**Mu’tah University**
**College of Graduate studies**

# Novel Methods for Enhancing the Performance of Genetic Algorithms

**By**
**Esraa Omar Alkafaween**

**Supervisor:**
**Dr. Ahmad Basheer Hassanat**

**A thesis submitted to the College of Graduate Studies in partial fulfillment of the requirements for the Master's degree in Computer Science in the Department of Information Technology, Mu’tah University.**

**الآراء الواردة في الرسالة الجامعية لا تُعبر**

**بالضرورة عن وجهة نظر جامعة مؤتة**

## ACKNOWLEDGMENT

**Allah, the Almighty**

Who gave me the strength, perception and insight for the completion of this work.

**My Parents**

I dedicate this work to my dear father who taught me success, and to my beloved mother who gave me compassion and love.

**My Supervisor**

Who conveys the light of his knowledge to others, who guides perplexed questioners to the right answers, who has stood by me every step of the way to completion of this work, every appreciation and respect to Dr. Ahmad Hassanat. Many thanks go also to Marion Cobban for editing this thesis.

**Esra'a Omar Alkafaween**

# TABLE OF CONTENTS

## LIST OF TABLES

# LIST OF FIGURES

## LIST OF EQUATIONS

## Abbreviations

| | |
|---|---|
| ATSP | Asymmetric Travelling Salesman Problem |
| COWGC | Cut on worst gene crossover |
| COWLRGC | Cut On Worst L+R Crossover |
| CX | Cycle Crossover |
| DCGA | Diversity Control Oriented Genetic Algorithm |
| DGA | Distributed Genetic Algorithm |
| DPGA | Dual Population Genetic Algorithm |
| FgGA | Fine grained Genetic Algorithm |
| GAs | Genetic Algorithms |
| GATC | Genetic Algorithm Time Complexity |
| GPGA | Global Parallel Genetic Algorithm |
| IBRGBWGM | Insert Best Random Gene Before Worst Gene Mutation |
| IBRGBRGM | Insert Best Random Gene Before Rand Gene Mutation |
| MOEA | Multi-Objective Evolutionary Algorithm |
| MPGA | Multi-population Genetic Algorithm |
| MTSP | Multi Travelling Salesman Problem |
| OX | Order Crossover |
| PDGA | Primal Dual Genetic Algorithm |
| PGA | Parallel Genetic Algorithm |
| PMX | Partially Matched Crossover |
| PR | Population reduction |
| RWS | Roulette Wheel Selection |
| RGIBNNM | Random Gene Inserted Beside Nearest Neighbour Mutation |
| SAC | Select Any Crossover |
| SAM | Select Any Mutation Algorithm |
| SBC | Select the Best Crossover |
| SBM | Select The Best Mutation Algorithm |
| SGA | Simple Genetic Algorithm |
| STSP | Symmetric Travelling Salesman Problem |
| SUS | Stochastic Universal Sampling |
| SWGLM | Swap Worst Gene Locally Mutation |
| TSP | Travelling Salesman Problem |
| WGIBNNM | Worst Gene Inserted Beside Nearest Neighbour Mutation |
| WGWRGM | Worst Gene With Random Gene Mutation |
| WGWWGM | Worst Gene With Worst Gene Mutation |
| WGWNNM | Worst Gene With Nearest Neighbour Mutation |
| WGWWNNM | Worst Gene With The Worst Around the Nearest Neighbour Mutation |
| WLRGWRGM | Worst LR Gene With Random Gene Mutation |

# Abstract

## Novel Methods for Enhancing the Performance of Genetic Algorithms

**Esra'a Omar Alkafaween**
**Mutah University, 2015**

Genetic algorithm (GA) is a branch of so-called evolutionary computing (EC) that mimics the theory of evolution and natural selection, where the technique is based on an heuristic random search. It is considered a powerful tool for solving many optimization problems.

Crossover and mutation are the key to success in genetic algorithms. Today, with the existence of several methods of crossover and mutation operators, our decision becomes more difficult to determine which method is best suited to each problem, and needs more trial and error.

In this thesis we propose new methods for crossover operator namely: cut on worst gene (COWGC), cut on worst L+R gene (COWLRGC) and Collision Crossovers. And also we propose several types of mutation operator such as: worst gene with random gene mutation (WGWRGM) , worst LR gene with random gene mutation (WLRGWRGM), worst gene with worst gene mutation (WGWWGM), worst gene with nearest neighbour mutation (WGWNNM), worst gene with the worst around the nearest neighbour mutation (WGWWNNM), worst gene inserted beside nearest neighbour mutation (WGIBNNM), random gene inserted beside nearest neighbour mutation (RGIBNNM), Swap worst gene locally mutation (SWGLM), Insert best random gene before worst gene mutation (IBRGBWGM) and Insert best random gene before random gene mutation (IBRGBRGM).

In addition to proposing four selection strategies, namely: select any crossover (SAC), select any mutation (SAM), select best crossover (SBC) and select best mutation (SBM). The first two are based on selection of the best crossover and mutation operator respectively, and the other two strategies randomly select any operator. So we investigate the use of more than one crossover/mutation operator (based on the proposed strategies) to enhance the performance of genetic algorithms.

Our experiments, conducted on several Travelling Salesman Problems (TSP), show the superiority of some of the proposed methods in crossover and mutation over some of the well-known crossover and mutation operators described in the literature. In addition, using any of the four strategies (SAC, SAM, SBC and SBM), found to be better than using one crossover/mutation operator in general, because those allow the GA to avoid local optima, or the so-called premature convergence.

**الملخص**

**طرق جديدة لتحسين أداء الخوارزمية الجينية**

**إسراء عمر الكفاوين**

**جامعة مؤتة، 2015**

الخوارزميات الجينية (Genetic Algorithms) هي إحدى أساليب الذكاء الاصطناعي والتي تعتبر فرع من فروع الحوسبة التطورية (Evolutionary Computation) حيث تعتمد على مبدأ التطور والانتخاب الطبيعي (Natural Selection)، ويمكن استخدامها في تحسين وإيجاد حلول للمسائل المعقدة ومسائل الحل الأمثل (Optimization Problems).

تعتبر عمليه التداخل الابدالي (Crossover) وعمليه الطفرة (Mutation) هما مفتاح النجاح في الخوارزميات الجينية. على مدار السنين ظهر العديد من الأنواع لهذه العمليات، و لكن تكمن الصعوبة في اختيار النوع الملائم من التداخل الابدالي والطفرة لحل مشكلة معينة، حيث نحتاج إلى المزيد من التجارب لإصدار القرار بشأن أي الأنواع من تلك العمليات أفضل.

في هذه الاطروحة قمنا بتقديم أنواع جديدة من التداخل الابدالي والطفرة، حيث أن كلا من هذه الأنواع يتبع دليلا معينا ليغير مسار الخوارزمية الجينية، بالاضافة إلى أننا قدمنا أربعة استراتيجيات تدعى: اختيار أفضل تداخل إبدالي (SBC)، اختيار اي تداخل إبدالي (SAC)، اختيار أفضل طفرة (SBM)، اختيار اية طفرة (SAM)، اثنان منها لعملية التداخل الابدالي واثنان لعملية الطفرة. تقوم هذه الاستراتيجيات على استخدام أكثر من تداخل إبدالي او طفرة في آن معا، استراتيجية افضل تداخل إبدالي واستراتيجية افضل طفرة تختار في كل مرة افضل عملية تداخل ابدالي/ طفرة من مجموعة من العمليات، بينما استراتيجية اختيار اي تداخل ابدالي / طفرة تختارفي كل مرة نوعا عشوائيا من الانواع الموجودة.

قمنا بتجربة الطرق والاسترتيجيات المقترحة على مسألة البائع المتجول (TSP)، حيث تبين من النتائج تفوق بعض العمليات المقترحة للتداخل الابدالي وللطفرة على عمليات تداخل ابدالي وعمليات طفرة موجودة مسبقا، بالاضافة الى انه تبين تفوق الاسترتيجيات الاربعة والتي تختار احدى طرق التداخل الابدالي/طفرة في كل جيل، بدلا من استخدام عملية تداخل ابدالي لوحدها او عملية طفرة لوحدها لجميع الاجيال. وهذ ادى الى تحسن اداءالخوارزمية الجينية، حيث انها تجنبت الوقوع في حل محلي (local optima) أو ما يسمى ب التقارب السابق لاوانه (premature convergence) مما يؤدي إلى تحسين أداء الخوارزمية الجينية من حيث الدقة.

# Chapter 1
# Introduction

Genetic algorithms (GAs) are adaptive heuristic techniques based on a natural selection mechanism, and are part of what is known as the evolutionary algorithm (EA). The basic principles of genetic algorithms were introduced by Johen Holland in the 1970's in the University of Michigan (Holland, 1975).

GA is a random, non-linear, and discrete process, that does not require a mathematical formulation, where Optima evolve from one generation to another (Man, Tang, & Kwong, 1996). It has emerged as important in solving large complex problems, which require a long time according to traditional programming techniques, compared with GA, which has a tremendous amount of alternative solutions, where the solution is often optimal or near to optimal, over an appropriate time (Hendricks, Wilcox, & Gebbie, 2014).

The GA has proved its strength and durability in solving many problems, and thus it is considered as an optimization tool (Golberg, 1989), (Whitley, 1994) and (Tsang & Au, 1996). This explains the increase in and expansion of its popularity among researchers in many areas, such as image processing (Paulinas & Ušinskas, 2015), speech recognition (Benkhellat & Belmehdi, 2012); (Gupta & Wadhwa, 2014), software engineering (Srivastava & Kim, 2009), computer networks (Mohammed & Nagib, 2012), robotics (Ayala & dos Santos Coelho, 2012), etc.

GA was inspired by the Darwinian theory of "survival of the fittest" (Zhong, Hu, Gu, & Zhang, 2005), (Mustafa, 2003) and (Eiben & Smith, 2003), by producing new chromosomes (individuals) through recombination (crossover) and mutation operations, i.e. the fittest individual is more likely to remain and mate. Therefore the inhabitants of the next generation will be stronger, because they are produced from strong individuals, i.e. the solution evolves from one generation to another.

Over the years, GAs have evolved from what was prevalent in the era of Holland (Bäck & Schwefel, 1993) to cope with some of the requirements such as optimizing multi-modal problems and time-dependent optima. Therefore, many of the types or extensions of the standard GA appeared, such as multi-population GAs and parallel GAs and these types have also developed. The ultimate goal is to bring diversity to the population and thus increase the efficiency of the GA.

## 1.1 Simple GAs

A simple (or Basic) GA (SGA) is depicted in Figure (1.1) (Al-Angari & ALAbdullatif, 2012).

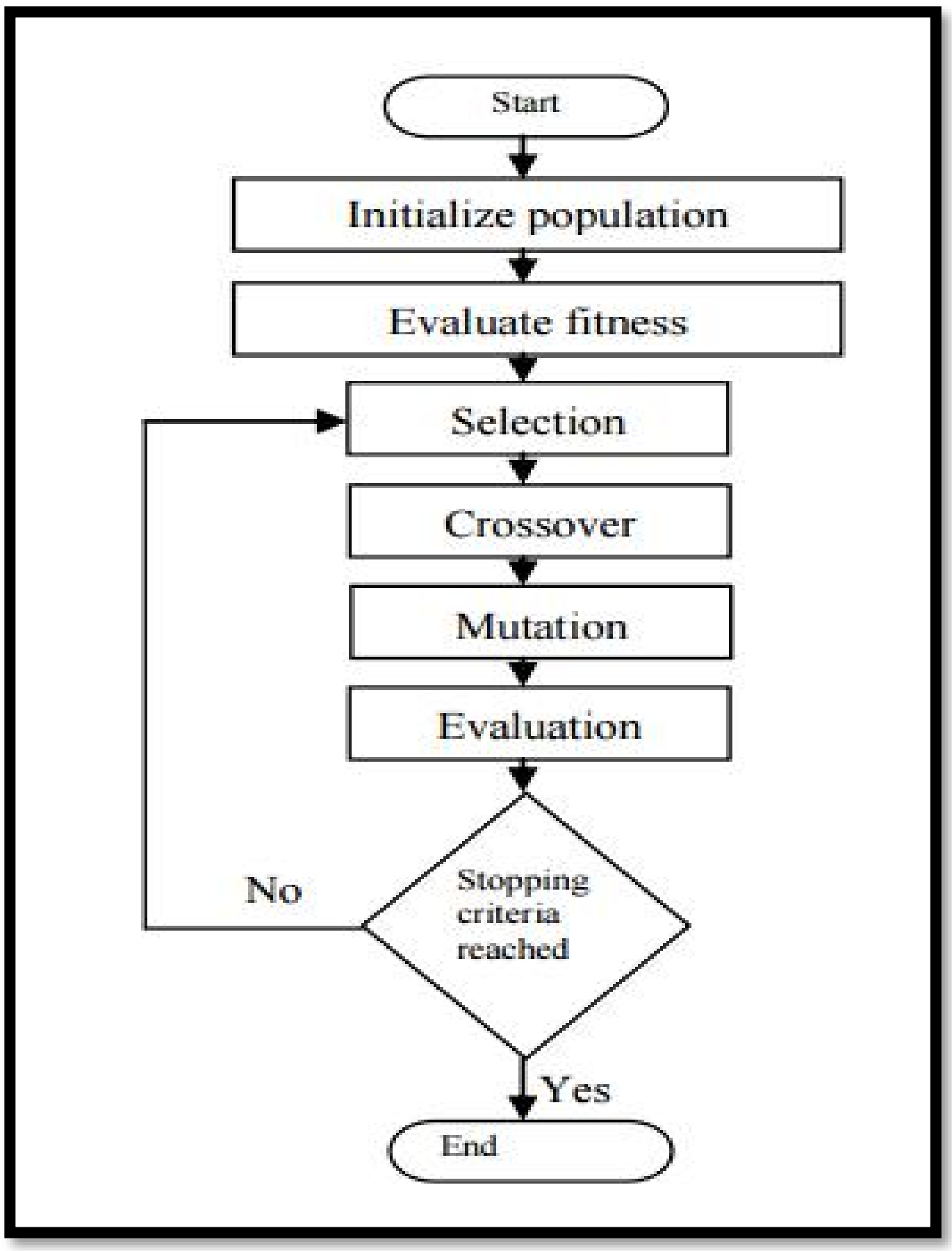

**Figure 1.1.** Flow chart of a typical GA **(Al-Angari & ALAbdullatif, 2012)**

The first step in the implementation of a GA is to connect between the real world and the world of the GA. This is called “encoding” or “representation” of the problem solutions (individuals) (Eiben & Smith, 2003). Encoding depends mainly on the problem to be solved, where there are many encoding techniques, and therefore an appropriate technique must be selected. Some encoding techniques are as follows:

1. Binary encoding: A chromosome in this type of encoding is represented using binary string; one example of the use of this type is the Knapsack problem (see Figure (1.2)).
2. Permutation encoding: Chromosomes in this type represent a position in a sequence, where this type is used in ordering problems, e.g. TSP (see Figure (1.3)).

Chromosome A : 101010101111101010
Chromosome B : 111010101010000111

**Figure 1.2**.Binary encoding example

| | |
|---|---|
| Chromosome A: | 652138974 |
| Chromosome B: | 287456913 |

**Figure 1.3.** Permutation encoding example

3. Value encoding: Each chromosome is represented using a sequence of some values. These values are possibly character, real number, etc. (Kumar R. , 2012) (see Figure (1.4)) that can be used in a neural network.
4. Tree encoding: Each chromosome is a tree of some objects (Kumar A. , 2013).This encoding is used in genetic programming (see Figure1.5).

| | |
|---|---|
| Chromosome A | left, up, right, down |
| Chromosome B | FDRTEWSQTOIUYKLPM |
| Chromosome C | 0.5,0.3,1.2,0.2,3.3,2.2 |

**Figure 1.4.** Value encoding example

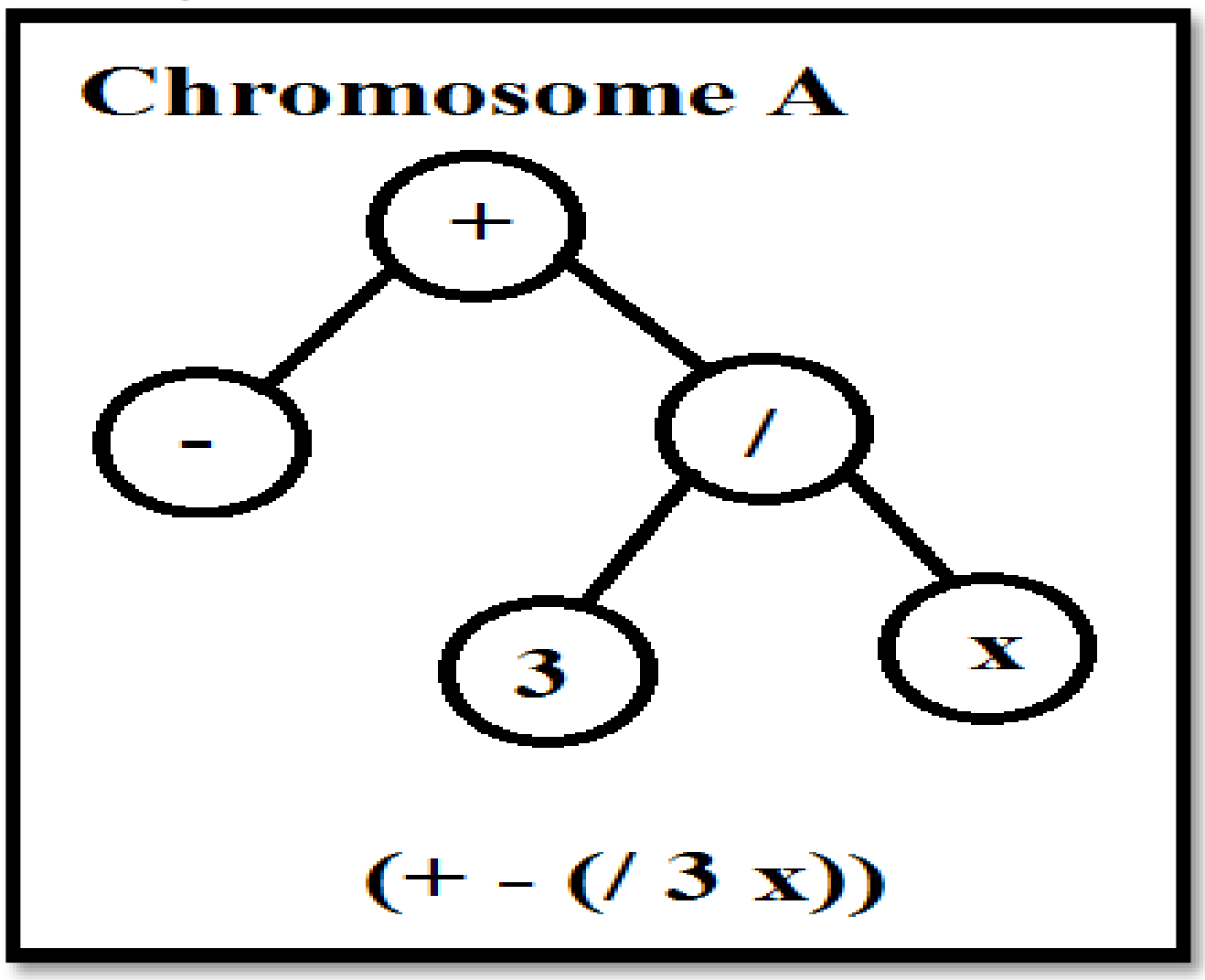

**Figure 1.5.** Tree encoding example

The next steps and the working principles of SGA and the developments of this simple algorithm are described as follows (Shyr, 2010):

1. **Start:** Create a random population of potential solutions (Michalewicz, 2013) (initial populations) consisting of *n* individuals.
2. **Fitness:** Evaluate the fitness value f(x) of each individual, x, in the population.
3. **New population**: Repeat the following steps to create a new population until completion of the new population.
4. **Selection (Reproduction)**: The process of choosing the “best” parents in the community for mating, “best” being defined based on the current problem, has an active role in solving premature convergence resulting from a lack of diversity in the population, therefore indentifying the appropriate selection technique is a critical step (Shukla, Pandey, & Mehrotra, 2015).There are several types of selection, including: Roulette Wheel, Elitism, Rank, Tournament, Stochastic Universal Sampling (SUS). We will mention the most common selection types, as follows:

a. Roulette Wheel Selection (RWS), also known as fitness proportionate (see Figure (1.6)), is a traditional selection technique, which assigns a fitness to each individual (chromosome) in the population using fitness function. The best solution is measured by fitness level which is used to link a probability of selection with each chromosome (Pedersen, 1998).

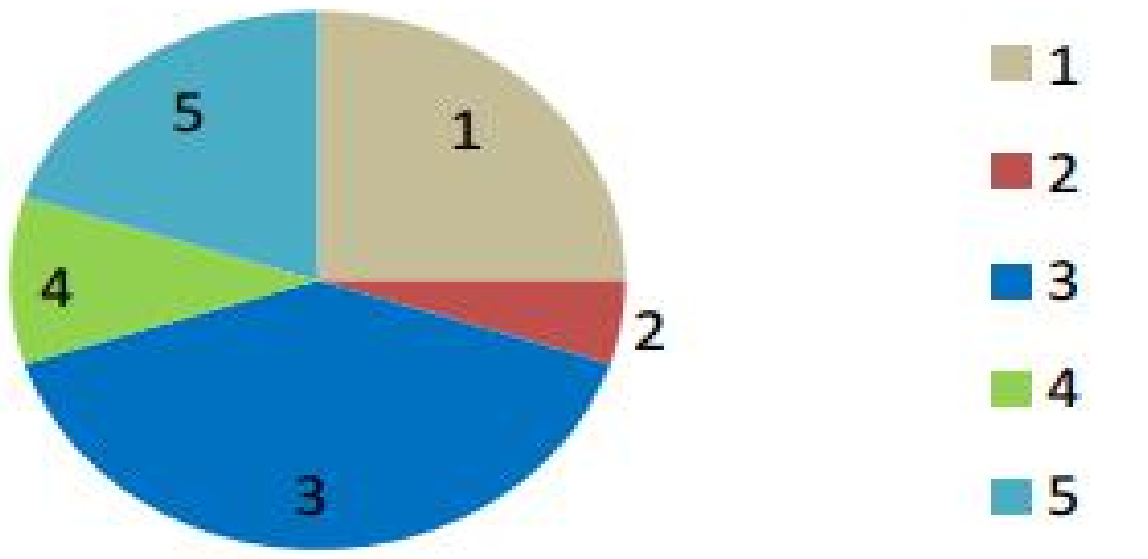

| Population | Fitness |
|---|---|
| 1 | 25.0 |
| 2 | 5.0 |
| 3 | 40.0 |
| 4 | 10.0 |
| 5 | 20.0 |

**Figure 1.6.** Roulette Wheel selection **(Sharma & Mehta, 2013)**

From the Figure above, the fitness of the third individual is found to be higher than the others, so it is expected that the roulette wheel will select the third individual over any other.

b. Elitism Selection: In the beginning, the best individuals are copied from the current generation and then the evolution proceeds (crossover, mutation).This prevents the loss of the best solution that was reached, so elitism helps in the rapid improvement of the performance of the GA (Sharma & Wadhwa, 2014).
c. Rank Selection: The individuals in this type are selected according to their rank, but first we need to sort the chromosomes based on their fitness value. The best chromosome will have rank (N) and the worst chromosome attains rank 1 (Shukla, Pandey, & Mehrotra, 2015).
d. Tournament Selection (Figure (1.7)): Using this type, two or more individuals are randomly chosen from a population; the fitter (individual with best fitness) is selected as a parent and inserted into the mating pool. The number of individuals who will compete is called the size of the tournament (Noraini & Geraghty, 2011). This method revealed its effectiveness in much of the research because of several features, such as less time complexity (Oladele & Sadiku, 2013).

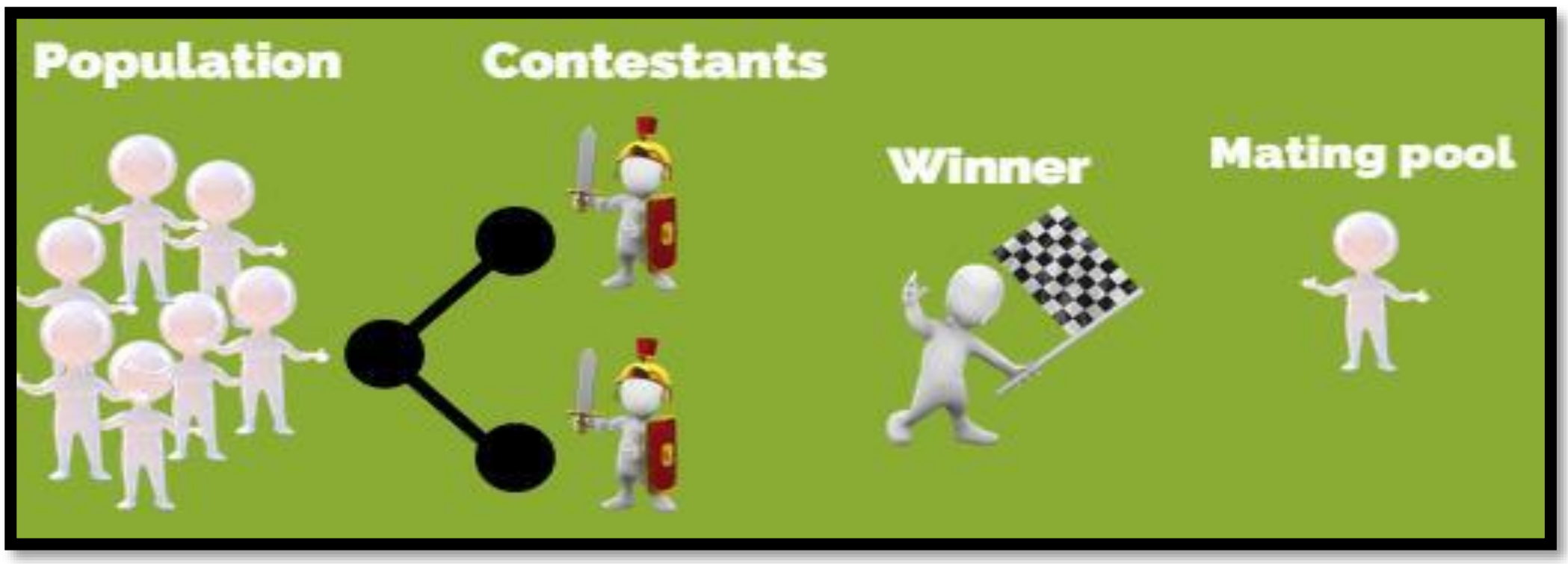

**Figure 1.7.**Tournament selection (Tournament size=2)

5. **Crossover (Recombination)**: This process takes two parents (chromosomes) to create a new offspring by switching segments of the parent genes. It is more likely that the new offspring (children) will contain good parts of their parents, and consequently perform better as compared to their ancestors. A number of different crossover operators can be used in GAs (Kaya & Uyar, 2011), beginning with one-point crossover and two-point crossover, then evolving into several techniques to accommodate some situations. Here is a brief definition of the two types:

a. One-point crossover: Randomly determines the point in the chromosomes, and then exchanges genes after this point between the parents to produce two of the children (see Figure (1.8)).

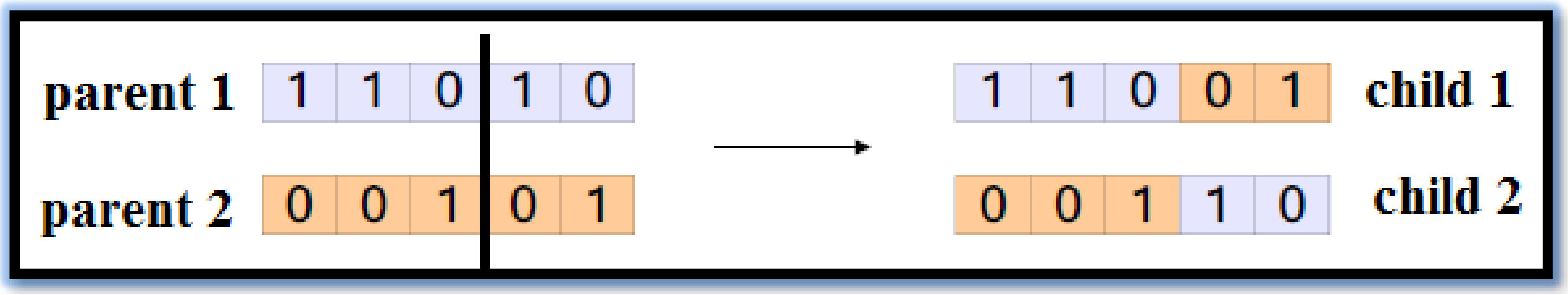

**Figure 1.8.**One-Point crossover

b. Two-point crossover: In this crossover two points are selected on the parent chromosomes. Exchange of genes between the two points in the parents, to produce two of the children (see Figure (1.9)).

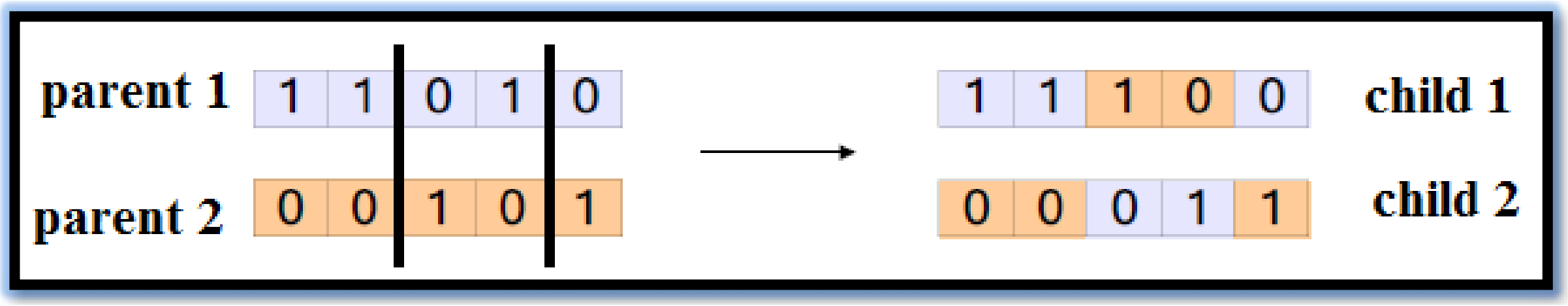

**Figure 1.9.**Two-Point crossover

6. **Mutation:** This is where there is a change or a switch between specific genes within a single chromosome to create chromosomes that provide new solutions for the next generation, with the aim of obtaining the best possible solutions, and thus introduce a certain level of diversity to the population, and as a result this also does not fall into the local optimum (Korejo, Yang, Brohi, & Khuhro, 2013) (see Figure (1.10)). There are many types of mutation, beginning with bit inversion mutation (see Figure (1.11)) and evolving into many types that fit several locations.

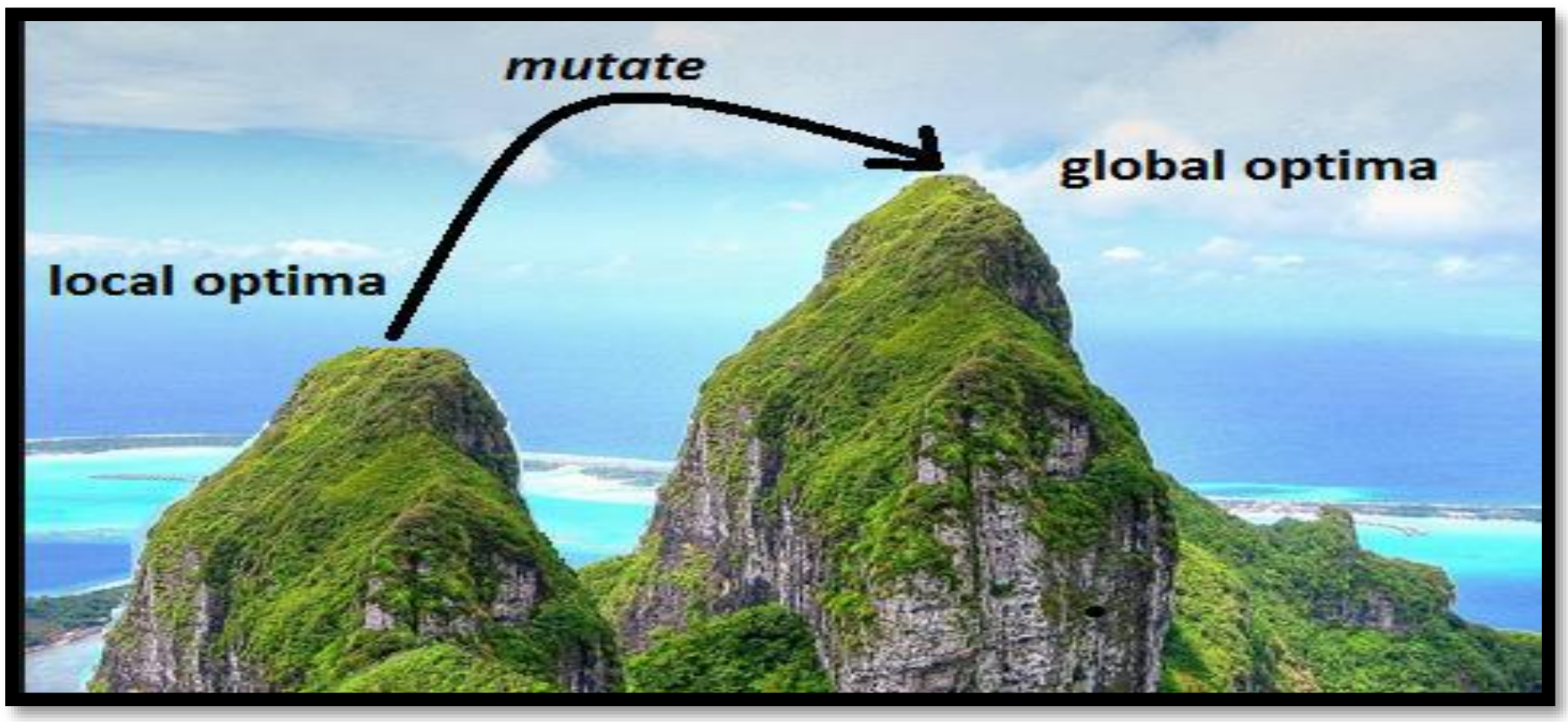

**Figure 1.10.**Overcoming local optima by mutation

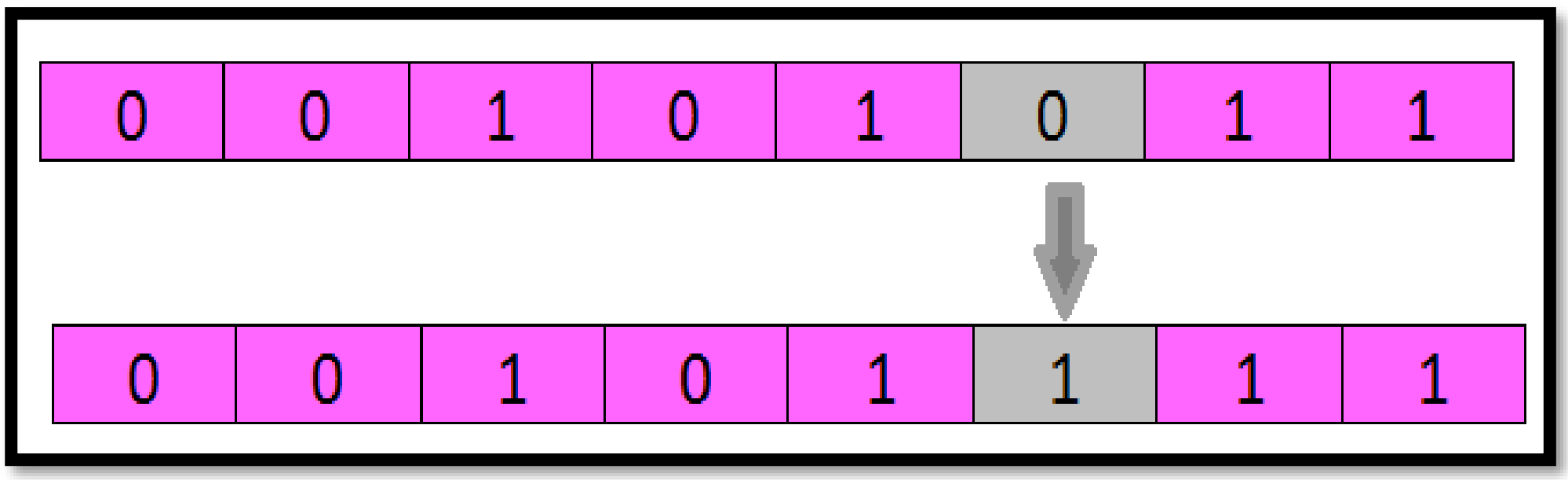

**Figure 1.11.** Example of mutation in a binary string

7. **Termination (stopping) criteria:** There are several stopping conditions that have been applied to the simple GA (Safe, Carballido, Ponzoni, & Brignole, 2004):

1- reach to the highest limit of generations.
2- The opportunity to make changes in future generations has become negligible, which means the chance of diversity in the population has become low (Eiben & Smith, 2003); (Safe, Carballido, Ponzoni, & Brignole, 2004).
3- Improving fitness remains below the threshold value (Eiben & Smith, 2003).

## 1.2 GA Parameters

There are four basic parameters of GA, including the following:

1. Crossover rate: This determines the number of times a crossover occurs in the chromosomes in one generation, and is between 0%-100%; crossover rate is a delicate matter.
2. Mutation rate: This determines how many genes mutate in one generation; it is between 0%-100%, and is also a sensitive matter.
   Increases or decreases in the rate of mutation and crossover can have both a negative and positive effect, as suggested by some studies, such as (Tuson & Ross, 1995) and (Davis, 1991).
3. Population size: Our selection of the size of the population is a sensitive issue, if the size of the population is very small, which means little search space, so it is possible to reach the local optimum. However, if the size of the population is too large, this would increase the area of search and increase the mathematical load and thus the process becomes slow (Roeva, Fidanova, & Paprzycki, September2013), therefore, the size of the population must be reasonable.
4. Number of generations: A number of cycles before termination. In some cases, hundreds of loops are sufficient and others are not, and this depends on the complexity of the problem.

All these GA parameters are important because they determine the quality of the solution (Yang, 2002).

## 1.3 Thesis Contribution

The efficiency of a GA is based on the appropriate choice of the genetic parameter (selection, crossover and mutation) and strategy parameters (Eiben, Michalewicz, Schoenauer, & Smith, 2007) associated with them, such as crossover ratio and mutation ratio (Yang, 2002). Many researchers have shown the effect of the two operators – crossover and mutation – on the success of the GA, and where success lies in both, whether crossover is used alone, mutation alone or both, as in (Spears, 1992) and (Deb & Agrawal, 1999).

This thesis proposes three new crossover operators, and ten new mutation operators to enhance the performance of the GA, in addition to proposing four selection strategies: two strategies for crossover, that use more than one crossover operator and two strategies for mutation that use more than one mutation operator. (The proposed work will be discussed in detail in Chapter 3 and Chapter 4).

## 1.4 Thesis Overview

This thesis consists of 6 chapters. The first chapter presents a brief introduction to the GA, then the simple GA and its components are explained in detail, as well as development of these components. Finally, the important parameter in the GA and its impact on the search process are described and explained.

The remaining chapters will be presented in this thesis as follows: in the second chapter we shall review the related work in the GA field. We aim to highlight previous work over the years, in terms of types of crossover and mutation. In addition, the evolution of the standard GA that uses a multi-population (multi-population GA) and previous work carried out in this area, as well as the parallel GA and some of the previous work are described and explained. A summary of the chapter is presented at the end.

In the third chapter, we shall propose new methods for crossover operator and new strategies that use multi crossover operator. We shall detail how the new methods and strategies work, and compare these methods with pre-existing methods by applying these new types to the specific problem. We shall also investigate the effect of using these methods on the performance of the GA.

New mutation operators and new mutation strategies that use more than one mutation operator at the same time will be discussed in Chapter 4. We shall detail how the new methods and strategies work, and compare mutation methods and mutation strategies with pre-existing methods, as well as investigating the effects of using these methods on the performance of the GA.

Chapter 5 describes the results of applying both of the proposed methods (crossover and mutation) by setting up experiments on a known problem using different benchmark data.

The last chapter summarizes the overall results, draws final conclusions and indicates possible future directions.

### 1.5 Summary

This chapter has presented an introduction to GAs, and a simple summary of the emergence of a GA.

GAs enable the production of suggestions for solving various problems, depending on genetics. In general, GAs are concerned with how to produce new chromosomes (individuals) that possess certain features through recombination (crossover) and mutation operators, which happen to inherited groups in order to make new individuals. GAs are dependent on several components, including: selection, crossover, mutation, representation (encoding) the problem and appropriate fitness function.

Much research has demonstrated the importance of GA parameters, such as: mutation rate, crossover rate, population size, and the extent of its importance in increasing the efficiency of the GA.

# Chapter2
# Literature Review

This chapter describes the basic GA, by means of answering several questions, such as: how have GAs evolved over the years? Where have developments in GA showed its superiority in many areas? Some previous studies that have been conducted for simple GA, multi-population GA, and the parallel GA are also highlighted.

## 2.1 Genetic algorithm (GA) revisited

GAs have a number of alternative solutions, and the resultant solution is closer to the optimal, where it begins with a number of random solutions (population). These solutions (individuals) are then encoded according to the current problem, and the quality of each individual is evaluated through a fitness function, after which the current population changes to a new population by applying three basic operators: selection (as described in the first chapter), crossover and mutation.

### 2.1.1 Crossover operator

In a switch between opposite values of the parent chromosomes, to create a new chromosome, it is more likely that the new chromosome will contain good parts of the parents and will consequently perform better as compared to them. Over the years, many types of crossover have been developed, and comparisons have been made between these different types. They began with the one-point crossover, then evolved into several techniques to accommodate a number of situations, including: Uniform crossover (Syswerda, 1989), Multi-point crossover (Spears & De Jong, 1990), Heuristic crossover (Grefenstette, Gopal, Rosmaita, & Van Gucht, 1985), Ring crossover (Kaya & Uyar, 2011), Arithmetic crossover, and for the order-based problem, the Partially Matched crossover (PMX) (Goldberg & Lingle, 1985), Cycle crossover (CX) (Oliver, Smith, & Holland, 1987), Order crossover (OX) (Golberg, 1989) and some other types.

Our selection of the type of crossover depends mainly on the type of encoding used. This can sometimes be very complicated but normally improves the performance of the GA (Man, Tang, & Kwong, 1996). However, the question of whether this kind of crossover is better than the others remains open. In this regard we cannot draw a significant conclusion about which is better, because most of the comparisons between the different types were conducted on a small group of test problems and more trial and error was needed. In order to overcome this problem, several researchers have developed new types of GA that use more than one crossover operator at the same time (Dong & Wu, 2009) (Hilding & Ward, 2005). Here is a brief definition of some of the more common types:

a. Modified crossover: This operator for permutation problems, which is an extension of the one-point crosover, so choose the cut point at random from the first parent. Then, the first child is created by appending genes (before cutting point) in the second parent to the first part. Repair the rest of the gene to prevent duplicates, (see Figure (2.1)).

| | | | | | | | | |
|---|---|---|---|---|---|---|---|---|
| parent 1 : | **1** | **2** \| | **5** | **6** | **4** | **3** | **8** | **7** |
| parent 2 : | 1 | 4 | 2 | 3 | 6 | 5 | 7 | 8 |
| offspring : | **1** | **2** | 4 | 3 | 6 | 5 | 7 | 8 |

**Figure 2.1.**Modified crossover **(Potvin, 1996)**

b. Uniform crossover: Its popularity was spread in 1989 by Syswerda (Syswerda, 1989). This crossover is used with a certain fixed ratio ($p_c$) to integrate (mixing) of genes between parents (Figure (2.2) (Magalhães-Mendes, 2013)).

| | | | | | | | | |
|---|---|---|---|---|---|---|---|---|
| Parent 1 | 0.32 | 0.22 | 0.34 | 0.89 | 0.23 | 0.76 | 0.78 | 0.45 |
| Parent 2 | 0.12 | 0.65 | 0.38 | 0.47 | 0.31 | 0.56 | 0.88 | 0.95 |
| Random number | 0.32 | 0.22 | 0.34 | 0.89 | 0.23 | 0.76 | 0.78 | 0.45 |
| Prob.Cross=0.7 | < 0.7 | < 0.7 | < 0.7 | > 0.7 | < 0.7 | > 0.7 | > 0.7 | < 0.7 |
| Offspring1 | 0.32 | 0.22 | 0.34 | 0.89 | 0.23 | 0.76 | 0.78 | 0.45 |
| Offspring2 | 0.12 | 0.65 | 0.38 | 0.47 | 0.31 | 0.56 | 0.88 | 0.95 |

**Figure 2.2.**Uniform crossover **(Magalhães-Mendes, 2013)**

c. Arithmetic crossover (Figure (2.3)): In this type parents are combined linearly for the production of children using the following formulae (Yalcinoz, Altun, & Uzam, 2001):

$$C_i^{gen+1} = a.C_i^{gen} + (1-a).C_j^{gen} \quad (1)$$

$$C_j^{gen+1} = (1-a).C_i^{gen} + a.C_j^{gen} \quad (2)$$

where $a$ is a random weighting factor in the range [0,1] , $C_i^{gen}$ is the first parent, and $C_j^{gen}$ is the second parent **.**

a = 0.7

| | | | | |
|---|---|---|---|---|
| parent 1 | (0.3) | (1.4) | (0.2) | (7.4) |
| parent 2 | (0.5) | (4.5) | (0.1) | (5.6) |
| child 1 | (0.36) | (2.33) | (0.17) | (6.87) |
| child 2 | (0.402) | (2.981) | (0.149) | (5.842) |

**Figure 2.3.** Arithmetic crossover

d. Partially Matched crossover (PMX): We choose two points randomly from chromosomes; the part between the two cut points is called the matching (mapping) section; the first section of the first chromosome is copied to the second child, and the second section from the second chromosome is copied to the first child. The rest of the genes are copied to the offspring, and the genes that are repeated in the same chromosome are replaced based on the matching section (Figure (2.4)).

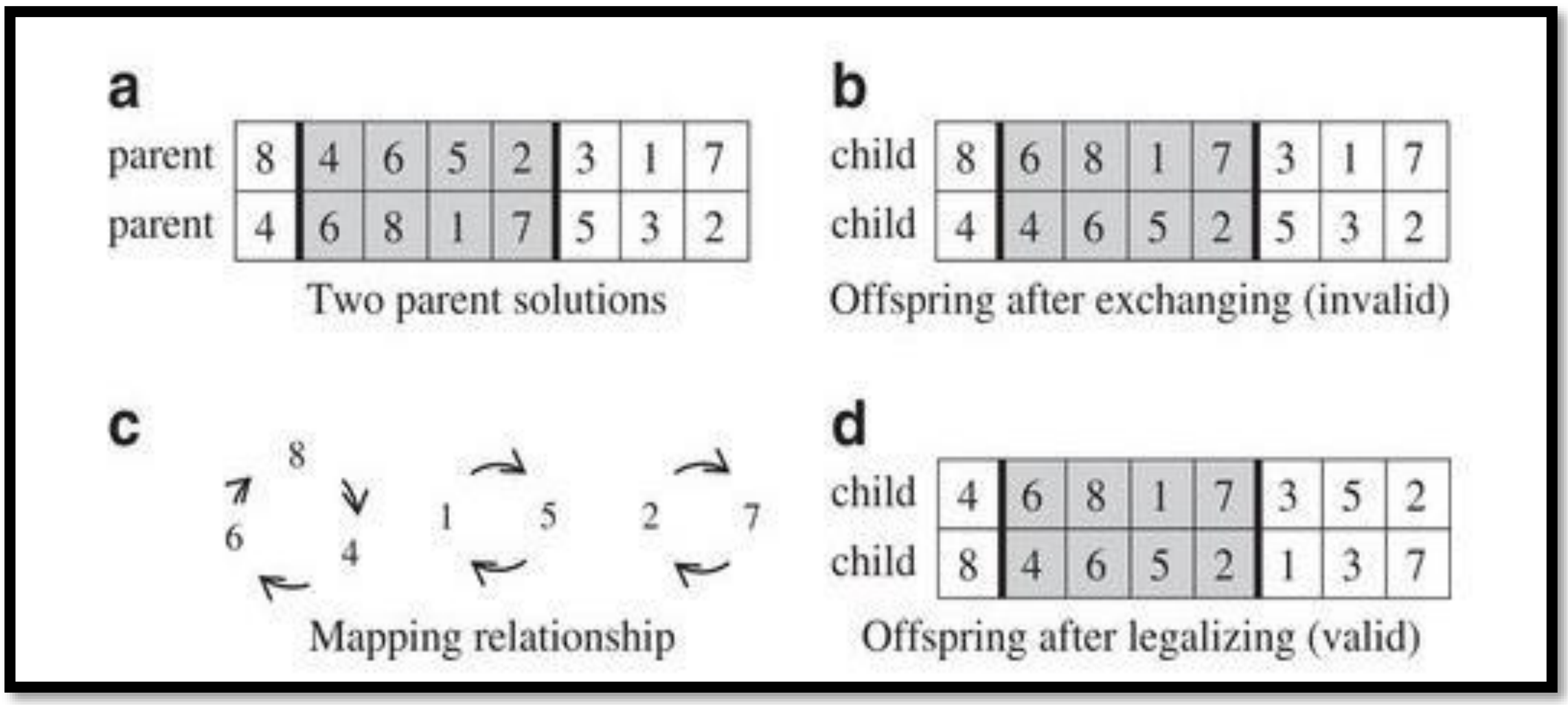

**Figure 2.4**. PMX crossover **(Lee, Wang, & Lee, 2015)**

### 2.1.2 Mutation operator

Mutation is one of the most important stages of the GA because of its impact on the exploration of global optima, where it is changed or is switched between specific genes within a single chromosome to create chromosomes that provide new solutions for the next generation.

Classical mutation (Bit-flip Mutation) was developed by Holland to deal with different encoding problems (e.g. TSP) that no longer fit (Larrañaga, Kuijpers, Murga, Inza, & Dizdarevic, 1999).Therefore, several types of mutation of various types of encoding were proposed, including::

Gaussian Mutation, Exchange Mutation (Banzhaf, 1990), Displacement Mutation (T I, 1992), Uniform Mutation and Creep Mutation (Soni & Kumar, 2014), Inversion Mutation (Fogel, 1990) and some other types. The problem lies in our selection of those types to solve a specific problem, and our decision becomes more difficult and needs more trial and error. In order to overcome this problem, several researchers have developed new types of GA that use more than one mutation operator at the same time (Hong, Wang, Lin, & Lee, 2002), (Hilding & Ward, 2005) and (Hong, Wang, & Chen, 2000).

Some of these types are described as follows (these types can be used for permutation encoding):

1. Exchange Mutation: Select two genes randomly and switch between their positions (Figure (2.5)).
2. Scramble Mutation: Select a subset of the genes randomly, and rearrange genes randomly in those locations (Figure (2.6)).
3. Inversion Mutation: Select two genes at random, then reverse the subset between them (Figure (2.7)).
4. Insert Mutation: Choose two of the genes randomly, convey the second to follow the first, then shift the rest of the genes (Figure (2.8)).

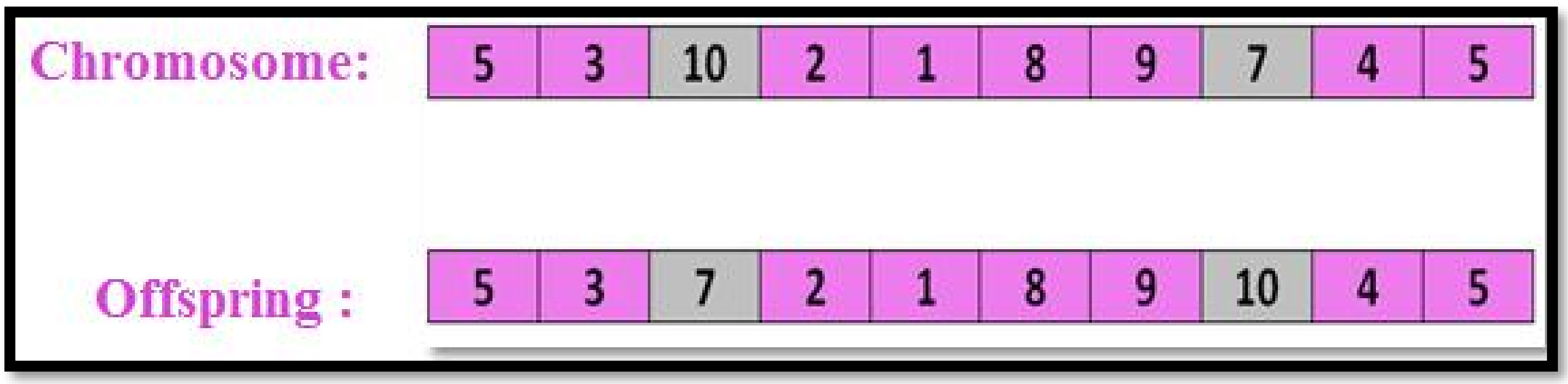

**Figure 2.5.** Exchange Mutation

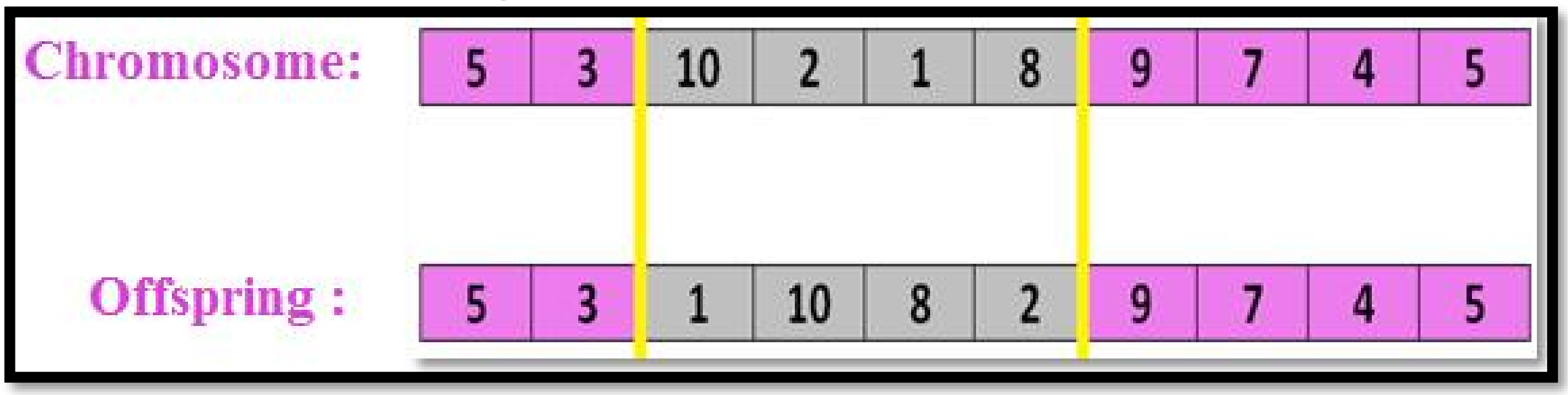

**Figure 2.6.** Scramble Mutation

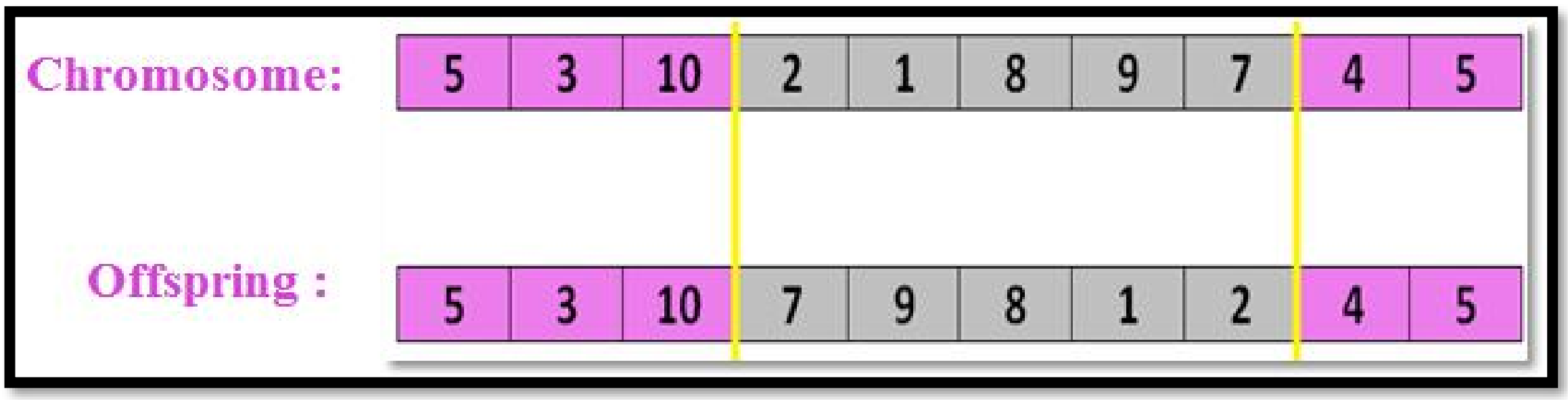

**Figure 2.7** Inversion Mutation

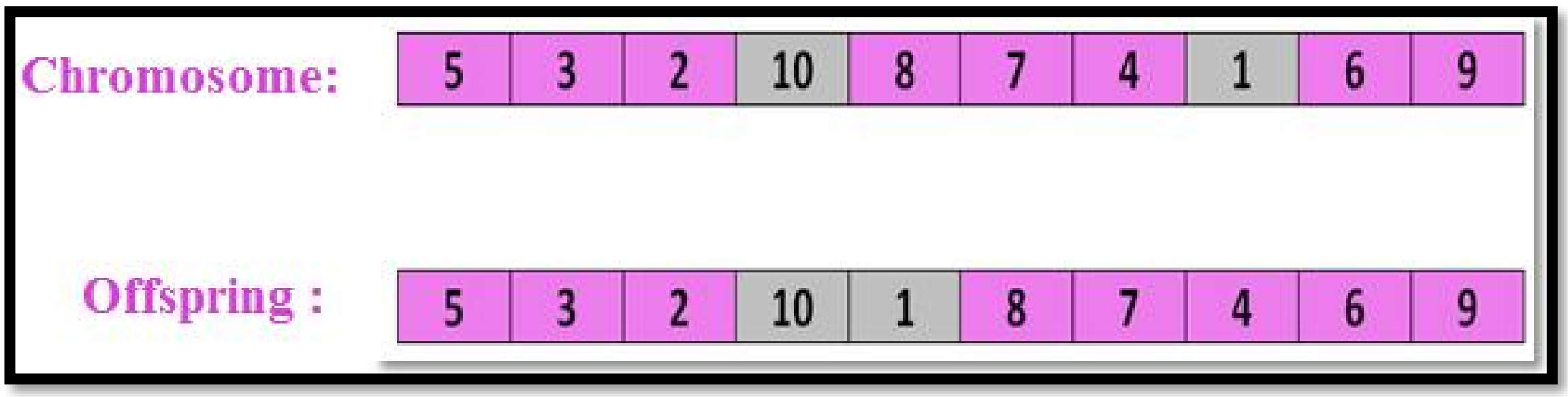

**Figure 2.8.** Insert Mutation

5. Rearrangement (Sallabi & El-Haddad, 2009): This method applies to the entire population in the Traveling Salesman Problem (TSP), but in this thesis we have made as a mutation operator ( see Figure (2.9)), in this operator the cost between any two adjacent cities (*city* $_i$ , *city* $_j$) denoted by $c_{i,j}$ , where i=1,2,3,...n-1 and j=i+1. The goal of that process to get the maximum cost between all the adjacent cities (max ($c_{i,j}$)), then *city* $_i$ is replaced by three other cities, one at a time. These cities are located on three different locations, i.e. in the beginning and the middle and the end, the original Location plus the best position will be accepted.

```
i = 1
S0 = S
max = 0
While i < n − 1
        If c_{i,i+1} > max
                Begin
                    Max = c_{i,i+1}
                    X = i
                End
S1 ← Swap city_x with city_1
S2 ← Swap city_x with city_{n/2}
S3 ← Swap city_x with city_n
S ← min (S0, S1, S2, S3)
Return (S)
End
```

**Figure 2.9.**The rearrangement operation procedure

### 2.1.3 Literature review (GAs)

Since the definition of the key principles of GAs provided by Holland, many reports and literature have emerged, such as: (Davis, 1991), (Beasley, Martin, & Bull, 1993), (Beasley, Bull, & Martin, 1993) and (Srinivas & Patnaik, 1994).

Zhong et al, " Comparison of performance between different selection strategies on simple genetic algorithms". Presented a comparative study of the selection approaches (roulette wheel, tournament selection), and applied seven test functions. This study showed that the tournament selection gave a better performance compared to the roulette wheel technique (Zhong, Hu, Gu, & Zhang, 2005).

Raja and Bhaskaran, " Improving the Performance of Genetic Algorithm by reducing the population size". Proposed a new methodology called population reduction (PR) to obtain the best individuals in the initial population, so that by division of the initial population into groups and applying the tournament selection to each group to choose the best individual, and after obtaining the best individuals of all the groups, these individuals enter into the normal GA. This methodology has been applied to the 0/1 knapsack problem, and the paper studied the effect of each of the following variables to find the best parameter of this method: 1-Selection Mechanisms: researchers chose Roulette Wheel selection, Rank-based selection and Tournament selection. 2-Population Size: applied different population size. 3-Crossover: applied single point crossover, two point crossover and uniform crossover. 4-Crossover Rate: used the variable rates in each experiment. 5-Mutation Rate: used the various rates in each experiment. Researchers compared this methodology with the simple GA, and from their experimental results it was determined that this method produces better results with less time than SGA. Also tournament selection is the best selection method with population reduction (PR) as compared with other selection methods (Raja & Bhaskaran, 2013).

Chiroma et al, "Correlation Study of Genetic Algorithm Operators: Crossover and Mutation Probabilities". Conducted a survey of previous research that discussed the best values of the critical variables that control the quality of the solution, to serve as a guide for researchers in the future. These variables are the size of the population, mutation rate, and crossover rate. They also discussed the three variables in terms of increase or decrease in their proportion. If there is a large population size, there is high efficiency in the search and this is better than a small size of population that generates inferior results. The size of the population in this research ranged from (12 - 4000); however, this represents a moderate population size. For crossover rate, a decrease or increase in the ratio also leads to a lack of exploration or ignores some of the solutions. The crossover rate ranged from (0.1 -1), and the mutation rate ranged from (0.001 - 1) (Chiroma, Abdulkareem, Abubakar, Zeki, Gital, & Usman, 2013).

Shukla et al, " Comparative Review of Selection Techniques in Genetic Algorithm". Presented a comparative study of various selection methods using many factors, such as time complexity and convergence rate. The paper shows that the tournament selection is more effective than other methods (Shukla, Pandey, & Mehrotra, 2015).

Vekariaet al, " Selective crossover in genetic algorithms: An empirical study". Suggested a crossover called "selective crossover" that mimics Dawkin's theory of natural evolution, and in storing the knowledge of previous generations it has been used as an extra real-value vector. This crossover method was compared with two types of crossovers: two-point and uniform crossover, and the results showed that this outperformed some of the other methods (Vekaria & Clack, 1998).

(Potvin, 1996) and (Larrañaga, Kuijpers, Murga, Inza, & Dizdarevic, 1999) presented a review of how to represent the travelling salesman problem, in addition to explaining the advantages and disadvantages of different crossover and mutation operations. Singh and Singh proposed an enhanced edge recombination crossover for solving the travelling salesman problem (TSP) (Singh & Singh, 2014).

Hong et al, "Evolution of appropriate crossover and mutation operators in a genetic process*". Proposed an algorithm called the Dynamic Genetic Algorithm (DGA) in order to apply more than one crossover and mutation at the same time. The algorithm automatically selects the appropriate crossover and appropriate mutation, and also adjusts the crossover and mutation ratios automatically based on the evaluation results of the respective children in the next generation. The algorithm was compared with the simple GA that commonly uses only one crossover process and one process of mutation. Their results showed enhancement of the GA performance (Hong, Wang, Lin, & Lee, 2002).

Ray et al, "New operators of genetic algorithms for traveling salesman problem". Proposed a new algorithm called SWAP_GATSP with two new operators: Knowledge-Based Multiple Inversion operator, which was done before selection and Knowledge-Based Swapping operator where the process was carried out after the process of crossover. This algorithm was applied to the travelling salesman problem for various data sets, and their results showed superiority as compared to other methods (Ray, Bandyopadhyay, & Pal, 2004).

Hilding and Ward, "Automated Crossover and Mutation Operator Selection on Genetic Algorithms". Proposed an Automated Operator Selection (AOS) technique, by which they eliminated the difficulties that appear when choosing crossover or mutation operators for any problem. In their work, they allowed the GA to use more than one crossover and mutation operators, and took advantage of the most effective operators to solve problems. The operators were automatically chosen based on their performance, and therefore the time spent choosing the most suitable operator was reduced. The experiments were performed on the 0/1 Knapsack problem. This approach proved its effectiveness as compared with the traditional GA (Hilding & Ward, 2005). Many studies also wanted to use more of the crossover and mutation approach, as in (Hong, Kahng, & Moon, 1995), (Herrera, Lozano, Pérez, Sánchez, & Villar, 2002), (Elsayed, Sarker, & Essam, 2011) and (Osaba, Onieva, Carballedo, Diaz, & Perallos, 2014).

Ahmed, "Genetic algorithm for the traveling salesman problem using sequential constructive crossover operator". Proposed sequential constructive crossover (SCX) to solve the TSP problem, and the basic idea of this method is to choose a point randomly called the crossover site, then employ a method of sequential constructive crossover before the crossover point by using better edges. The rest of the genes after the crossover site are exchanged between parents to get two children; if there is already a gene, then replace it with an unvisited gene (Ahmed, 2010).

Kaya and Uyar, "A novel crossover operator for genetic algorithms: ring crossover". Proposed a new crossover called "ring crossover" (RC). In this type, parents were arranged in the form of a ring, and then a cut point was chosen at random. The other point was the length of the chromosome; the first child arises from the point (first cut) in the clockwise direction and the other child arises in the counterclockwise direction. They applied this type of crossover to six functions such as: Sphere Function , Axis Parallel Hyper-Ellipsoid Function, Rotated Hyper-Ellipsoid Function, Normalized Schwefel Function, Generalized Rastrigin Function and Rosenbrock's Valley Function, and the proposed operator obtained results with better performance than the other types of crossover studied (Kaya & Uyar, 2011).

Ismkhan and Zamanifar, "Study of some recent crossovers effects on speed and accuracy of genetic algorithm, using symmetric travelling salesman problem". Review some recent crossover operators, including: PMX, extended PMX (EPMX), Greedy Subtour Crossovers (GSXs), Distance Preserving Operator (DPX), Unnamed Heuristic Crossover (UHX) and Very Greedy Crossover (VGX).These operators were applied to the TSP problem, and comparisons conducted between them in terms of accuracy and time. The results showed that the GA that uses heuristic in crossover, e.g. UHX and DPX, is more accurate than when using non-heuristic crossovers, e.g. PMX and EPMX (Ismkhan & Zamanifar, 2015).

Hong et al, "A Dynamic Mutation Genetic Algorithm". Proposed a Dynamic Mutation Genetic Algorithm (DMGA) to apply more than one mutation at the same time to generate the next generation. The mutation ratio is also dynamically adjusted according to the progress value that depends on the fitness of the individual, and so decreases the ratio of mutation if the mutation operator is inappropriate, and *vice versa,* increasing the ratio of mutation if the operator is appropriate (Hong & Wang, 1996).

Louis and Tang proposed a new mutation called Greedy-swap mutation, so that two cities are chosen randomly in the same chromosome, and switching between them if the length of the new tour obtained is shorter than the previous ones (Louis & Tang, 1999). Soni and Kumar studied many types of mutations that solve the problem of travelling salesmen (Soni & Kumar, 2014).

Dong and Wu, "Dynamic Crossover and Mutation Genetic Algorithm Based on Expansion Sampling". Proposed a dynamic crossover rate, where the crossover rate is calculated through the ratio between the Euclidean distance between two individuals and the Euclidean distance between the greatest and lowest fitness of the individual in the population. The process of crossover between individual "long distance individuals" is thus effective because of differences among themselves, and this would avoid inbreeding and thus overcome premature convergence (Dong & Wu, 2009).

Deep and Mebrahtu," Combined mutation operators of genetic algorithm for the travelling salesman problem". Proposed an Inverted Exchange mutation and Inverted Displacement mutation, which combine inverted mutation with exchange mutation and also combines inverted mutation with displacement mutation. The experiment was performed on the TSP problem and the results were compared with several operators that already exist (Deep & Mebrahtu, 2011).

Dong and Wu, "Dynamic Crossover and Mutation Genetic Algorithm Based on Expansion Sampling". Proposed a dynamic mutation probability, where the mutation rate is calculated by the ratio between the fitness of the individual and the fittest in the population. This ratio helps the algorithm to get out of local optima and also leads to diversification of the population (Dong & Wu, 2009). Patil and Bhende, "Comparison and Analysis of Different Mutation Strategies to improve the Performance of Genetic Algorithm". Presented a study of the various mutation-based operators in terms of performance, improvement and quality of solution. A comparison was made between Dynamic Mutation algorithm, Schema Mutation Genetic Algorithm, Compound Mutation algorithm, Cluster-based adaptive mutation algorithm and Hyper Mutation-based Dynamic Algorithm (Patil & Bhende, 2014).

## 2.2 Diversity and Multi-Population Genetic Algorithm (MPGA)

Many factors must be taken into account when establishing the population, including the following (Gupta & Ghafir, 2012): the diversity, selection pressure, the problem difficulty, the fitness function, and the search space (Figure (2.10)).

Diversification is how chromosomes are different from each other in the population. We are always interested in obtaining diversification, which this lacks, and is the main reason for the population to reach local minima, so that genetic processes do not produce offspring superior to the parents. This is what is known as premature convergence (Nicoară, 2009).

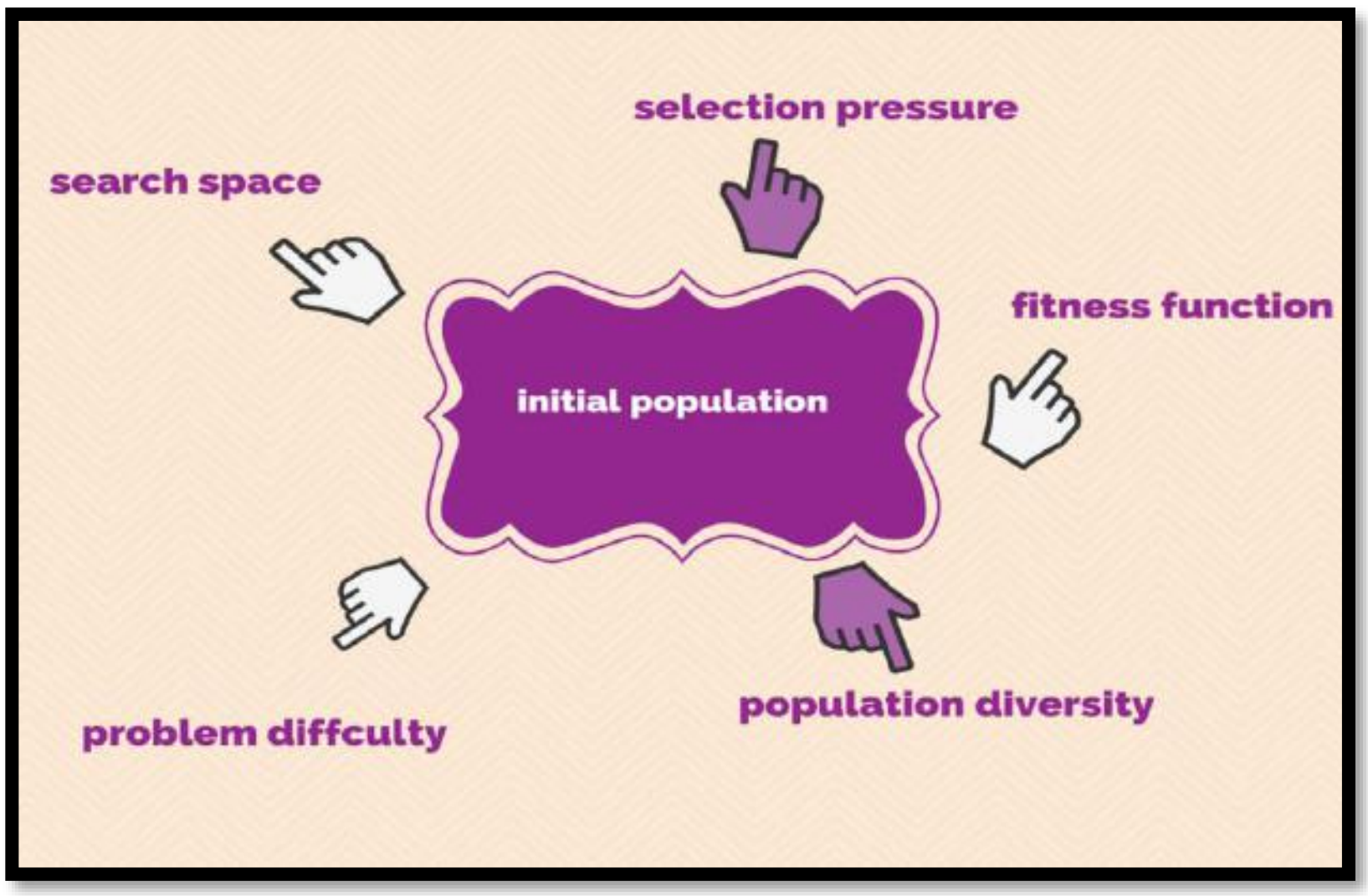

**Figure 2.10.** Factors affecting the initial population

In recent years, many techniques have emerged to facilitate and maintain the diversification of the population and thus reduce premature convergence (Korejo, Yang, Brohi, & Khuhro, 2013).This would help improvement through global exploration support and gain access to many global and local optima, in addition to dealing with the multi-modal function (Gupta & Ghafir, 2012); (Siarry, Pétrowski, & Bessaou, 2002). These techniques include:

**1. Multi-Population Genetic Algorithm (MPGA):**

This is also called the “island model”, the principle of which is to divide the population into a sub-population, where it is more likely that each island (sub-population) follows a different search path (Whitley, Rana, & Heckendorn, 1999). Good individuals are exchanged between the sub-population through the migration process (see Figure (2.11) (Siarry, Pétrowski, & Bessaou, 2002)), and new individuals are generated through the crossover operator. This allows exploration of areas not yet discovered. The number of individuals to be replaced between the parts of the population (sub-population) is called the migration rate, and this allows control of the degree of diversity within the sub-population (Siarry,

Pétrowski, & Bessaou, 2002). Another variable to encourage the diversity of sub-population is migration interval: the variable controls the number of times for migration.

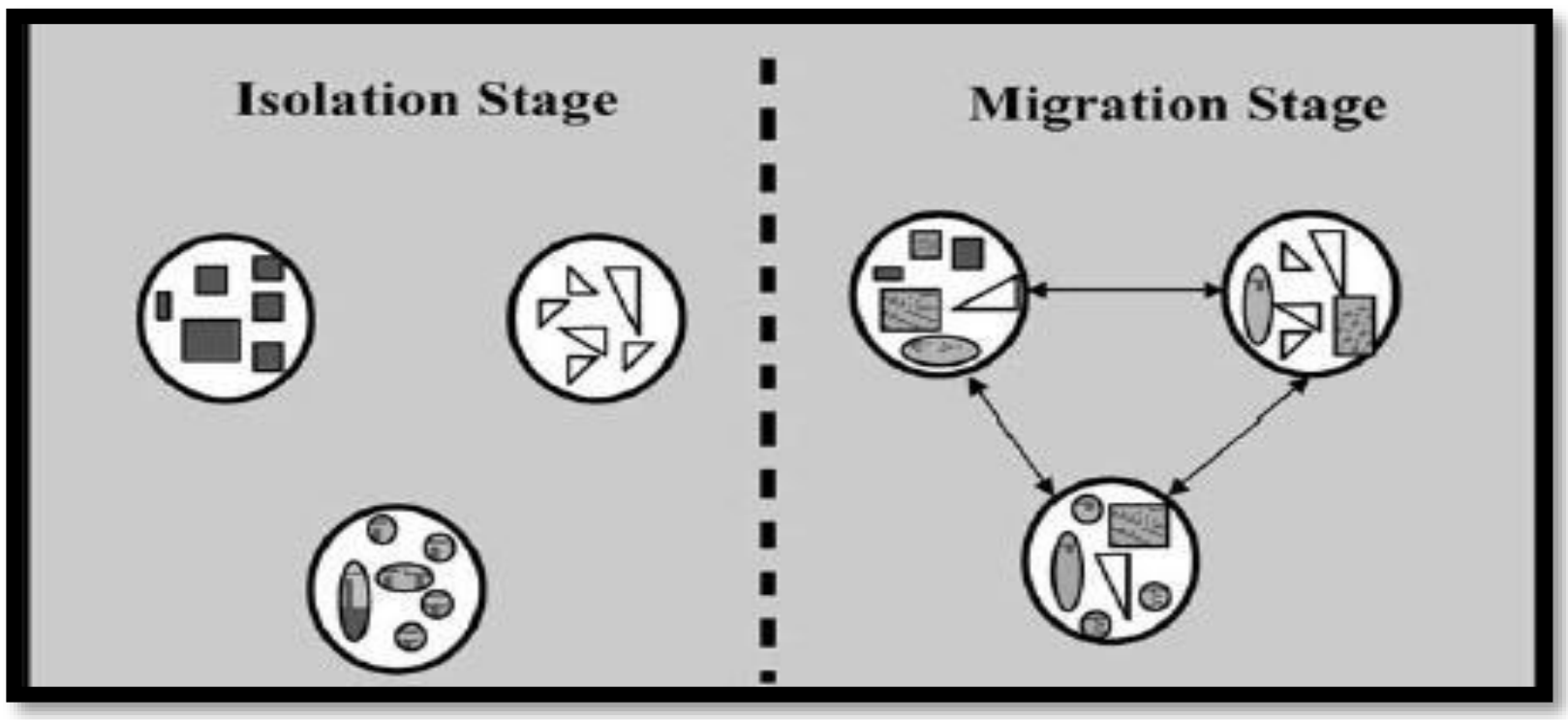

**Figure 2.11.** Symbolic representation of the migration stage used by MPGA **(Siarry, Pétrowski, & Bessaou, 2002)**

It is possible to integrate a multi-population with what is known as the speciation, which combines chromosomes based on the genetic similarity. The similarity is measured by distance, such as the Euclidean distance. In this method, multiple peaks are produced, and therefore there is more than one solution to the problem rather than with the single peak.

**2. Nitching Method:**

This maintains the diversity of the population and allows GAs to explore more of the area of search, in order to specify multiple peaks. Nitching reduces the impact of some of the phenomena, such as genetic drift (Booker, 1987), and is also useful in finding a single solution (best one) to complex problems and multiple solutions for other problems (Mahfoud, 1995). Examples of such applications are multi- modal functions, multi-objective optimization and classification, etc.

**3. Primal dual GA (PDGA)** (Yang, 2003)**:**

This method deals with a couple of chromosomes which are called "primal-dual". Chromosomes registered in the population of the GA are called "primal", while other chromosomes that have the farthest distance (in genotype) in a given distance space for the primal chromosomes are called "dual chromosomes", while the conversion function from primal to dual using a distance measure (e.g. Hamming distance) is called "primal-dual mapping". The algorithm chooses chromosomes that have less fitness, to give them the opportunity of moving to their superior dual. This algorithm increases the efficiency of the search.

There are also many techniques that support diversity in the population, such as: Dual Population GA (DPGA) (Park & Ryu, 2006), Crowding, Injection Strategy, Diversity Control Oriented GA (DCGA), Elitist, Restricted Mating, Multi-Objective Evolutionary Algorithm (MOEA) and many others.

Gupta and Ghafir (Gupta & Ghafir, 2012) presented concise knowledge of a different set of methods that maintain the diversity of the population. This is essential for continued success in the evolutionary system.

### 2.2.1 Exploration and Exploitation

Any effective search algorithm trying to find the global peaks must use two techniques: Exploration and Exploitation. Effective algorithms are concerned with finding a balance between these two concepts (Vekaria & Clack, 1998).

**1. Exploration:** gaining access to new and promising areas in the search space. The goal of this technique is diversification in order to avoid falling into premature convergence. This is a global search.

**2. Exploitation**: using knowledge that has been gained from exploration to find new best points. This process uses the term “refining” (intensification) to investigate a specific region; in other words, this is a local search.

A study by Back (1994) showed that the selection method (selection pressure) controls the degree of exploitation and exploration together, where increasing or decreasing the selection pressure leads to exploitation or exploration respectively.

(Eiben & Schippers, 1998) discussed how to achieve these two concepts, and it was observed that the common opinion was that crossover and mutation operations are used for exploration, while selection is used for exploitation. From their point of view, this matter (to achieve exploration and exploitation) needs a deep understanding of evolutionary algorithms.

The previous two works of research and other research that has been mentioned in the research paper (Črepinšek, Liu, & Mernik, 2013), where the aim of this paper is to provide treatment for exploration and exploitation, discussed several issues, including: how to achieve exploration and exploitation in Evolutionary Algorithms (EAs)? How to control exploration and exploitation? and achieving exploration and exploitation balance through diversity.

### 2.2.2 Literature review (MPGA)

Branke et al, “A multi-population approach to dynamic optimization problems”. Used a new technique called “Self Organization Scout” (SOS) based on the concept of a Forking GA (FGA) to enhance the search in a dynamic landscape. In this scheme, the population is composed of parents and children, parents who explore the search space and children who locate previously detected optima. This population is capable of simultaneously tracking many different promising peaks of the search space. As a result, this research showed that it is suitable for dynamic optimization (Branke, Kaußler, Smidt, & Schmeck, 2000).

Park and Ryu, “A dual population genetic algorithm with evolving diversity”. Proposed a new multi-population evolutionary algorithm termed a dual population GA (DPGA-ED).This is an updated version of the DPGA. The algorithm was inspired by two mechanisms –complementary and dominance –and this approach uses the main and reserve population, each one having an evolutionary purpose. The main population works as usual in a normal GA, and the reserve population keeps the individuals that are different from those in the main population, and exchanges information between the two parties to maintain diversity. Best children resulting from the exchange are sent to the main population, and the rest are sent to the reserve population. This study presented a new mechanism for information exchange between the populations and new parent selection methods (see Figure (2.12)).This technique is based on the selection of four parents, two from the main population and two from the reserve population. The choice of parents in the main population is done by the objective function, while the choice of parents in the reserve population is done by specific fitness function; six of the children arise through one crossbreeding process and two from the inbreeding process. Two of the children that were obtained from the process of inbreeding and two who were obtained from the process of crossbreeding (processes of parents in the main population), become candidates for the following generation in the main population. After assessing the children using fitness function on the main population, the best two children go to the main population. These steps also take place in the reserve population. The experimental results from various multi-modal optimization problems showed that the proposed algorithm is better than any other algorithms (Park & Ryu, 2007).

Hong et al, “Dynamically adjusting migration rates for multi-population genetic algorithms”. Presented the development of variables for migration (especially the migration rate) in the multi-population GA and their impact on the quality of the solutions, which use the methods of adjustment of the migration rate, by calculating the difference in fitness between two migration intervals (but neighbouring), where with individuals in the adjacent sub-population, if they have a good effect on the current sub-population, the rate of migration (from the neighbour) increases, and on the contrary, if they have a negative impact the rate of migration decreases (Hong, Lin, Liu, & Lin, 2007).

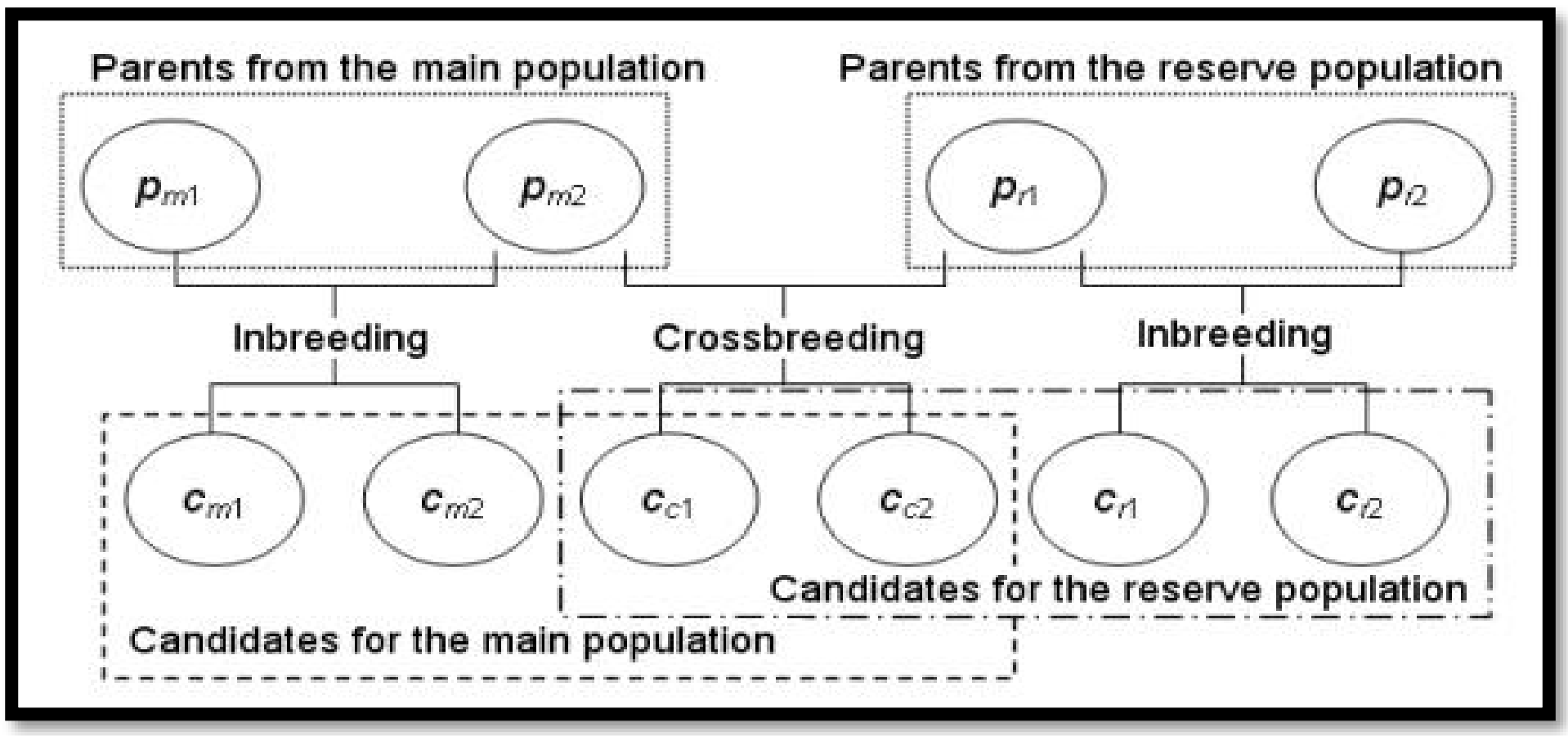

**Figure 2.12.**Offspring generated by inbreeding and crossbreeding, and the candidate for each population of the next generation **(Park & Ryu, 2007)**.

## 2.3 Parallel Genetic Algorithm (PGA)

Sequential GAs have proven their effectiveness in many areas and applications but there are many problems in finding solutions. It may take days or perhaps weeks, in addition to the possibility of the occurrence of a sequential GA in local optima (Nowostawski & Poli, 1999). Our need for diversity through, for example, an increase in the size of the population or by increasing the number of the population (the curse of dimensionality), will inevitably lead to deterioration of the performance of the GA (Umbarkar & Joshi, 2013); therefore the parallel GA emerged to address the problem of computational speed, and this has contributed to solving most of the problems.

The fundamental principle of parallel programs is to divide tasks into parts, and, by using a multiprocessor, parts are solved at the same time (Cantú-Paz, 1998). According to (Hart, Baden, Belew, & Kohn, 1996) the justification behind the use of the PGA is where this algorithm supports the following:

1. Minimizes the time needed to solve problems and evaluation costs.
2. For solutions, contributes to improving the quality.
3. The possibility of using a large size of population, especially in some difficult problems (to support effectiveness).

Some of the parallel methods work on a single population, and others work on a multi-population, therefore there are two models used by the parallel GA (Talbi & Bessiere, 1991):

1. Standard parallel model: single population.
2. Decomposition model: segmentation of the population to sub-population (which is the natural way for parallelization) and migrations among them.

The parallel GA is divided into three main parts (Figure (2.13) (Golub & Budin, 2000)):

1. **Master-slave GA model** (Figure (2.14)): Also called global (GPGA) or distributed fitness function, in this type the population is in the master, and the slaves perform the process of assessing individuals (fitness evaluation), apply mutations, and sometimes the process of crossover. When individuals are evaluated (the process is done in parallel) there is no need for contact, but there is contact when the slave receives individuals or upon completion of the evaluation returns the values of the fitness.

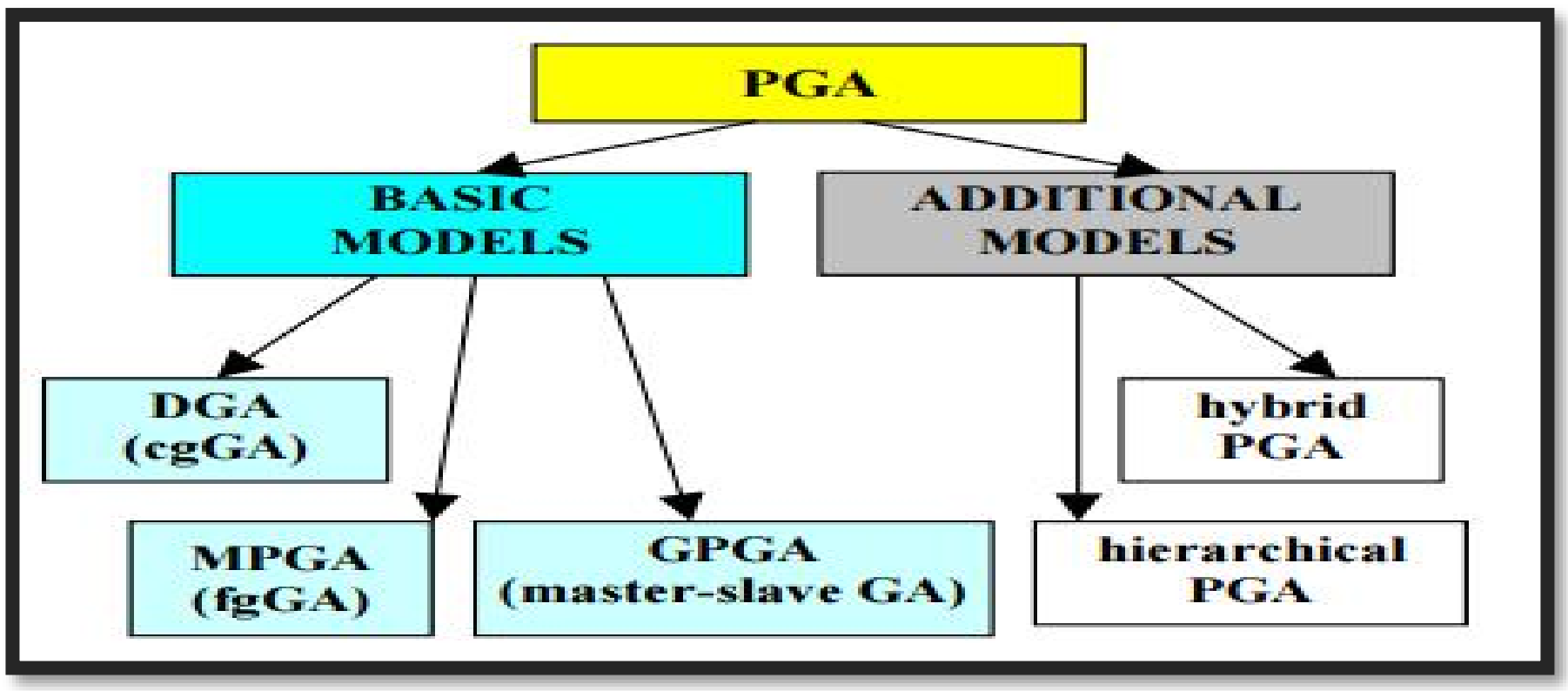

**Figure 2.13.** PGA models **(Golub & Budin, 2000)**

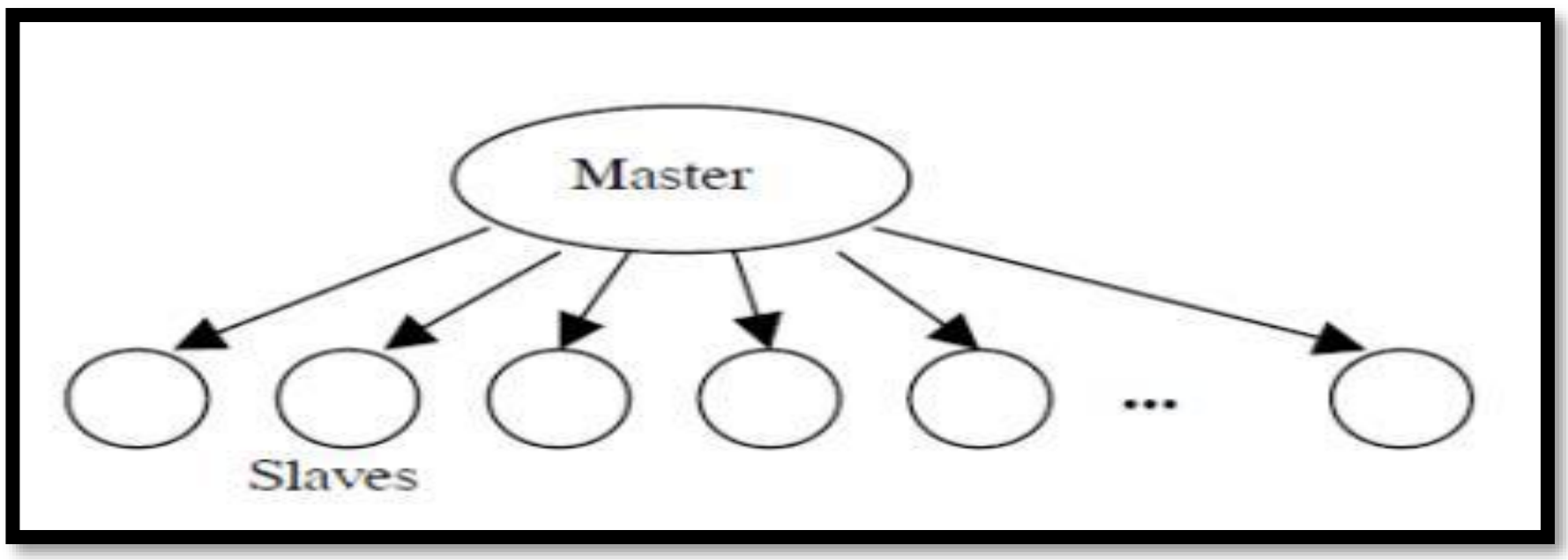

**Figure 2.14.**Master-Slave GA **(Al-Angari & ALAbdullatif, 2012)**

2. **Coarse grained GA:** This has several names, such as: distributed (DGA) or multiple-demes GA, multiple-population, island parallel GA. The idea is to divide the population into demes (small number of sub-populations), where each deme works in isolation. Here, migration between sub-populations induces the proliferation of good individuals, thus improving the solution. There are many topologies in migration, including (Figure (2.15)): ring migration, neighbourhood migration, injection migration. Many variables can control the path and efficiency of the search, such as: the migration rate, migration interval, the size of the sub-populations and the number of sub-populations.

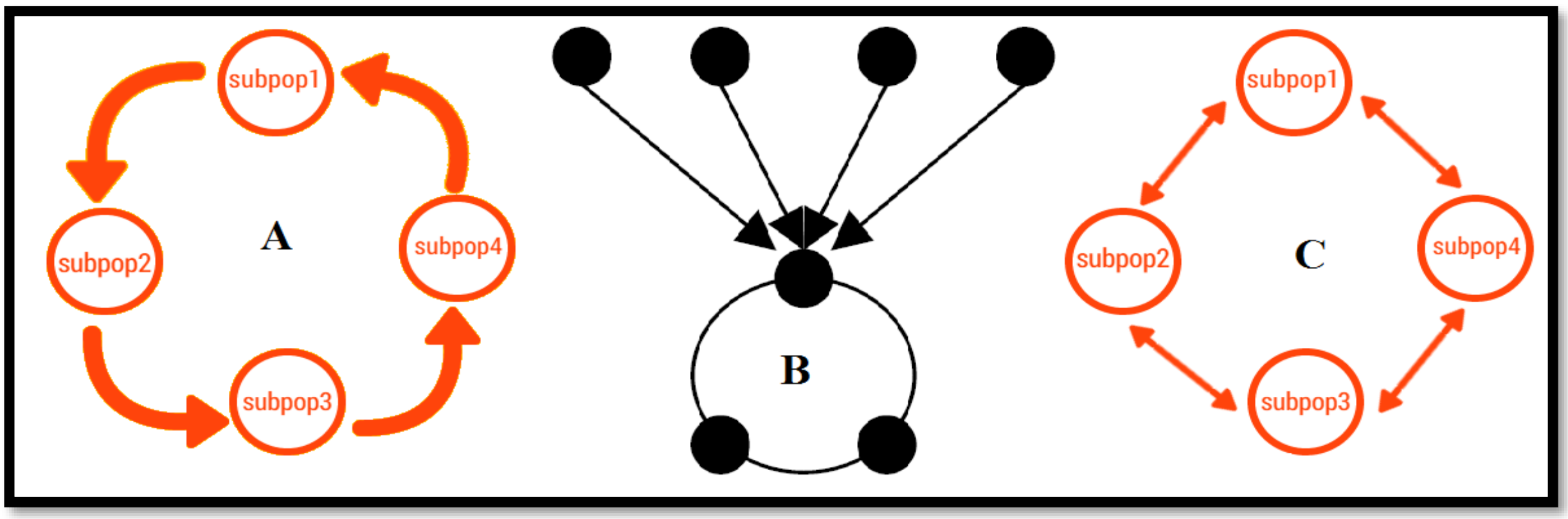

**Figure 2.15.** Migration topology A-Ring migration B-Injection Island **(Golub & Budin, 2000)** C-Neighbourhood topology

One obstacle to this algorithm is the possibility that new individuals result as ineffective and this means that the resulting genetic material is incompatible. The reason for incompatibility in general is that two sub-populations with good individuals when combined produce bad individuals (Baluja, 1993).

3. **Fine grained GA (FgGA):** also called Massively PGA (MPGA) or Diffusion GA (Figure (2.16)).

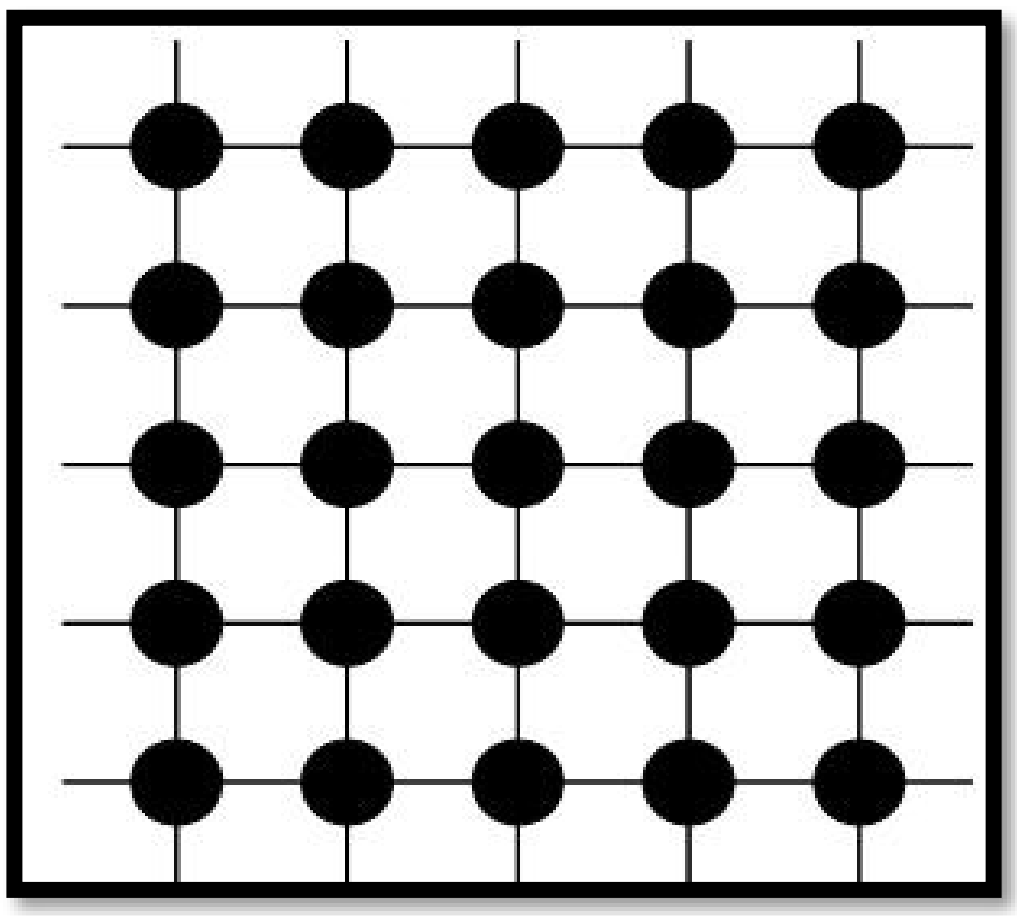

**Figure 2.16**. FgGA **(Cantú-Paz, 1998)**

FgGA has one population, where the individual can mate with its neighbours only, and overlaps between neighbourhoods allow good individual spreads in the entire population. Neighbourhood shape and neighbourhood size are considered as variables of the function to determine the amount of overlap (Sarma & De Jong, 1996). Different neighbourhood shapes and sizes are shown in Figure (2.17).

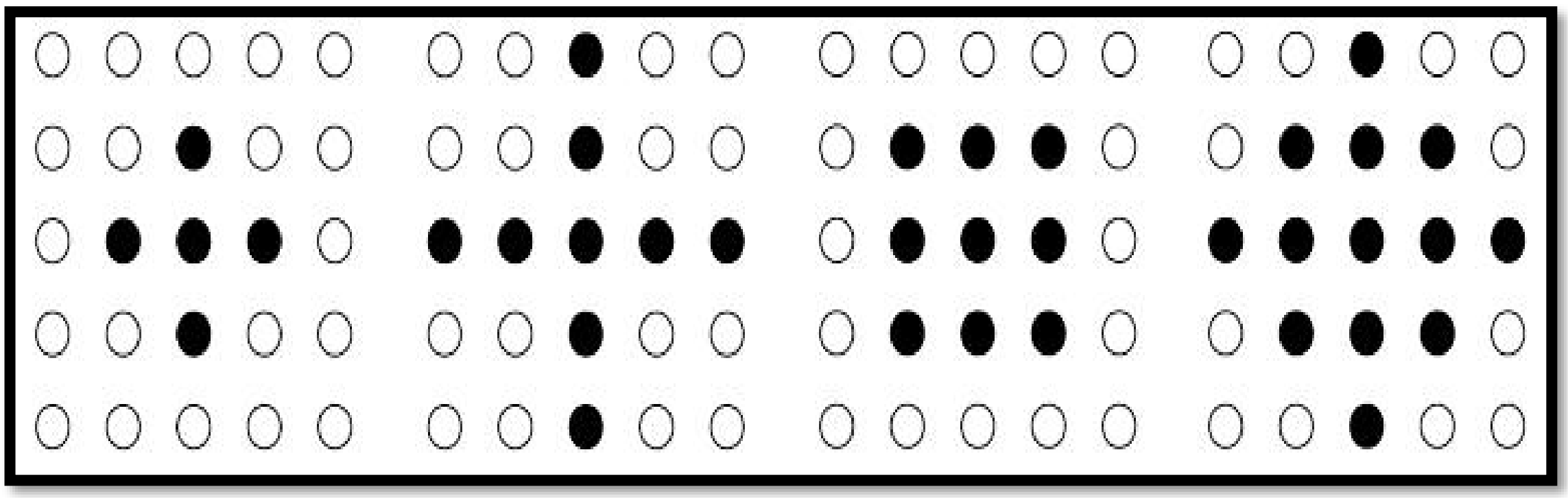

**Figure 2.17.** Different sizes [5, 9, 9, 13] and shapes of neighbourhoods in FgGA **(Sarma & De Jong, 1996)**

4. **Hierarchical PGA** (Figure (2.18)) is another type which is created through the integration of the three types with each other. This type contributes to minimization of the execution time (Cantu-Paz & Goldberg, 2000).
5. **Hybrid PGA** is another type of parallel GA which is the integration of the parallel GA with the optimization method.

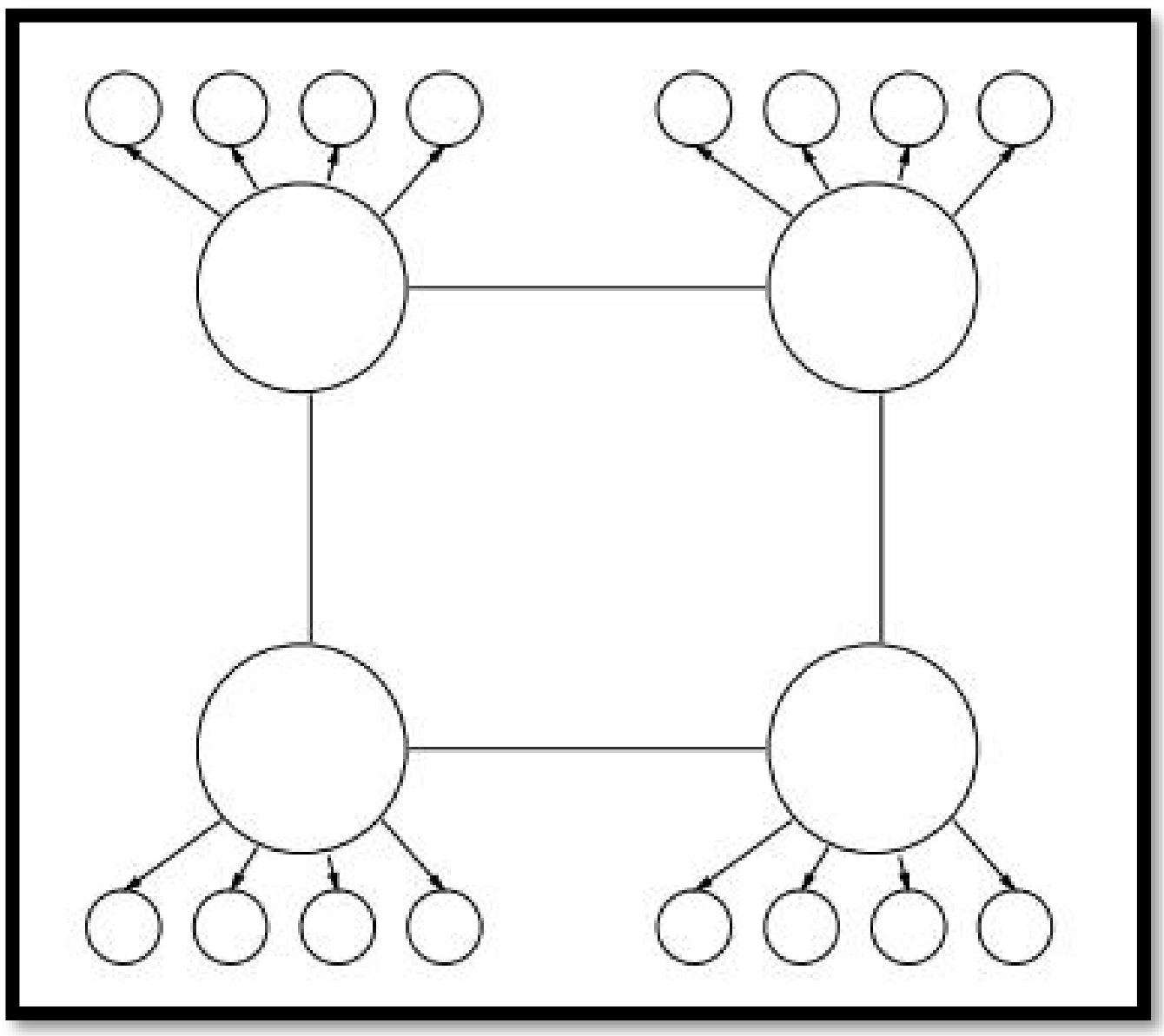

**Figure 2.18.** A schematic of a hierarchical PGA. At the upper level this hybrid is a multi deme PGA where each node is a master slave GA **(Cantu-Paz & Goldberg, 2000)**.

Parallel algorithms have many applications, such as: graph partitioning problems, scheduling problems, financial balancing problems, optimization of VLSI circuits, optimization in material engineering, optimization of server load or database queries and others. In (Konfršt, 2004) there is a brief overview of some research that mentions applications for PGA.

### 2.3.1 Literature review (PGA)

Pettey et al, “A parallel genetic algorithm”. Proposed a migration method, that copies the good individual in the sub-population to the rest of the neighbouring sub-populations. Migration takes place after every generation (Pettey, Leuze, & Grefenstette, 1987). In his book, Russell showed that inbreeding (mating among relatives) reduces the diversity of the population, while outbreeding (mating between individuals with non-relatives) increases diversity. He also said that inbreeding is a form of so-called positive assortative mating and outbreeding is a form of negative assorative mating (Russel, 1997).

Eldos, “A New Migration Model For Distributed Genetic Algorithms”. Proposed a new migration, where individuals are not allowed to migrate unless they exceed certain conditions. The condition exists in the first deme (source), and the other condition in the second deme (destination), where each sub-population (deme) has qualification criteria and the simplest criterion is the fitness of individuals (Eldos, 2006).

Ruciński et al, “On the impact of the migration topology on the island model”. Studied the impact of migration topology on the Parallel Global Optimization Algorithm, where several topologies were applied to more problems and different numbers of islands, and this showed that topology plays an important role in the quality of the solution and in the speed of convergence, where the impact appears evident in large networks (Ruciński, Izzo, & Biscani, 2010).

Chen and Li, “The Design and Analysis of an Improved Parallel Genetic Algorithm Based on Distributed System”. Proposed a new migration method which can be modified dynamically. They calculated the different degree of individuals in each deme (in the current generation) using a certain equation. Good individuals were stored in a buffer for each deme. The exchange of good individuals is based on certain values where they become a condition for migration. Good individuals move in the current deme to the buffer in the other deme, and then the best individuals are transmitted to the other deme and replaced by bad individuals in the deme (Chen & Li, 2012).

## 2.4 Travelling Salesman Problem (TSP)

TSP is considered one of the combinatorial optimization problems (Laporte, 1992), that is easy to describe but difficult to solve, and it is classified as among the problems that are not solved in polynomial time; in other words, it belongs to the NP-hard problem (Potvin, 1996). In 1832 a veteran travelling salesman used the term “travelling salesman” in the manual for the travelling salesman, which was published in Germany (Osorio, Pérez, & Pérez, 2002).

The TSP is to find the shortest path (tour) through a set of nodes (starting from a given node N and finishing at the same node), so that each node is visited only once.

The TSP classified into three types (Rao & Hegde, 2015):

1. Symmetric travelling salesman problem (STSP): The cost (distance) between any two cities in both directions is the same (undirected graph), i.e. the distance from *city1* to *city2* is the same as the distance from $city_2$ to $city_1$.
2. Asymmetric travelling salesman problem (ATSP):The cost (distance) between any two cities in both directions is not the same (directed graph), i.e. the distance from $city_1$ to $city_2$ is not the same as the distance from $city_2$ to $city_1$.
3. Multi travelling salesman problem (MTSP): is a TSP, but for more than one salesman.

Many techniques can be used to solve the TSP, such as (Reisleben & Merz, 1996): GA (Soni & Kumar, 2014) (Larrañaga, Kuijpers, Murga, Inza, & Dizdarevic, 1999), Simulated Annealing (Malek, Guruswamy, Pandya, & Owens, 1989), Hill Climbing, Ant Colony (Dorigo & Gambardella, 1997), Tabu Search (Gendreau, Hertz, & Laporte, 1994), Particle Swarm (Shi, Liang, Lee, Lu, & Wang, 2007), Elastic Nets (Durbin, Szeliski, & Yuille, 1989), Neural Networks (Aarts & Stehouwer, 1993), Nearest Neighbour and Minimum Spanning Tree algorithms (Karkory & Abudalmola, 2013), etc.

TSPs are used in various applications, including (Rai, Madan, & Anand, 2014): Computer wiring, Crystallography, Dartboard design, Job sequencing, Hole punching, Wallpaper cutting, Overhauling gas turbine, etc.

### 2.4.1 Solving TSP using GA

Over the years, a great deal of research has occurred in applying GAs to solve the TSP. Therefore, there are several representations of the TSP using a GA and each one of them has its own operators (Mohebifar, 2006).These representations include:

1. Binary representation: where each city is represented by bit string; for example, in 7 cities TSP, each city is represented using 4-bit strings: a tour 1→3→7→6→5→2→4 is represented: (00010011 0111 0110 0101 0010 0100).

   The operators used in this representation are classical mutation and classical crossover, but the problem lies in the creation of illegitimate cities, and here we need to repair algorithms.
2. Path representation: where there is a natural representation for the tour (Abdoun, Abouchabaka, & Tajani, 2012), for example: a tour 1→3→7→6→5→2→4 is represented: (1 3 7 6 5 2 4).

The classical operator is unsuitable because it certainly will produce the illegal tour. Therefore the crossover operators used in path representation are: partially mapped crossover (PMX), cycle crossover (CX), order crossover (OX), position-based crossover (POS), and other types. The mutation used in this representation includes: displacement mutation (DM), exchange mutation (EM), insertion mutation (ISM) and some other types.

3. Adjacency representation: If there is an edge from city *i* to city *j*, city *j* occupies the position of city *i*. For example, suppose that the parent: 38526417 represents the tour 13564287, so that city 3 is listed in position 1 because there is an edge from city 1 to city 3, and city 8 is listed in position 2 because there is an edge from city 2 to city 8, and so on (Potvin, 1996).

   The crossover operators used here are: alternating edges crossover, subtour chunks crossover, heuristic crossover.

4. Ordinal representation: In this representation, the tour is encoded as a list of cities. The element *i* in the list is the number in the range from 1 to (n-i+1) (Abdoun, Abouchabaka, & Tajani, 2012). In this representation there is a reference tour to serve the encoding; for example, suppose that the reference tour is: 1 2 3 4 5 6 7 8, and the current tour to be encoded is: 1 2 5 6 4 3 8 7; the ordinal representation in this tour is shown in Figure (2.19) (Potvin, 1996). The crossover operator that can be used in this representation is one-point crossover.

| Current Tour | Canonic Tour | Ordinal Representation |
|---|---|---|
| 1 2 5 6 4 3 8 7 | 1 2 3 4 5 6 7 8 | 1 |
| 1 2 5 6 4 3 8 7 | 2 3 4 5 6 7 8 | 1 1 |
| 1 2 5 6 4 3 8 7 | 3 4 5 6 7 8 | 1 1 3 |
| 1 2 5 6 4 3 8 7 | 3 4 6 7 8 | 1 1 3 3 |
| 1 2 5 6 4 3 8 7 | 3 4 7 8 | 1 1 3 3 2 |
| 1 2 5 6 4 3 8 7 | 3 7 8 | 1 1 3 3 2 1 |
| 1 2 5 6 4 3 8 7 | 7 8 | 1 1 3 3 2 1 2 |
| 1 2 5 6 4 3 8 7 | 7 | 1 1 3 3 2 1 2 1 |

**Figure 2.19.** Ordinal representation example **(Potvin, 1996)**

5. Matrix representation: There are three attempts to work on the matrix representation (Homaifar et al. (Homaifar, Guan, & Liepins, 1993), Seniw (Michalewicz, 2013), Fox and McMahon (Fox & McMahon, 1991)). In the proposed matrix by Homaifar et al., if there is an edge from $city_i$ to $city_j$, we put 1 in the $m_{ij}$ of the binary matrix M. For example, parent1 (p1) = (1, 3, 5, 4, 2) and parent2 (p2) = (2, 4, 3, 1, 5).The matrix representation is shown in Figure (2.20). The crossover operator used in this type is classical crossover.

P1

| | | | | |
|---|---|---|---|---|
| 0 | 0 | 1 | 0 | 0 |
| 1 | 0 | 0 | 0 | 0 |
| 0 | 0 | 0 | 0 | 1 |
| 0 | 1 | 0 | 0 | 0 |
| 0 | 0 | 0 | 1 | 0 |

P2

| | | | | |
|---|---|---|---|---|
| 0 | 0 | 0 | 0 | 1 |
| 0 | 0 | 0 | 1 | 0 |
| 1 | 0 | 0 | 0 | 0 |
| 0 | 0 | 1 | 0 | 0 |
| 0 | 1 | 0 | 0 | 0 |

**Figure 2.20.** Example of matrix representation **(Osorio, Pérez, & Pérez, 2002)**

TSP is considered as a minimization problem, therefore the fitness function can be expressed by calculating the cost of the tour. The cost is calculated by the Euclidean distance (ED) between the two cities as in the equation below:

$$ED = \sqrt{(x1 - x2)^2 + (y1 - y2)^2} \quad (3)$$

where (x1, y1) and (x2, y2) are coordinates of city$_i$ and city $_j$ respectively.

## 2.5 Summary

Today, GA is considered an important tool for solving many problems, and it is derived from genetics and natural selection. The important stages in this algorithm, which are made from generation to generation, include: the encoding stage, assessing the fitness, crossover, mutation and selection. The encoding phase or representation is the first stage, where every problem has a special encoding, the population is randomly generated, the fitness of each individual is calculated, and then parents are selected, crossover and mutation are applied to produce a new generation, the fitness of the new individual is calculated, and this process is repeated until a certain standard is achieved to finish.

There are also important parameters of the GA where these parameters have an influence on the search path and the quality of the solution and must be carefully selected. These variables are: the size of the population, the crossover rate, the mutation rate.

One problem associated with the GA is the loss of diversity in the population and the consequent possibility of falling into the local optima. This can lead to so-called “premature convergence”. This problem is related to the factors that must be taken into account when starting the algorithm (initial population), such as: diversity, selection, problem difficulty. The primary aim is to eliminate the problem, obtain the best solutions and find multiple peaks (highest peak is global optimum and lower peaks are local optima). Research has tended to provide several techniques to maintain diversity, and thus the GA can deal with uni-modal and multi-modal optimization functions. These techniques include: MPGA, nitching, PDGA, injection, dual population and more. These applications divide the population into sub-populations, and apply migrations between these branches (sub-population) to encourage the proliferation of good individuals, and therefore obtain the best solutions.

Migration between the populations is necessary, as this includes very important variables, namely the migration rate and migration interval, which control the amount of diversity in the population.

Applying GA in many cases (e.g. business) needs a large size of population, and thus needs a long time to calculate fitness. To speed up the GA, and reduce the computational time, the PGA has emerged with three basic types: master-slave PGA, coarse grained algorithm and fine grained algorithm. The PGA is used in many areas: finance, scheduling problems, graph theory and many others.

# Chapter 3
# Proposed Work for Crossover Operator

## 3.1 Design and Methodology

Crossover operators have a role in the balance between exploitation and exploration, which will allow the extraction of characteristics from both parents, and hopefully the resulting children will possess good characteristics from the parents (Gallard & Esquivel, 2001).

The search for the best solution (in GAs) depends mainly on the creation of new individuals (chromosomes) from the old ones. The process of crossover ensures the exchange of genetic material between parents and thus creates chromosomes that are more likely to be better than the parents. There is a large number of crossover methods in the literature, and so the question is: what is the best method to use?

To answer this question, and in an attempt to provide diversity in the population, we have proposed and implemented two strategies and three types of crossovers, to be compared with two of the well-known types, namely: Modified crossover and PMX crossover.

## 3.2 Crossover operator

We shall describe the three operators, in addition to the two strategies in detail as follows:

### 3.2.1 Cut on worst gene crossover (COWGC)

This method aims to exchange genes between parents by cutting the chromosome at the maximum point that increases the cost. This point (the worst gene) is chosen in both parents depending on the definition of the worst gene for each problem; the worst gene is the point that contributes the maximum to increase the cost of the fitness function of a specific chromosome. For example, the worst gene in the TSP problem is the city with the maximum distance from its left neighbour, while the worst gene in the Knapsack problem is the point with the lowest value to weight ratio, and so on.

Our algorithm needs to search along the chromosome to find this gene in both parents. The two worst genes are compared to get the worst of both; the index of this point is considered as a cut point in the parent that has the worst gene. The genes after this cut point of the two parents are swapped as in the "Modified Crossover" (Davis, 1985). Figure (3.1) shows an example of (COWGC). The sign ">" means the worst, i.e. "greater than" if the problem is a minimization problem, and "less than" otherwise.

The cut point (CP) can be calculated for the minimization problem using:

$$CP = \underset{1 \le i < n}{\operatorname{argmax}}(Distance(C[i], C[i+1]) \quad (4)$$

and for the maximization problem:

$$CP = \underset{1 \le i < n}{\operatorname{argmin}}(Distance(C[i], C[i+1]) \quad (5)$$

where $C$ represents the chromosome, $i$ is the index of a gene within a chromosome, and the distance function for the TSP can be calculated using either Euclidian distance or the distances table between cities. The previous equations are used for both parents, and the cut point of the parent that exhibits the maximum distance is used for the crossover operation.

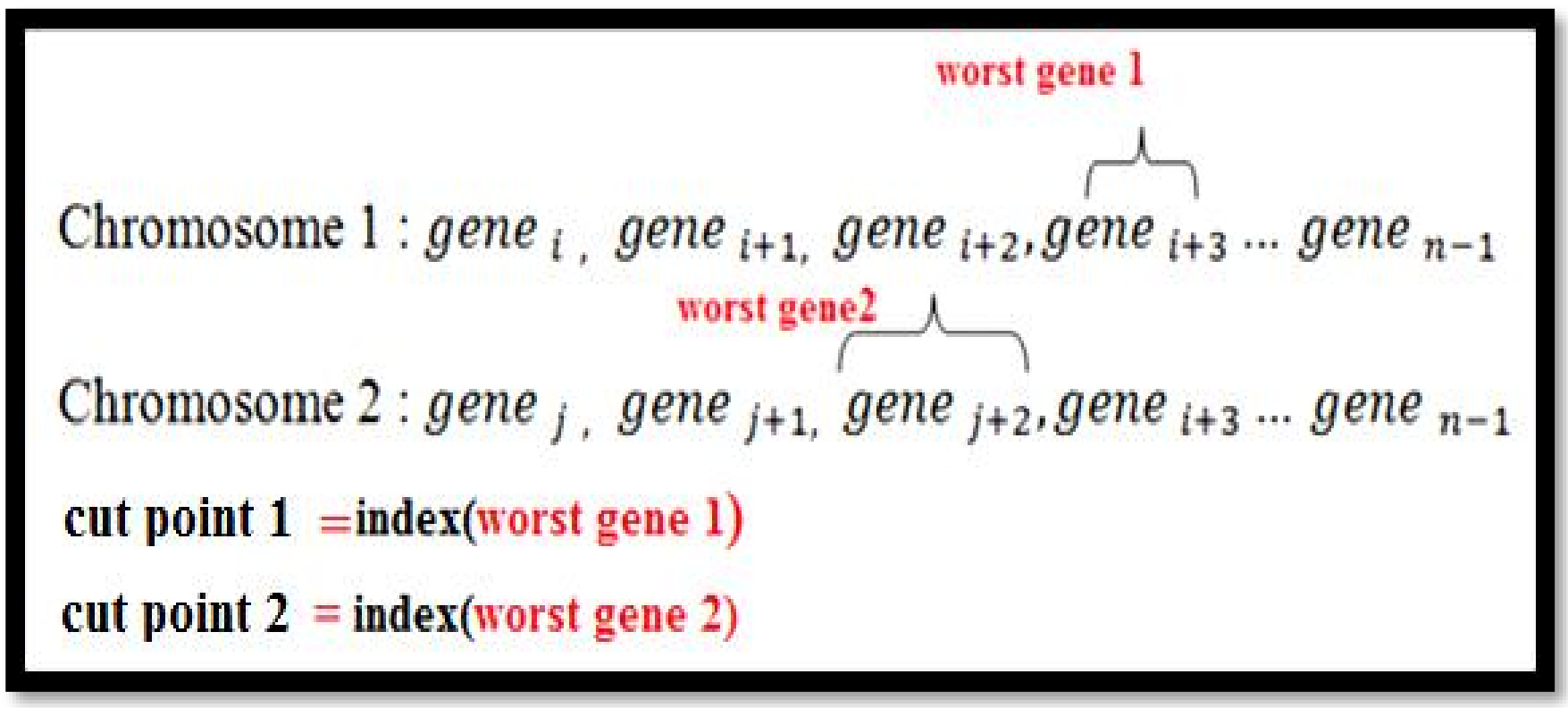

**Figure 3.1**. Example of COWGC

**Example 1. (TSP problem)**

If we assume that we represent the TSP problem using path representation and this is natural representation of a tour, each tour is represented by a list of n cities such as: the tour 1-5-4-8-6-7-9 is represented as (1548679).

Figure (3.2) shows the distances between cities. Suppose we choose randomly two parents from the mating pool:

Parent 1: 1 3 8 7 5 6 2 9 4

Parent 2: 1 5 9 8 4 3 7 6 2

| City | 1 | 2 | 3 | 4 | 5 | 6 | 7 | 8 | 9 |
|---|---|---|---|---|---|---|---|---|---|
| 1 | 0 | 2 | 8 | 5 | 20 | 6 | 25 | 30 | 4 |
| 2 | | 0 | 5 | 3 | 15 | 8 | 52 | 21 | 12 |
| 3 | | | 0 | 27 | 6 | 10 | 20 | 14 | 7 |
| 4 | | | | 0 | 8 | 4 | 17 | 60 | 2 |
| 5 | | | | | 0 | 22 | 6 | 8 | 5 |
| 6 | | | | | | 0 | 15 | 6 | 8 |
| 7 | | | | | | | 0 | 10 | 9 |
| 8 | | | | | | | | 0 | 30 |
| 9 | | | | | | | | | 0 |

**Figure 3.2.** TSP example distances between cities

To apply COWGC:

a. Step1: find the worst gene in the first parent, which is 6, because the distance from 5 to 6 is the maximum and equal to 22 (distance 1), and the worst gene in the second parent is 4, because the distance from 8 to 4 is the maximum and equal to 60 (distance 2).
b. Step 2: using equation (1) the cut point of parent 1 is the index (6) and the cut point of parent 2 is index (4).
c. Step 3: If (distance1) > (distance2), then
d. Apply the Modified crossover in both parents at index (6).
   Else apply the Modified crossover in both parents at index (4) (see Figure (3.3)).

**parent 1 : 1 3 8 7 | 5 6 2 9 4**

**parent 2 : 1 5 9 8 | 4 3 7 6 2**

**offspring1 : 1 5 9 8 3 7 6 2 4**

**offspring2 : 1 3 8 7 5 9 4 6 2**

**Figure 3.3.** Two offspring output using COWGC

### 3.2.2 Cut On Worst L+R Crossover (COWLRGC)

This method is similar to the COWGC, the only difference being that the worst gene is found by calculating the distance between both its neighbours: the right and the left. The cut point can be calculated using:

$$CP = \underset{2 \le i < n-1}{\operatorname{argmax}}(Distance(C[i], C[i-1] + Distance(C[i], C[i+1]) \quad (6)$$

The worst gene is the one that is the sum of the distances with its left and right neighbours and is the maximum among all genes within a chromosome.

**Example 2. (TSP problem)**

Based on Example (1), suppose we choose randomly two parents from the mating pool:

Parent 1: 1 4 2 8 9 6 3 7 5

Parent 2: 1 9 5 7 8 2 3 4 6

To apply COWLRGC:

a. Step 1: using equation (3) calculate CP1 for the first parent and CP2 for the second parent. CP1 will be at city 8, because the total distance from city 8 to city 2 and from city 8 to city 9 is the maximum and is = 51 (distance 1). For the second parent, CP2 will be at city 3, because the total distance from city 3 to city 2 and from city 3 to city 4 is the maximum distance and is = 32 (distance 2).

b. Step 2: If distance 1 > distance 2, then apply Modified crossover for both parents based on CP1 (city 8), to create two offspring (see Figure (3.4)). Else apply Modified crossover for both parents based on CP2.

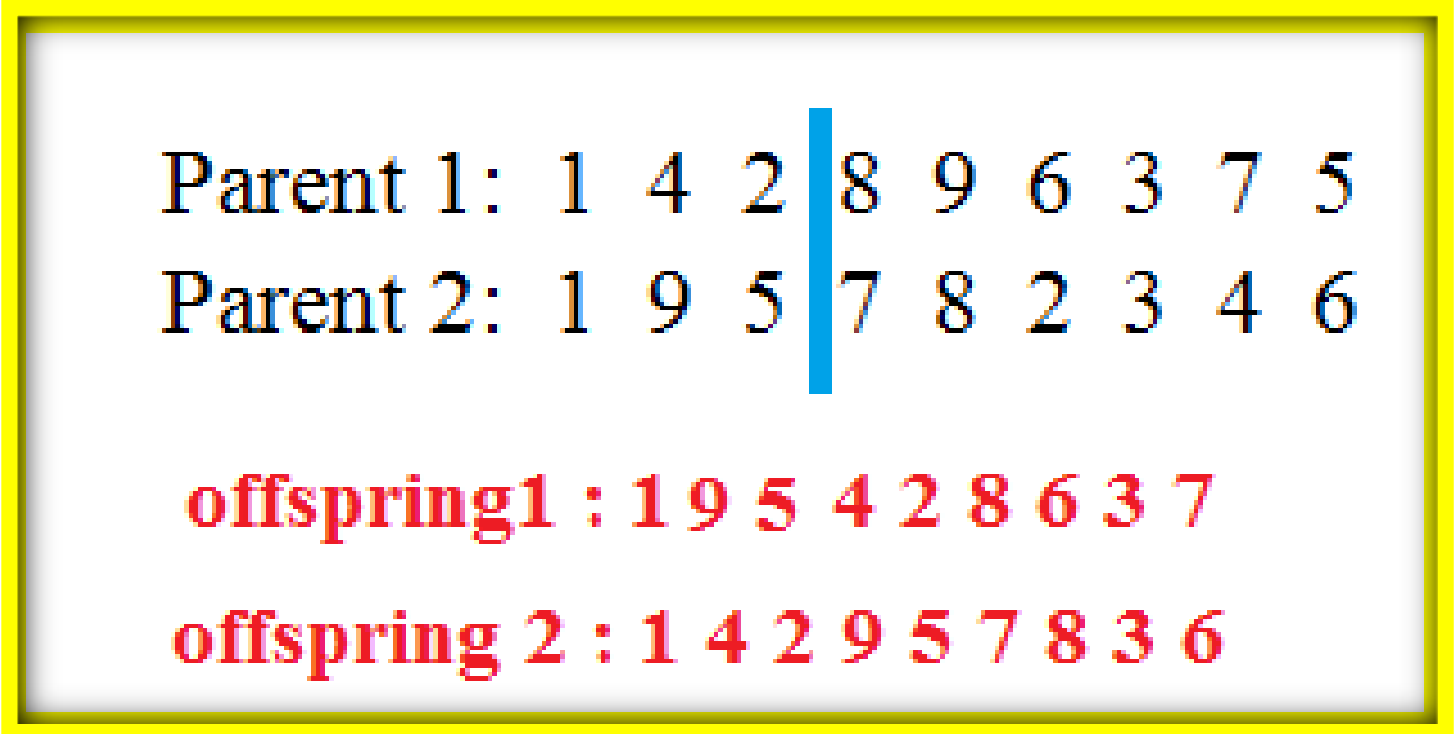

**Figure 3.4.** Two offspring output using COWLRGC

### 3.2.3 Collision Crossover

This type of crossover is inspired by the principle of the head-on elastic collision, when two objects are moving towards each other, with specific velocity and mass for each. If the collision happens, the direction and speed of both objects can be calculated using:

$$v'_1 = \frac{m_1 - m_2}{m_1 + m_2} v_1 + \frac{2m_2}{m_1 + m_2} v_2 \qquad (7)$$

$$v'_2 = \frac{2m_1}{m_1 + m_2} v_1 - \frac{m_1 - m_2}{m_1 + m_2} v_2 \qquad (8)$$

where $v_1$ and $v_2$ are the velocities of objects 1 and 2 respectively, m1 and m2 are the masses of objects 1 and 2 respectively, $v_1$' and $v_2$' are the new velocities after collision of objects 1 and 2 respectively. We assume both objects are moving in the opposite direction, so that one of the velocities should be negative (see Figure (3.5)).

Depending on the masses and velocities of the moving objects, the direction and the new velocities can be determined. There are several possibilities after collision, such as: if the 1st object was heavier and faster it will continue in the same direction and so the other object; if both objects are similar they might reflect to the opposite direction or become stationary.

In applying this physics to do crossover, we assume that each gene has its own mass, e.g. masses for Parent 1 = $\{m_{11}, m_{12} \ldots m_{1n}\}$, and masses for Parent 2 = $\{m_{21}, m_{22} \ldots m_{2n}\}$. Choosing which of these masses depends on the problem itself; for the TSP, we assumed that each city has a mass, which is its distance from its left and right neighbours. For the 01-Knapsack problem, the mass of a gene might be the ratio of its value to its weight.

To do crossover we assume that both chromosomes (parents) are moving towards each other (opposite direction, 180 degrees head-on elastic collision). The velocity of each parent is its total cost, thus each gene within a chromosome has a mass and a velocity.

When both parents collide, each gene (depending on its velocity and mass) will be either reflected, become stationary, or keep moving in the same direction. This can be known from the sign and value of the new velocity.

If the gene is reflected or becomes stationary (v' = 0) this means that the gene is “good”, i.e. it is a small distance from its neighbours, and therefore it should remain in its place to form child (1). Other genes which carried on moving in the same direction are removed from their places in the new child (1), leaving gaps that need to be filled from parent 2. The same procedure is applied to child (2).

Equations 4 and 5 decide which remains and which leaves, and the gap places are filled by the other parent, ensuring that no gene is repeated. To foster randomness and diversity we opted for changing the velocity of the moving chromosomes using a random number from 1 to the cost of the chromosome, rather than fixing it to be the chromosome’s cost.

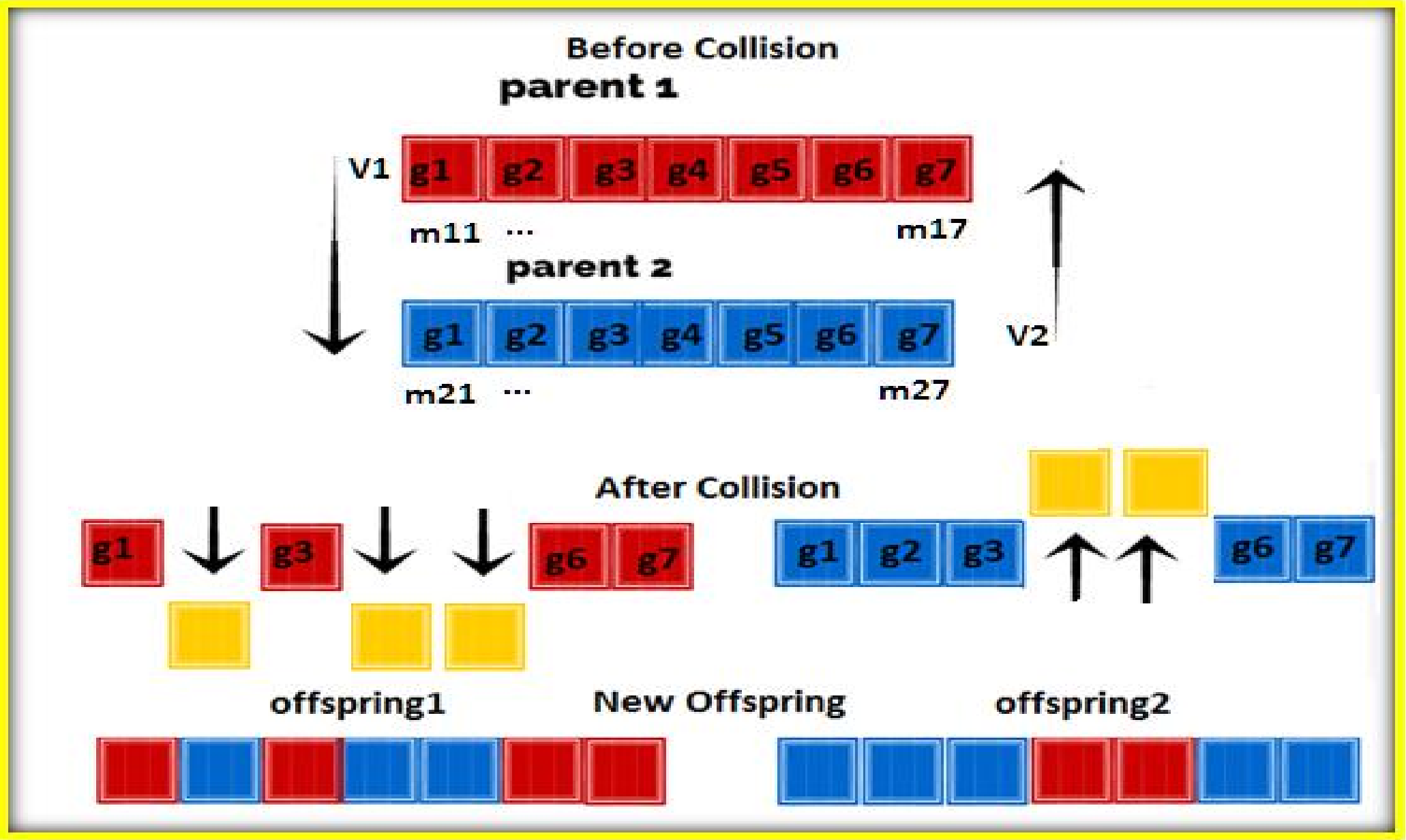

**Figure 3.5**. Collision crossover

### 3.2.4 Multi Crossover Operator Algorithms

A traditional GA commonly uses one crossover operator. We propose using more than one crossover operation, anticipating that different operators will produce different patterns in the offspring and provide some sort of diversity in the population, so as to improve the overall performance of the GA.

To do this we opted for two selection approaches: the best crossover, and a randomly chosen crossover.

#### 3.2.4.1 Select the best crossover (SBC)

This algorithm applies multiple crossover operators at the same time on the same parents, and considers the best two children to be added to the population, to prevent duplication; only the best and not found in the population is added.

In this work, we applied all the aforementioned crossovers (COWGC, COWLRGC and Collision Crossovers) to the two randomly chosen parents. The best two children that do not already exist in the population are added, though not necessarily from the same operation. This anticipates that such a process encourages diversity in the population, and thus avoids falling into local optima.

#### 3.2.4.2 Select any crossover (SAC)

This algorithm is similar to SBC, the difference lying in applying only one crossover operator each time; the selection strategy is random.

Randomly choose one of the aforementioned crossovers (COWGC, COWLRGC and Collision Crossovers) in the two randomly chosen

parents, and add the two new children to the population. We reckon that in each generation or so, the algorithm chooses a different type of crossover. This means that different types of children will result, and this is what we are aiming for, thus increasing diversification in the population, and attempting to enhance the performance of the GA.

### 3.3 Experimental Results and Discussion

To evaluate the proposed methods, we conducted two experiments on different TSP problems. The aim of the first experiment was to examine the convergence to the minimum value of each method separately. The second experiment was designed to examine the efficiency of the SBC and SAC algorithms and compare their performance with the Modified crossover and PMX crossover using real data.

#### 3.3.1 Experiments Set 1

The aim of these experiments was to measure the effectiveness of the proposed crossover operators (COWGC, COWLRGC and Collision Crossovers). The results of the experiments were compared to the well-known crossover (the one-point modified crossover (Davis, 1985)) and PMX crossover (Goldberg & Lingle, 1985), in addition to measuring the performance of the GA when using either SAC or SBC.

These crossover operators were tested using three test data: the first was random cities, where the coordinates of the cities were chosen randomly, and the second and third test data are real data – “bier127“ and “a280” taken from TSPLIB (Reinelt & Gerhard, 1996), each of them consisting of 100, 127, and 280 cities respectively. Mutation used in this experiment is Exchange mutation (Banzhaf, 1990).

The GA parameters that were selected in the first test included the following: population size: 100, the probability of crossover: (83%), mutation probability: (2%), and the maximum generation was 2000.
In the second test we used the same parameters, applied to the same problems, but the crossover probability was increased to (92%). The results of the second test are shown in Figures (3.9, 3.10, and 3.11).

Results from the first test indicate that the best performance is recorded by the SBC, followed by the Collision crossover and SAC, in most cases the performance of the Collision crossover was better than that of the SAC, because it showed good convergence to a minimum value. The Collision crossover showed a faster convergence rate than other crossover operators, followed by the modified crossover. The performance of each method is shown in Figures (3.6), (3.7) and (3.8).

Increasing the crossover ratio (the second test) does not enhance the performance of all the method significantly. A closer look at Figures (3.6-3.11) reveals that the SBC algorithm outperformed all other methods in the speed of convergence. In addition, the Collision crossover and the modified crossover showed rapid convergence to a near optimal solution as compared to other methods.

Despite the SBC outperforming the Collision crossover and SAC, the Collision crossover and SAC is still better than the SBC in terms of the time consumed, because SBC tries all the available crossovers and selects the best, which consumes more time. While the SAC selects any one randomly, and the Collision crossover is only one crossover operator by definition. Moreover, the differences between the results of the three methods are not significant. In addition, if the number of crossover operators in the SBC and SAC increased, both algorithms might become more efficient, but this is at the expense of the SBC's consumed time.

Some crossovers showed better performance than others, and this does not mean that the rest of the crossovers have proved their failure, as they can be effective when used by SBM and SAM, where they urge diversity through the different patterns of individuals and hence increases the efficiency of both algorithms and help escaping local minima.

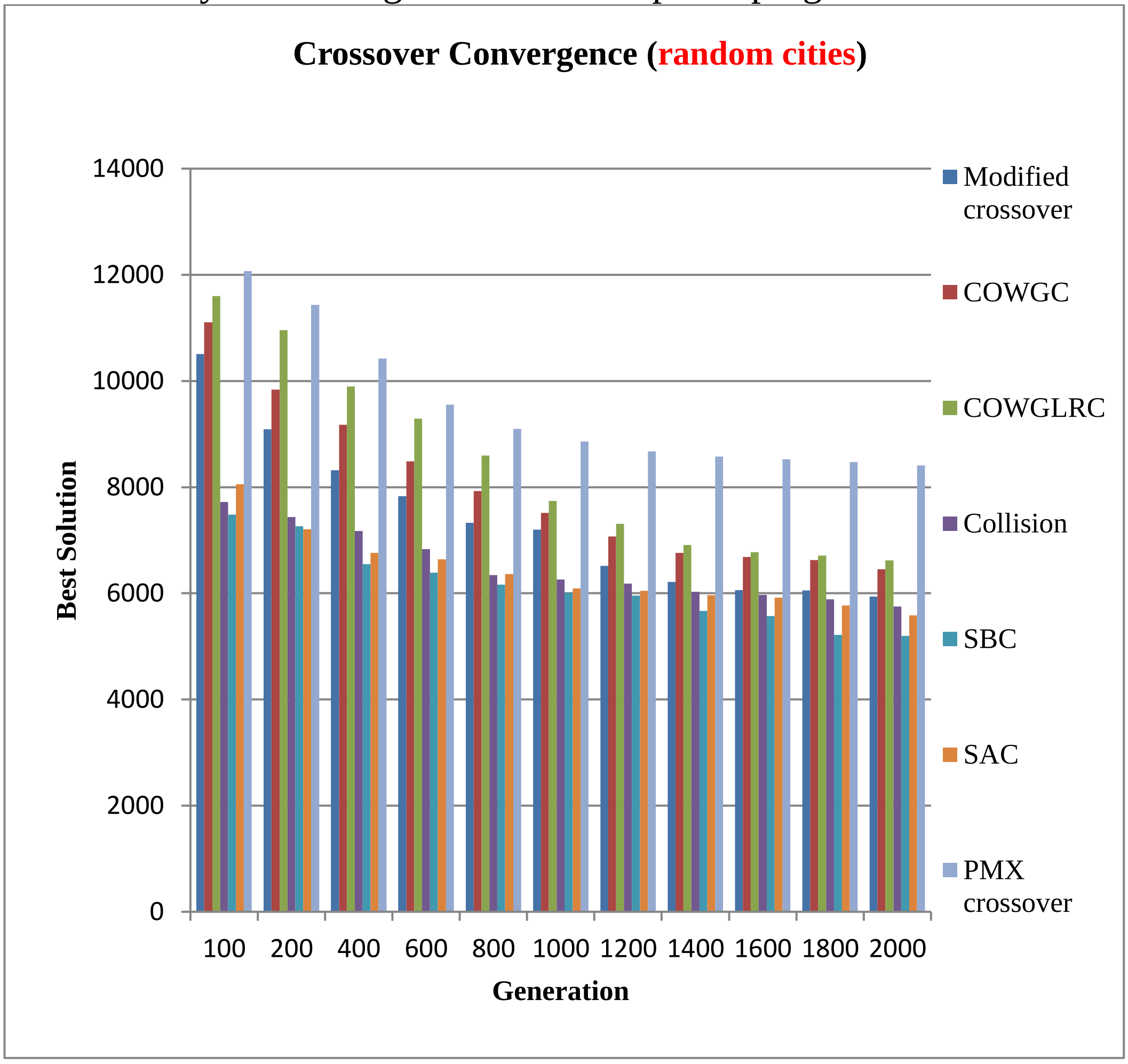

**Figure 3.6.**Crossover convergence with crossover ratio = 0.83 (random cities)

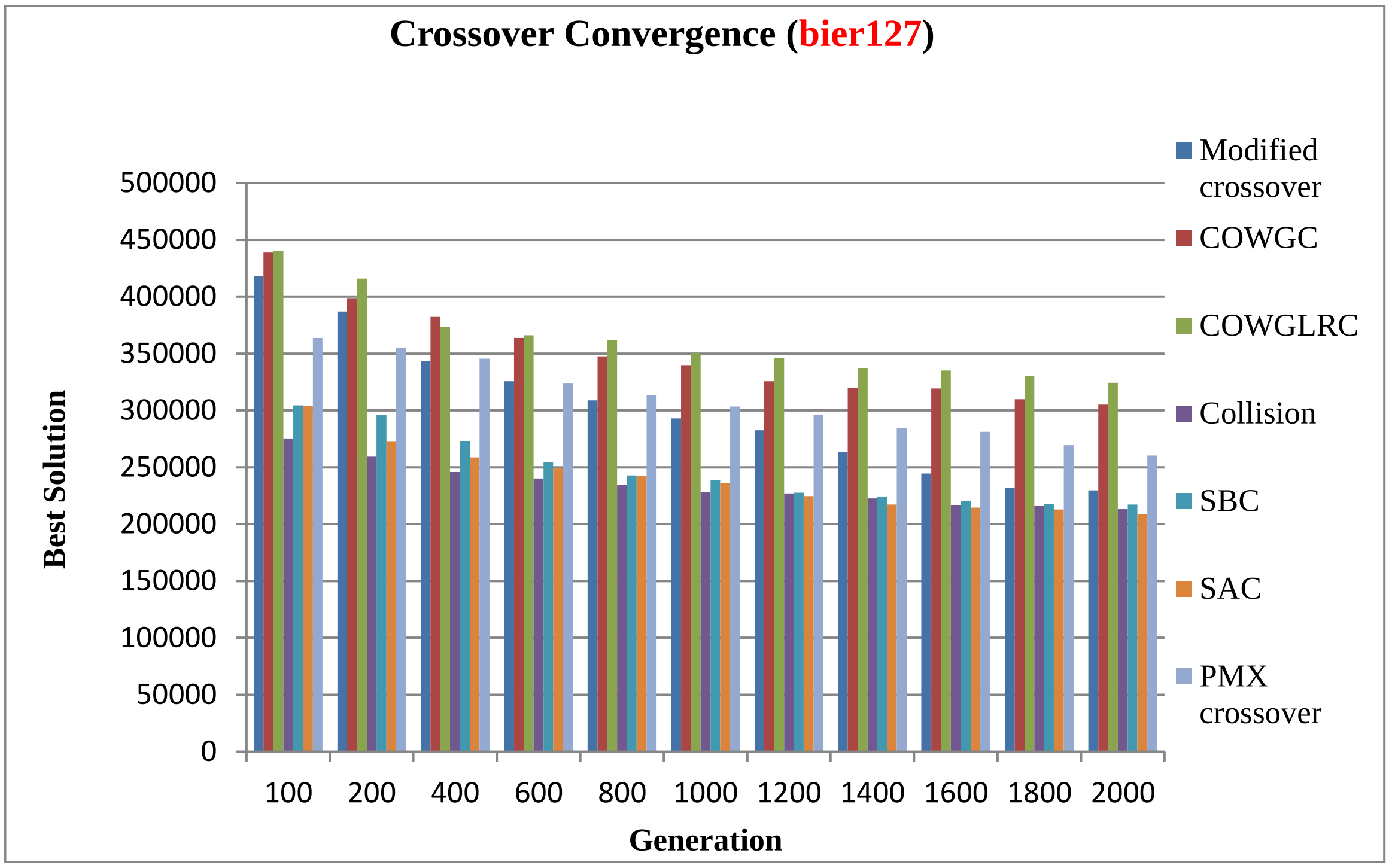

**Figure 3.7.**Crossover convergence with crossover ratio = 0.83 (bier127)

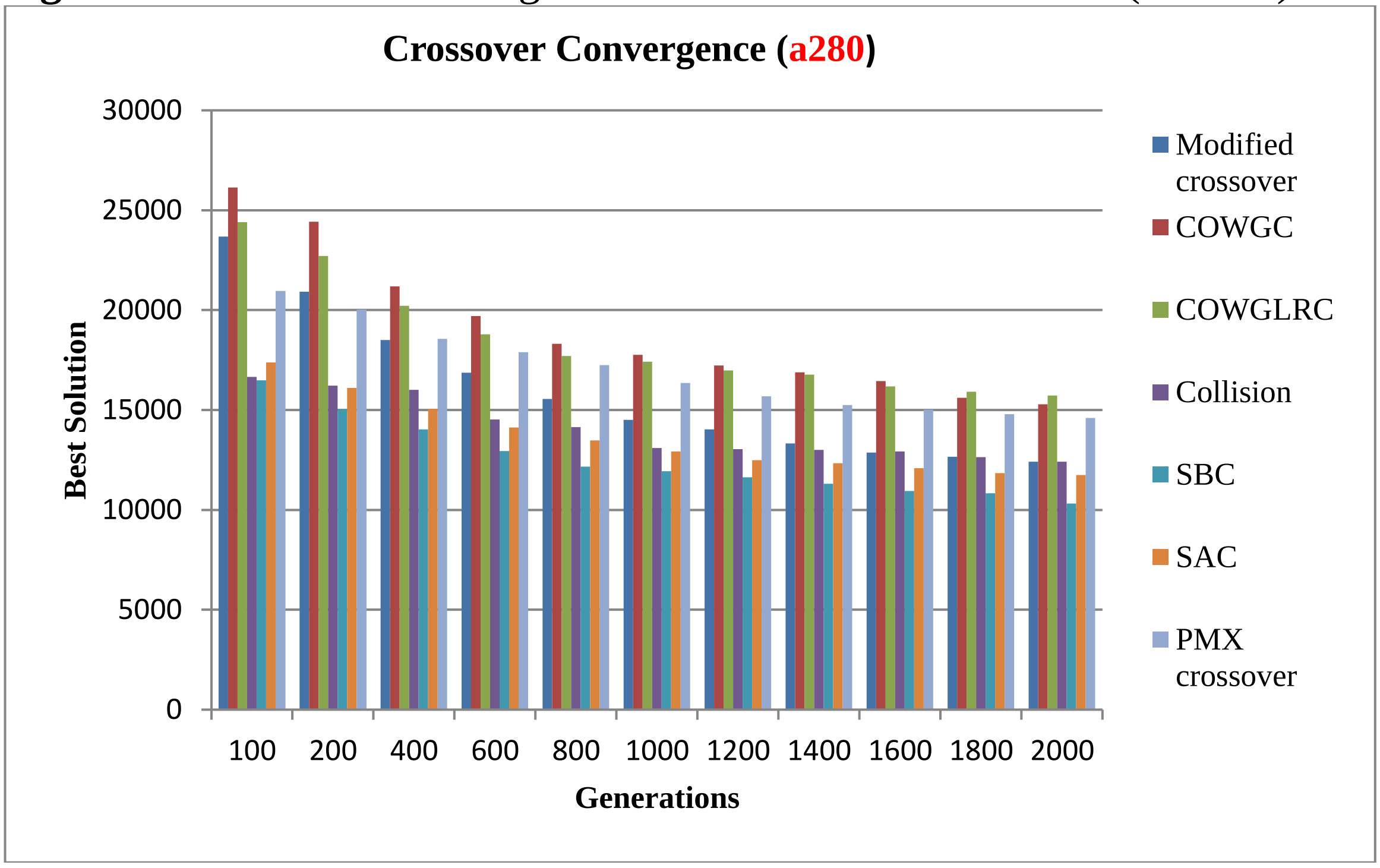

**Figure 3.8.**Crossover convergence with crossover ratio = 0.83 (a280)

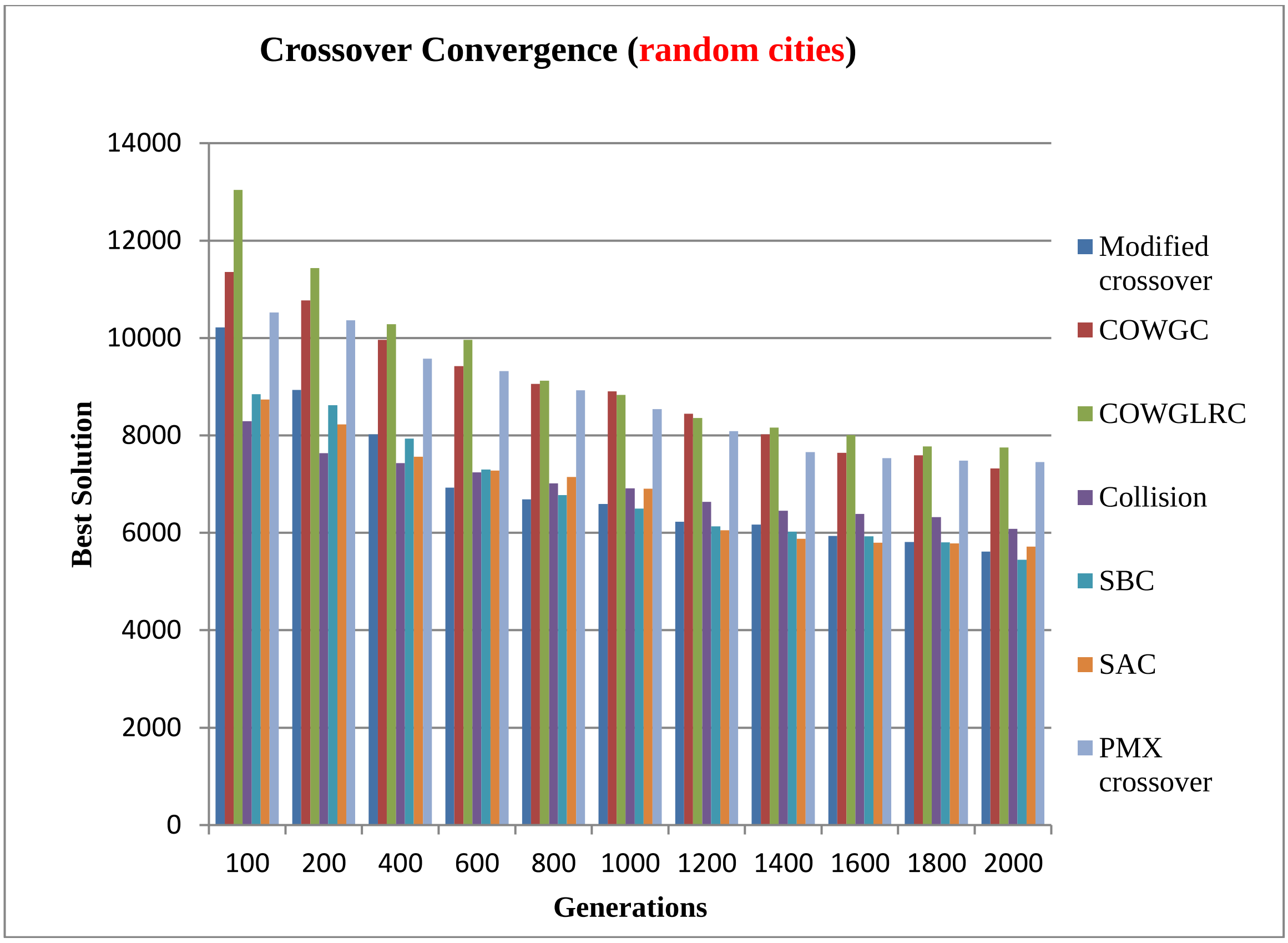

**Figure 3.9**. Crossover convergence with crossover ratio = 0.92 (random cities)

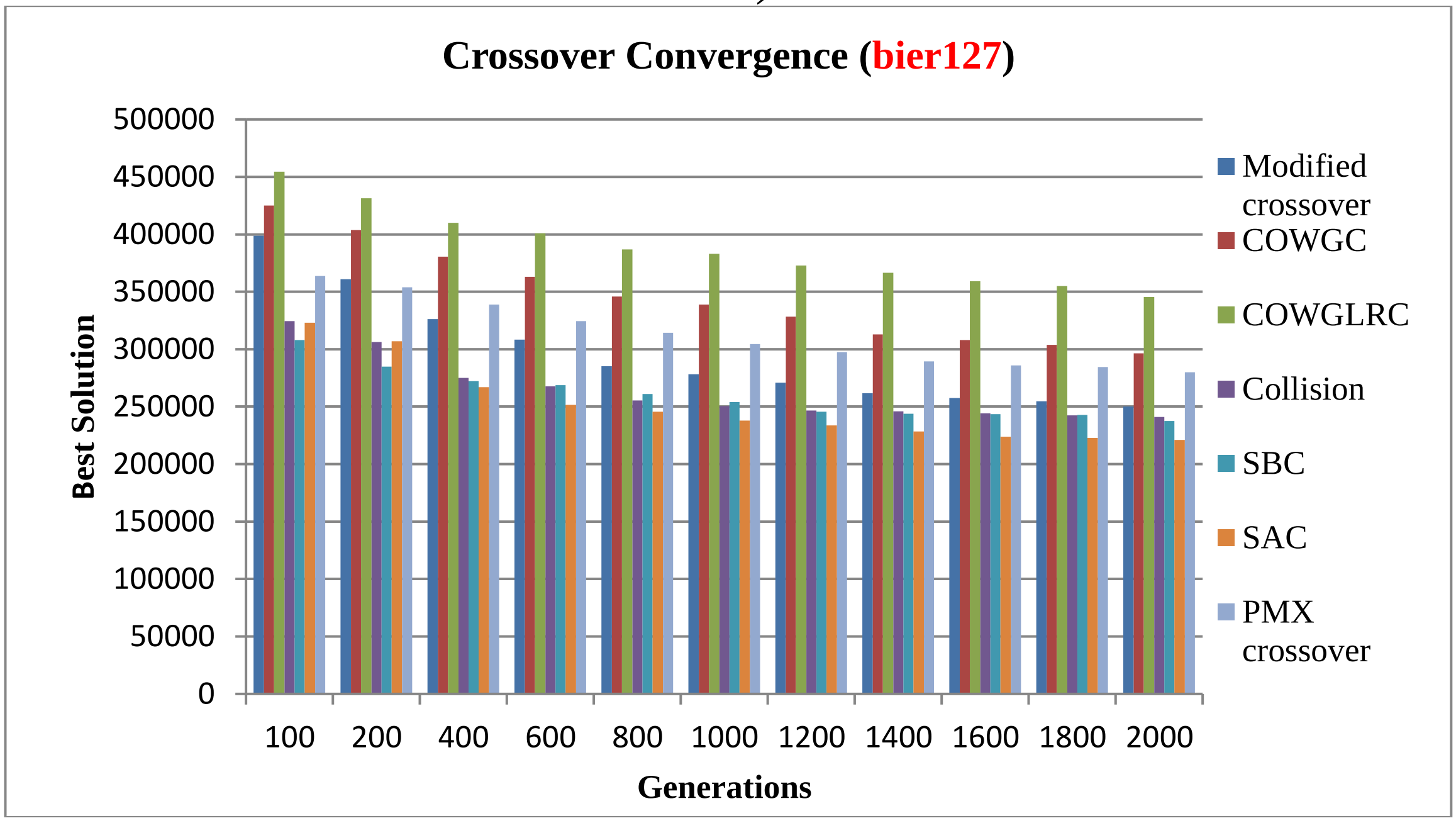

**Figure 3.10.** Crossover convergence with crossover ratio = 0.92 (bier127)

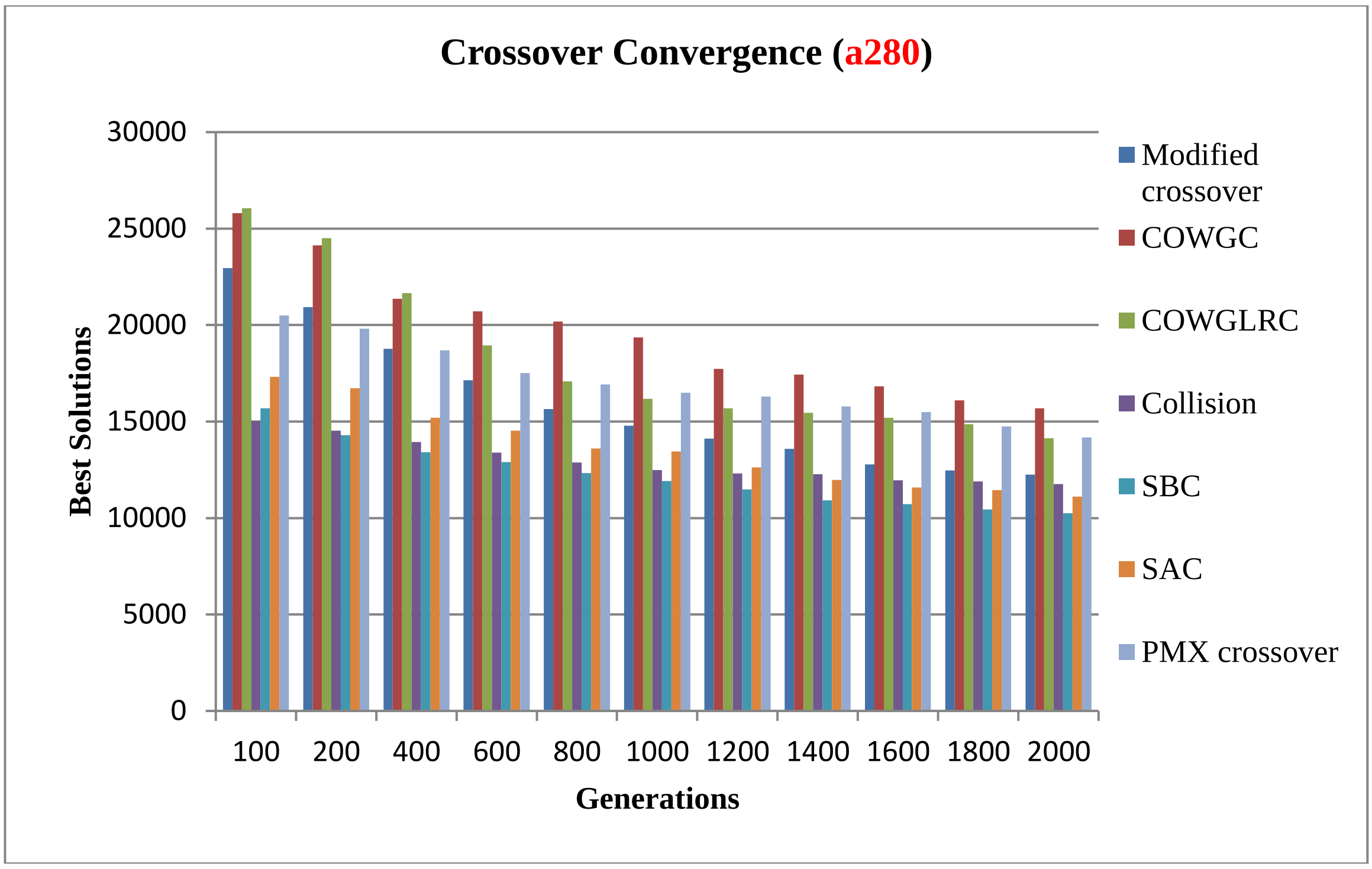

**Figure 3. 11.** Crossover convergence with crossover ratio = 0.92 (a280)

### 3.3.2 Experiments Set 2

In these experiments, we wanted to measure the effectiveness of the SBC and SAC algorithms in converging to an optimal solution. These algorithms, in addition to the Collision crossover, Modified crossover and PMX crossover, were tested using eleven real TSP problems taken from the TSPLIB, including: rat783, a280, u159, ch130 bier127, kroA100, pr76, berlin52, att48, eil51, pr144 (the numbers attached to the problem names represent the number of cities).

The GA parameters that were selected in the first test for all algorithms included the following: the crossover ratio: 100%, mutation ratio: 0%, the size of population was 200, and the maximum number of generations was 8000 (see Table (3.1)). In the second test the same parameters were used except for the size of population, which was reduced to 100 (see Table (3.2)).

**Table 3.1.**Comparison of crossover methods on converging to optimal solution (population size = 200)

| Name | #City | SBC | SAC | Collision crossover | PMX crossover | Modified crossover | Optimal Solution |
|---|---|---|---|---|---|---|---|
| **rat783** | 783 | 99581 | 104474 | **96647** | 143981 | 146769 | 8806 |
| **a280** | 280 | **16504** | 16881 | 17048 | 25131 | 24172 | 2579 |
| **u159** | 159 | 180580 | 213555 | **168037** | 291877 | 292269 | 42080 |
| **ch130** | 130 | 22407 | 24652 | **18724** | 29692 | 31834 | 6110 |
| **bier127** | 127 | **313554** | 335392 | 332013 | 433551 | 420988 | 118282 |
| **kroA100** | 100 | **53572** | 74023 | 57515 | 97490 | 97013 | 21282 |
| **pr76** | 76 | 226251 | 248414 | **214962** | 331358 | 320816 | 108159 |
| **berlin52** | 52 | 11409 | 13924 | **10224** | 16178 | 16142 | 7542 |
| **att48** | 48 | **47456** | 70669 | 51439 | 72269 | 75278 | 10628 |
| **eil51** | 51 | **642** | 735 | 649 | 949 | 878 | 426 |
| **pr144** | 144 | **309973** | 379911 | 335088 | 480611 | 504435 | 58537 |
| **Average** | - | 116539 | 134784.5 | 118395.1 | 174826.1 | 175508.5 | - |

As can be seen from Table (3.1), the results indicate the superiority of the SBC and the Collision crossover over the other methods, where the SBC performed the best for the problems: a280, bier127, kroA100, att48, eil51 and pr144. And the Collision crossover performed the best for the rest of the problems.

**Table 3.2**.Comparison of crossover methods on converging to optimal solution (population size = 100)

| Name | #City | SBC | SAC | Collision crossover | PMX crossover | Modified crossover | Optimal Solution |
|---|---|---|---|---|---|---|---|
| **rat783** | 783 | 99076 | 100301 | **96895** | 149925 | 152036 | 8806 |
| **a280** | 280 | **15157** | 16928 | 15335 | 25554 | 23027 | 2579 |
| **u159** | 159 | 196521 | **190217** | 192696 | 304470 | 277389 | 42080 |
| **ch130** | 130 | 20423 | 20942 | **20310** | 29501 | 29956 | 6110 |
| **bier127** | 127 | 333549 | 348203 | **324922** | 456519 | 412957 | 118282 |
| **kroA100** | 100 | **62968** | 81001 | 63269 | 98378 | 95493 | 21282 |
| **pr76** | 76 | **239327** | 254548 | 240782 | 336533 | 340641 | 108159 |
| **berlin52** | 52 | 13240 | 12336 | **11798** | 16456 | 14965 | 7542 |
| **att48** | 48 | 51959 | 60126 | **46315** | 77061 | 67630 | 10628 |
| **eil51** | 51 | 673 | 753 | **556** | 918 | 827 | 426 |
| **pr144** | 144 | 350733 | 333256 | **279038** | 455153 | 539113 | 58537 |
| **Average** | - | 125784.2 | 128964.6 | 117446.9 | 177315.3 | 177639.5 | - |

By reducing the population size to 100, the results degraded a bit as it can be seen from Table (3.2). Interestingly, the best performance was recorded by the Collision crossover, which outperformed the other methods in seven problems: rat783, ch130, bier127, berlin52, att48, eil51, pr144 followed by the SBC algorithm and SAC algorithm. The Modified crossover and PMX method showed slow convergence to near optimal solutions.

Having known that the Collision crossover came the second in the first set of experiments and the first in the second set of experiments in terms of finding the minimum solution and the fastest convergence (best solution in less number of generations), this put the proposed Collision crossover at the top of the compared methods.

Some of the solutions produced by the tested algorithms were close to the optimal solutions, but none could achieve an optimal solution. This shows the importance of using appropriate parameters along with crossover (such as population size, a higher mutation ratio) and appropriate number of generations, due to the effective impact of their convergence to near optimal solution.

## 3.4 Summary

In this chapter three methods and two strategies for crossover operator have been proposed. For the proposed types of crossover, two of them provide a heuristic search (worst gene) in determining the cutting point. The third type is inspired by the idea of the elastic collision.

The two strategies called SBC and SAC are trying to apply more than one crossover operator. While SBC applies all the specific types of crossover to the same parent and retains a good child, SAC applies a certain type of crossover randomly every time, in the hope that the algorithm will choose a different type every time. Both strategies aim to encourage diversity in the population, through new patterns that emerge when applying multi crossovers.

The proposed method has been tested using the well-known problem (TSP). Comparisons have also been made between the proposed type and the well-known Modified crossover and PMX crossover.

# Chapter 4
# Proposed Work for Mutation Operator

## 4.1 Design and Methodology

One of the difficulties of the GA is so-called premature convergence (Nicoară, 2009) and the reason for this is directly related to the loss of diversity (Suh & Van Gucht, 1987). Achieving population diversity is the desired goal, and according to that the search path becomes better, and also avoids trapping into a suboptimal solution. According to Holland, mutation is considered an important mechanism for maintaining diversity (Deb & Deb, 2014); (Wagner, Affenzeller, Beham, Kronberger, & Winkler, 2010), and explores new areas in the search space, thus avoiding the convergence of the local optimum (Korejo, Yang, Brohi, & Khuhro, 2013). The need for mutation is to prevent loss of genetic material, where the crossover does not guarantee access to the new search space (Deep & Mebrahtu, 2011), therefore, random changes in the gene through mutation helps in providing variations in the population (Yang, 2002).

Many researchers have resorted to preventing local convergence in different ways, and because mutation is a key operation in the search process, we find several mutation methods in the literature. The question is, what is the best method to use?

To answer this question, and in the hope of avoiding the local optima and increasing the diversification of the population, we have proposed and implemented two strategies and several types of mutations, to be compared with two of the well-known types namely: Exchange mutation and Rearrangement mutation (Sallabi & El-Haddad, 2009).

### 4.2 Mutation Operator

In the following we describe each mutation operator. It is important to note that mutation methods described in subsections 4.2.4 to 4.2.10 are designed especially for the TSP problem. However, they can be customized to fit some other problems.

#### 4.2.1 Worst gene with random gene mutation (WGWRGM)

To perform this mutation we need to search for the worst gene in the chromosome from index 0 to L-1, where L is the length of the chromosome. The worst gene varies depending on the definition of the "worst" for each problem; the worst gene is the point that contributes the maximum to increase the cost of the fitness function of a specific chromosome. For example, the worst gene in the TSP problem is the city with the maximum distance from its left neighbour, while the worst gene in the Knapsack problem (for instance) is the point with the lowest value to weight ratio, and so on.

After the worst gene is identified, we select another gene at random, and swap the genes in these positions as in the Exchange mutation. Figure (4.1) shows an example of (WGWRGM).

The worst gene (WG) can be calculated for a minimization problem such as TSP using:

$$\mathrm{WG} = \underset{1 \le i < n}{\mathrm{argmax}}(\mathrm{Distance}(C[i], C[i+1]) \qquad (9)$$

and for a maximization problem such as the Knapsack problem using:

$$\mathrm{WG} = \underset{0 \le i < n}{\mathrm{argmin}}(\mathrm{Distance}(C[i], C[i+1]) \qquad (10)$$

where *C* represents the chromosome, i is the index of a gene within a chromosome, and the distance function for the TSP can be calculated using either Euclidian distance or the distances table between cities. In the case of TSP, searching for the WG starts at index 1, assuming that the route-starting city is located at index 0, while this is not the case for other problems such as the Knapsack problem (equation (10)).

The previous equations are used for the chromosome, and the worst gene of this chromosome that exhibits the maximum distance is used for the mutation operation.

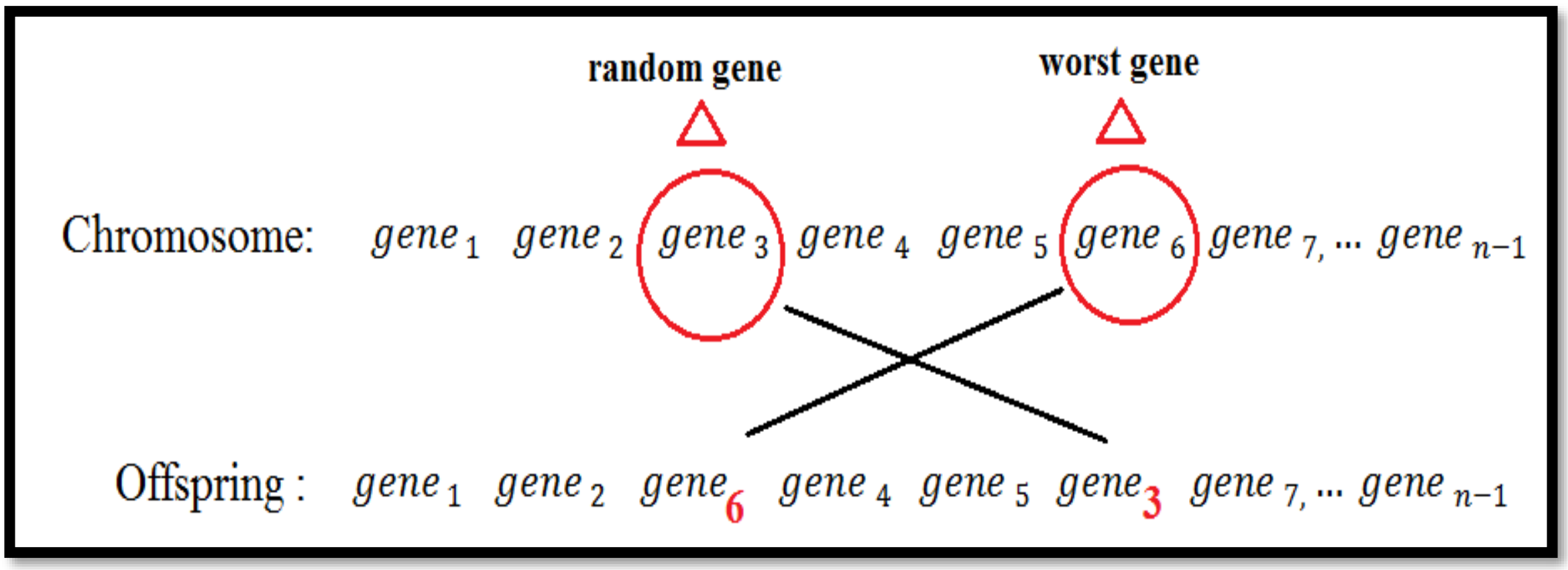

**Figure 4.1.** Example of WGWRGM

**Example 1. (TSP problem)**
Suppose that the chromosome chosen to be mutated is:
CHR1: A→B→E→D→C→A as depicted in Figure ((4. 2) (a)).

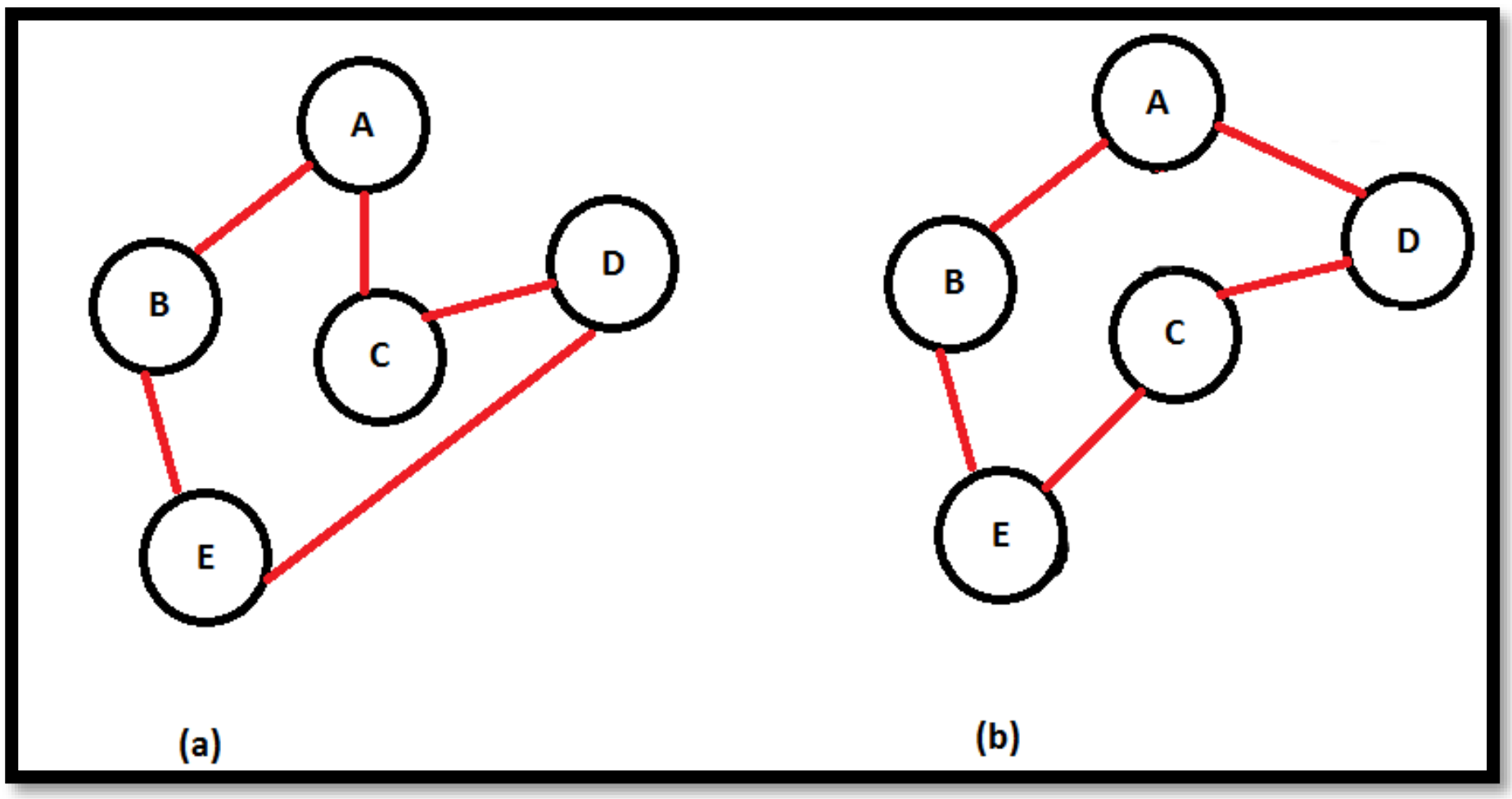

**Figure 4.2.** Example of applying WGWRGM to a specific chromosome of particular TSP

To apply WGWRGM:

a. Step 1: Find the worst gene in the parent; according to the graph, the worst gene is (D).
b. Step 2: Suppose that the city which has been selected at random is (C).
c. Step 3: Apply the Exchange mutation in this chromosome by swapping the positions of the two cities(see Figure (4.2) (b)). The output offspring: A→B→E→C→D→A.

### 4.2.2 Worst gene with worst gene mutation (WGWWGM)

This type is similar to the WGWRGM, the only difference being that we search for two worst genes. Both worst genes exchange positions with each other. Finding both worst genes is similar to finding the two maximum value algorithms if the problem being dealt with is a minimization problem, and for the maximization problem, finding the two minimum values of the algorithm can be used, as the definition of the worst gene concept is different from one problem to another (see Figure (4.3)).

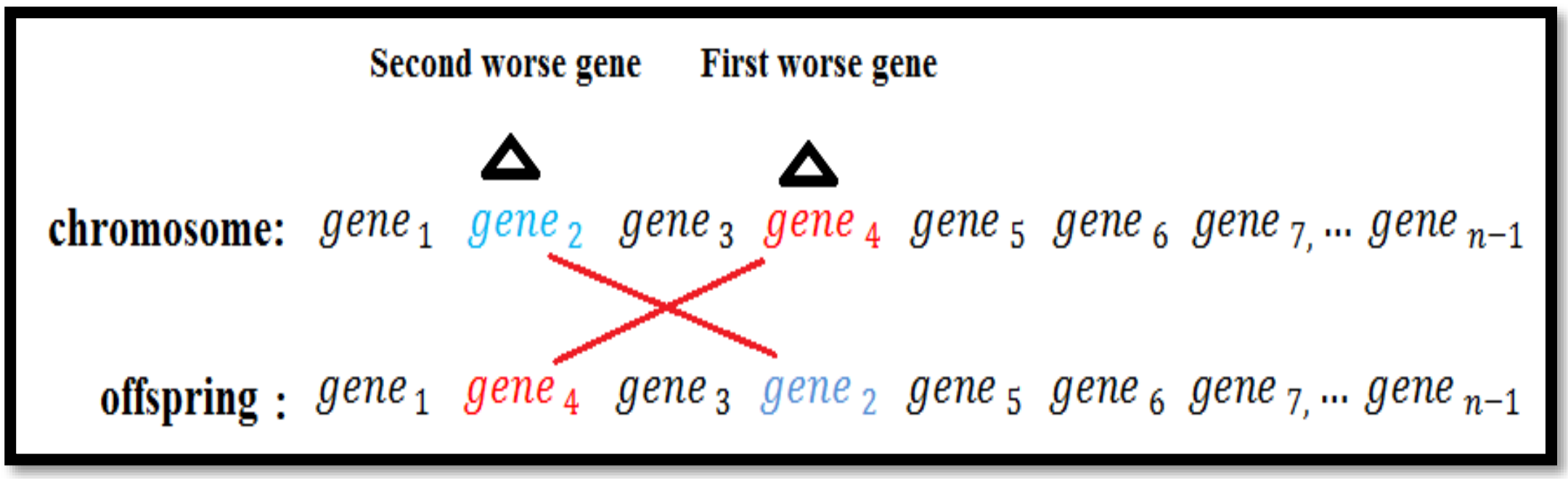

**Figure 4.3.** Example of WGWWGM

### 4.2.3 Worst LR gene with random gene mutation (WLRGWRGM)

This method is also similar to the WGWRGM, the only difference being that the worst gene is found by calculating the distance between both its neighbours: the left and the right.

The worst gene (WLRgene) can be calculated for the TSP using:

$$W_{LRgene} = \underset{1 \le i < n-2}{\operatorname{argmax}}(\text{Distance}(C[i], C[i-1] + \text{Distance}(C[i], C[i+1]) \qquad (11)$$

$$W_{LRgene} = \underset{1 \le i < n-2}{\operatorname{argmin}}(\text{Distance}(C[i], C[i-1] + \text{Distance}(C[i], C[i+1]) \qquad (12)$$

Equation (11) can be used for minimization problems, and Equation (12) for maximization problems. The extreme genes (the first and last ones) can be handled in a circular way, i.e. the left of the first gene is the last gene.

The worst gene (for minimization problems) is the one that is the sum of the distances with its left and right neighbours, which is the maximum among all genes within a chromosome, and the other way round for minimization problems. In this mutation, the position of the worst gene is altered with the position of another gene chosen randomly.

**Example 2. (TSP problem)**

Figure ((4.4) (a)) represents the chromosome chosen to be mutated which is: Chromosome: A→B→E→H→F→D→C→A.

According to Figure ((4.4) (a)), (WLRgene) is the city (D) because the total distance from city D to city F and from city D to city C is the maximum. If city (H) is chosen randomly, the output offspring after applying WLRGWRGM mutation is: A→B→E→D→F→H→C→A (see Figure (4.4) (b)).

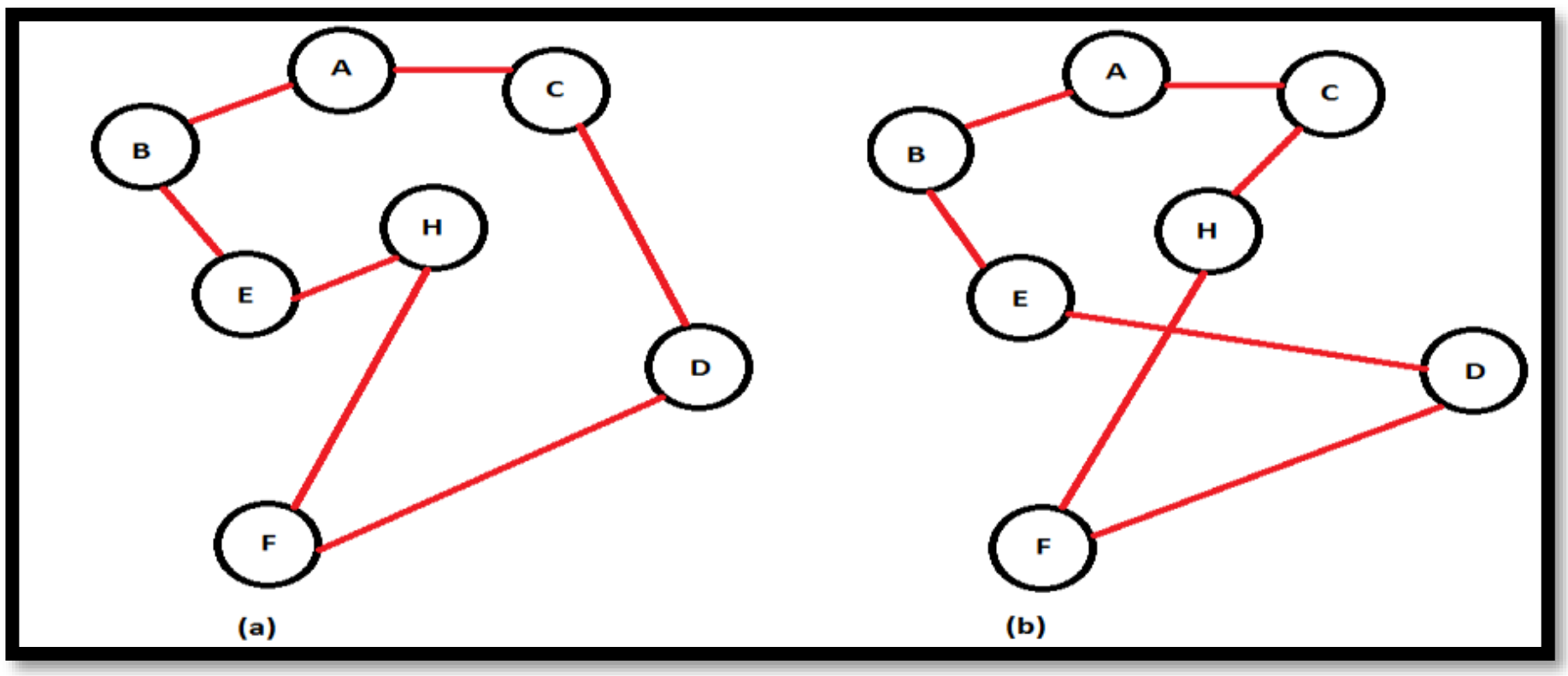

**Figure 4.4.** Example of applying WLRGWRGM to a specific chromosome of particular TSP

### 4.2.4 Worst gene with nearest neighbour mutation (WGWNNM)

This method uses the idea of the nearest neighbour cities (knowledge-based method), where it provides an heuristic search process for mutation, and is performed as follows:

a. Step 1: Search for the gene (city) in a tour characterized by the worst with its left and right neighbours (WLRgene) as in WLRGWRGM mutation; this city is called "worst city".
b. Step 2: Find the nearest city to the worst city (from the graph), and call it Ncity. Then search for the index of that city in the chromosome, calling it Ni.

We need to replace the "worst city" with another one around the "Ncity" other than the "Ncity". The term "around" is defined by a predefined range, centered at the "Ncity". We arbitrarily used (Ni ± 5) as a range around the index of the "Ncity". The out-of-range problem with the extreme points is solved by dealing with the chromosome as a circular structure.

c. Step 3: Select a random index within the range; the city at that index is called "random city".
d. Step 4: Swap between "worst city" and "random city".

**Example 3. (TSP problem)**

Suppose that the chromosome chosen to be mutated is:

Chromosome: A→B→F→D→E→C→H→A as depicted in Figure ((4.5) (a)).

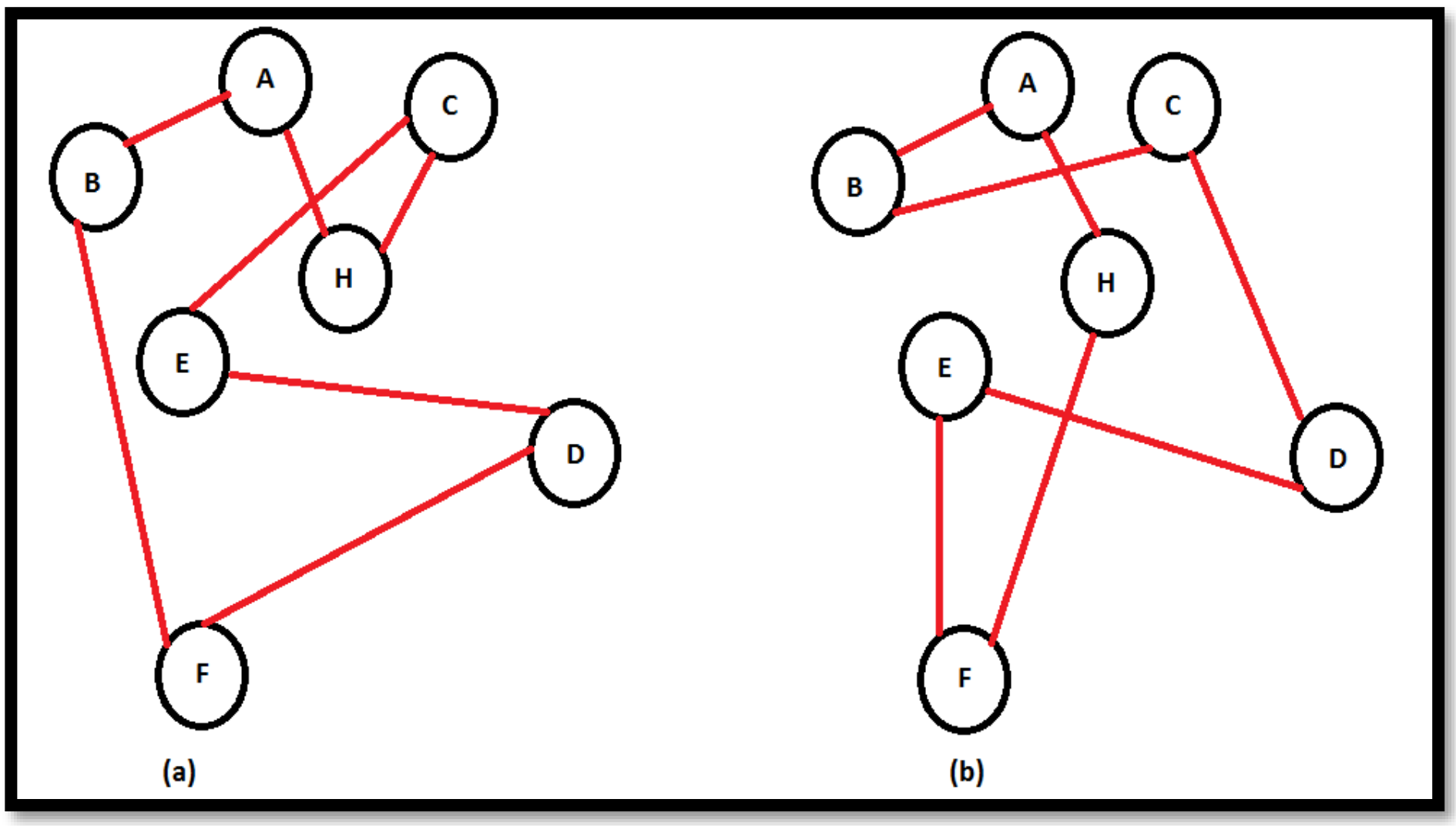

**Figure 4.5.** Example of applying WGWNNM to a specific chromosome of particular TSP

To apply WGWNNM:

a. Step 1: Find the (WLRgene) in the chromosome; according to the graph, the worst city is (F).
b. Step 2: Find the nearest city to the worst city, which is (E); this city is called Ncity.
c. Step 3: Search for a city around Ncity at random in the range (± 5); suppose we choose city (C).
d. Step 4: Apply the Exchange mutation in this chromosome by swapping the position of the two cities (see Figure (4.5) (b)). The output offspring is: A→B→C→D→E→F→H→A.

### 4.2.5 Worst gene with the worst around the nearest neighbour mutation (WGWWNNM)

This mutation is similar to the WGWNNM; the only difference is in the selection of the swapped city. The swapped city is not selected randomly around the nearest city as in WGWNNM, but rather is chosen based on its distance from the nearest city. By considering the furthest city from the nearest city to be swapped with the worst city, this brings nearest cities together, and sends furthest cities far away.

### 4.2.6 Worst gene inserted beside nearest neighbour mutation (WGIBNNM)

This type of mutation is similar to the WGWNNM, after finding the indices of the worst city and its nearest city. The worst city is moved to neighbour its nearest city, and the rest of the cities are then shifted either left or right depending on the locations of the worst city and its nearest city.

In other words, if the worst city was found to the right of its nearest city, the worst city is moved to the left of its nearest city, and the other cities are shifted to the right of the location of the worst city. If the worst city was found to the left of its nearest neighbour, the worst city is moved to the location prior to the location of its nearest city, and the rest of the cities between this location and the previous location of the worst city are shifted to the right of that location, and the other way round otherwise.

### 4.2.7 Random gene inserted beside nearest neighbour mutation (RGIBNNM)

This mutation is almost the same as the WGIBNNM, except that the "worst city" is selected randomly, and is not based on its negative contribution to the fitness of the chromosome. We reckon that RGIBNNM is an enhancement of the WGIBNNM to enforce some randomness.

### 4.2.8 Swap worst gene locally mutation (SWGLM)

This mutation is performed as follows:

a. Step 1: Search for the "worst gene", the same as for WLRGWRGM.
b. Step 2: Swap the left neighbour of the "worst gene" with its left neighbour, and calculate the fitness (C1) of the new child (F1).
c. Step 3: Swap the "worst gene" with its right neighbour, and calculate the fitness (C2) of the new child (F2).
d. Step 4: If C1 > C2, then return F2 as the legitimate child and delete F1, otherwise return F1 as the legitimate child and delete F2 (see Figure (4.6)).

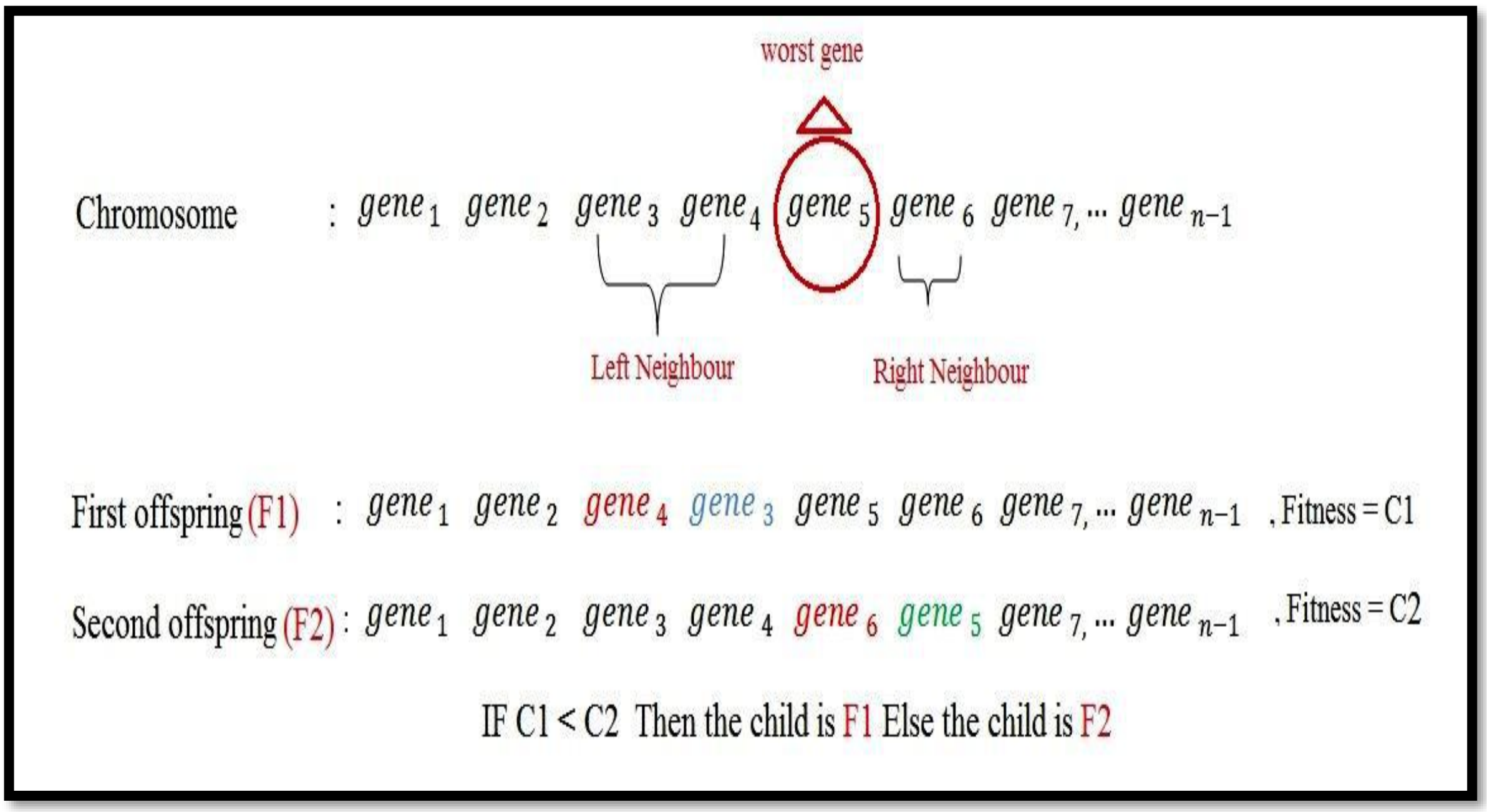

**Figure 4.6.** Example of SWGLM

**Example 4. (TSP problem)**

Suppose that the chromosome chosen to be mutated is:

Chromosome: A→B→F→E→H→D→C→A as depicted in Figure ((4.7) (a)).

To apply SWGLM:

a. Step 1: Find the "worst gene" in the chromosome. According to the graph, the worst city is (E).
b. Step 2: Swap between two left neighbours of E, which is (B and F), and the first offspring is: A→F→B→E→H→D→C→A.The cost of this offspring is C1 (see Figure (4.7) (b)).
c. Step 3: Swap between worst city (E) and its right neighbour. The second offspring is:
A→B→F→H→E→D→C→A.The cost of this offspring is C1 (see Figure (4.7) (c)).
d. Step 4: Compare the cost (C1, C2) and the least among them are the legitimate children.

Based on the graph the output offspring is: A→B→F→H→E→D→C→A (second offspring).

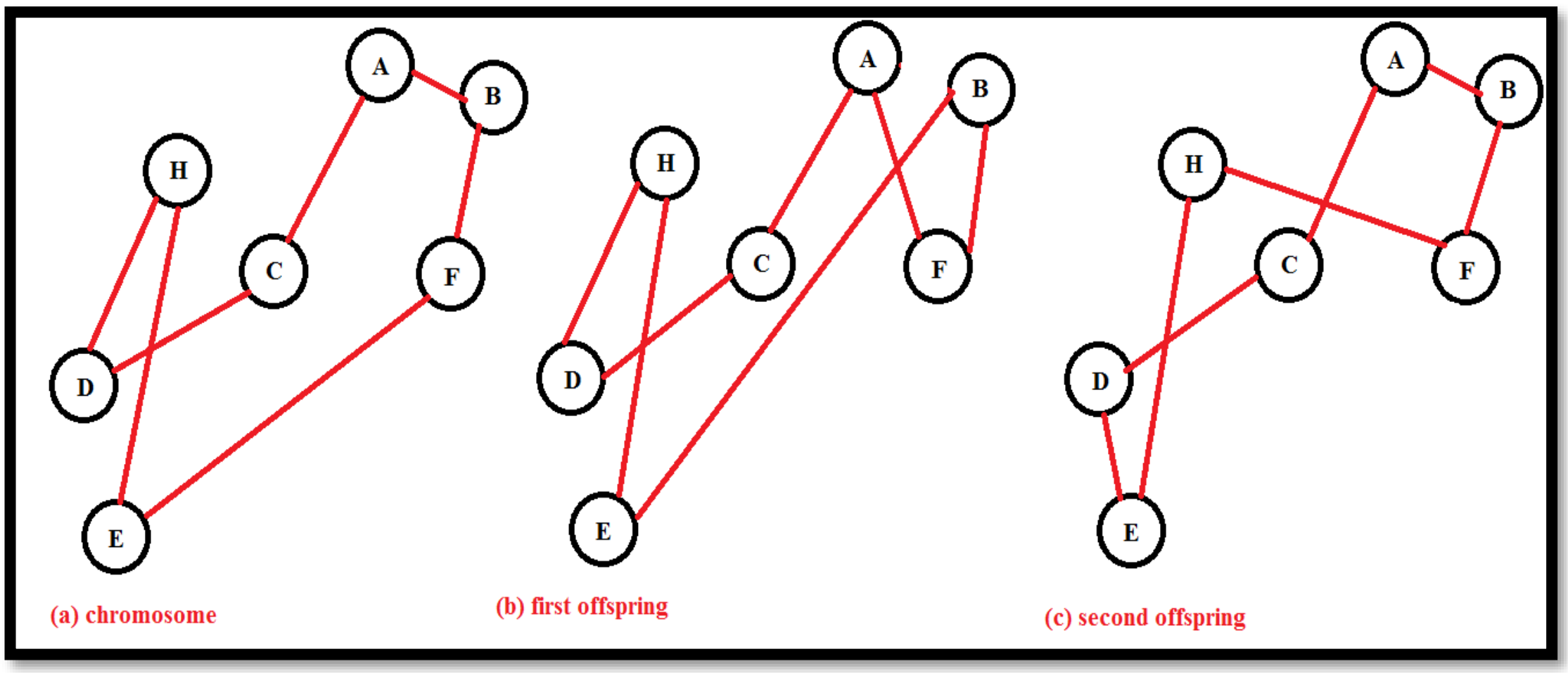

**Figure 4.7.** Example of applying SWGLM to a specific chromosome of particular TSP

### 4.2.9 Insert best random gene before worst gene mutation (IBRGBWGM)

This mutation works as follows:

1. Search for the city that is characterized by the worst city as in WGWRGM and find the index of its previous city.
2. Select a certain number of random cities (in this work we choose 5 random cities arbitrarily, excluding the "worst city" and its previous neighbour (PN)).
3. For each random city calculate the distance to the "worst city" (D1) and the distance to PN (D2).
4. Find the "best city" from the random cites, which is the one with the minimum (D1+D2).

5. Move the “best city” to be inserted between the “worst city” and PN.
6. Shift cities which are located between the old and the new location of “best city” to legitimize the chromosome.

**Example 5. (TSP problem)**

Figure ((4.8) (a)) represents the chromosome chosen to be mutated which is: Chromosome: A→B→E→D→ C→A.

According to Figure ((4.8) (a)), the worst gene is the city (E). According to the graph the best city is (C), and the output offspring after applying IBRGBWGM mutation is: A→B→C→E→D→A (see Figure (4.8) (b)).

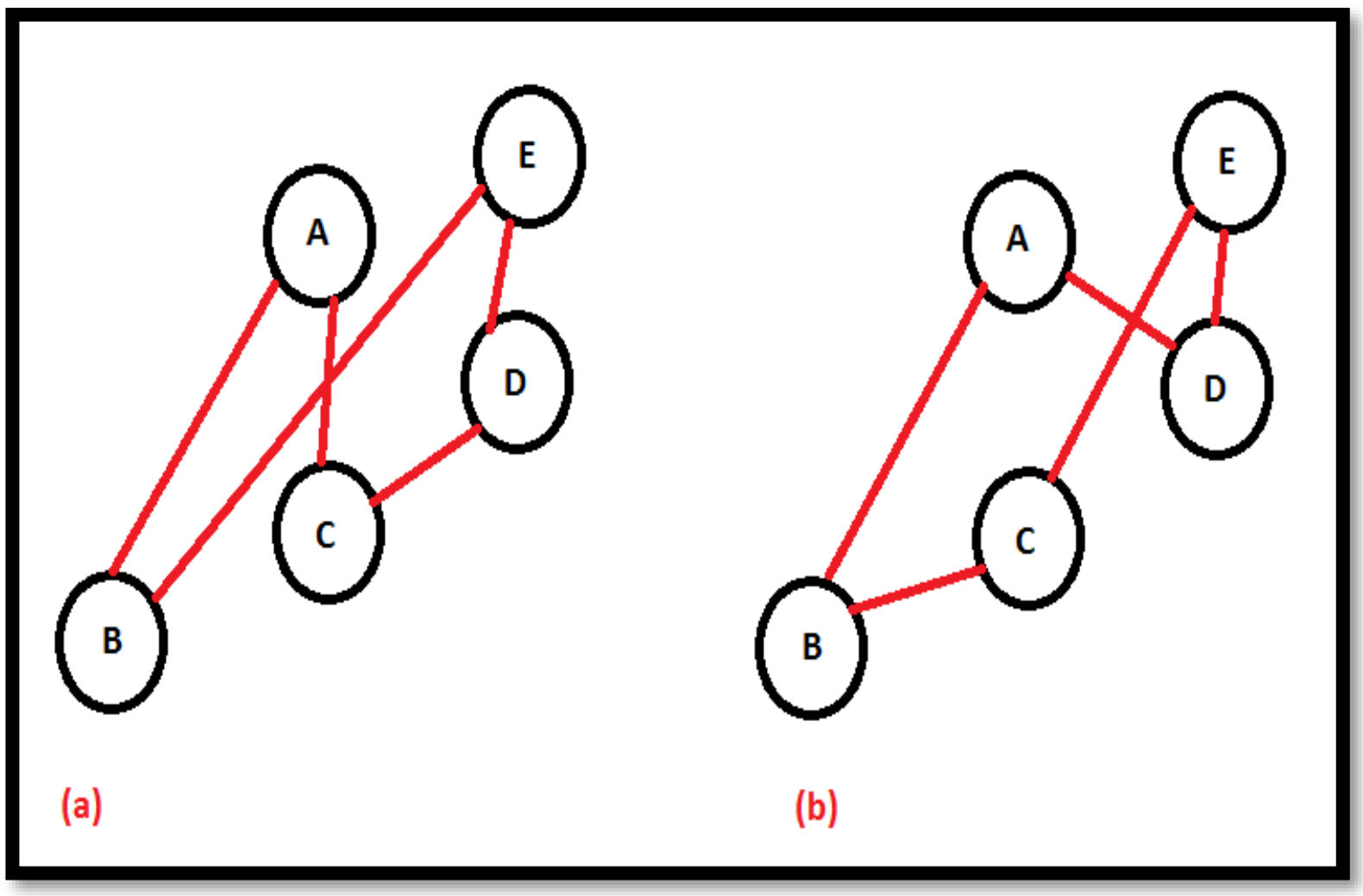

**Figure 4.8.** Example of applying IBRGBWGM to a specific chromosome of particular TSP

### 4.2.10 Insert best random gene before random gene mutation (IBRGBRGM)

This mutation is similar to IBRGBWGM, the only difference being that the “worst city” is not chosen based on any distance; rather it is chosen randomly to impose some diversity among the new offspring.

### 4.2.11 Multi Mutation Operators Algorithms

A traditional GA commonly used is one mutation operator. We propose using more than one mutation operator, which is the different mutation operators that hopefully lead to different directions in the search space, and thus increases diversification in the population, then improves the performance of the GA.

To do this we opted for two selection approaches: the best mutation, and a randomly chosen mutation.

#### 4.2.11.1 Select the best mutation algorithm (SBM)

This algorithm applies multiple mutation operators at the same time to the same chromosome, and considers the best one child to be added to the population, to prevent duplication; only the best and not found in the population is added.

In this work, from the beginning, we have defined the mutation methods to be applied. This algorithm implements all the aforementioned methods (WGWRGM, WGWWGM, WLRGWRGM, WGWNNM, WGWWNNM, WGIBNNM, RGIBNNM, SWGLM, IBRGBWGM and IBRGBRGM) one after the other, and from each method produces one offspring; the best child that does not already exist in the population is added.

This anticipates that such a process encourages diversity in the population, and thus avoids convergence to local optimal.

#### 4.2.11.2 Select any mutation algorithm (SAM)

This algorithm tries to apply multi mutation operators each time. The selection strategy is random, and it randomly chooses one of the aforementioned operators in a certain generation. Therefore, we reckon that in each generation a different operator of mutation is chosen. This means that there is a different direction of the search space, and this is what we are aiming for, increasing diversification, and attempting to enhance the performance of the GA.

## 4.3 Experimental Results and Discussion

To evaluate the proposed methods, we conducted two experiments on different TSP problems. The aim of the first experiment was to examine convergence to the minimum value of each method separately. The second experiment was designed to examine the efficiency of the SBM and SAM algorithms and compare their performance with the Exchange mutation using real data.

### 4.3.1 Experiments Set 1

The aim of these experiments was to measure the effectiveness of the proposed mutation operators (WGWRGM, WGWWGM, WLRGWRGM, WGWNNM, WGWWNNM, WGIBNNM, RGIBNNM, SWGLM, IBRGBWGM and IBRGBRGM). The results of the experiments were compared to the two already existing mutations, namely: Exchange mutation (Banzhaf, 1990), Rearrangement mutation (Sallabi & El-Haddad, 2009) in addition to measuring the performance of the GA when using either SAM or SBM.

These mutation operators were tested using three test data: the first was random cities, where the coordinates of the cities were chosen randomly, and the second and third test data were real data: “bier127“and “a280” taken from TSPLIB (Reinelt & Gerhard, 1996), each of them consisting of 100, 127, and 280 cities respectively.

The GA parameters that were selected included the following: population size: 100, the probability of crossover: (0%), mutation probability: (100%), and the maximum generation was 1600. The algorithm was applied eleven times to different generations starting with 100.

Results from the first test indicated that the best performance was recorded by the SBM, followed by the SAM. This compared well with the rest of the mutation methods, because it showed good convergence to a minimum value.

The efficiency of each one of the fourteen mutations is shown in Figures (4.9 - 4.11).A closer look at these Figures reveals that the SBM and SAM algorithms outperformed all other methods in the speed of convergence. Here we will analyze the results as follows:

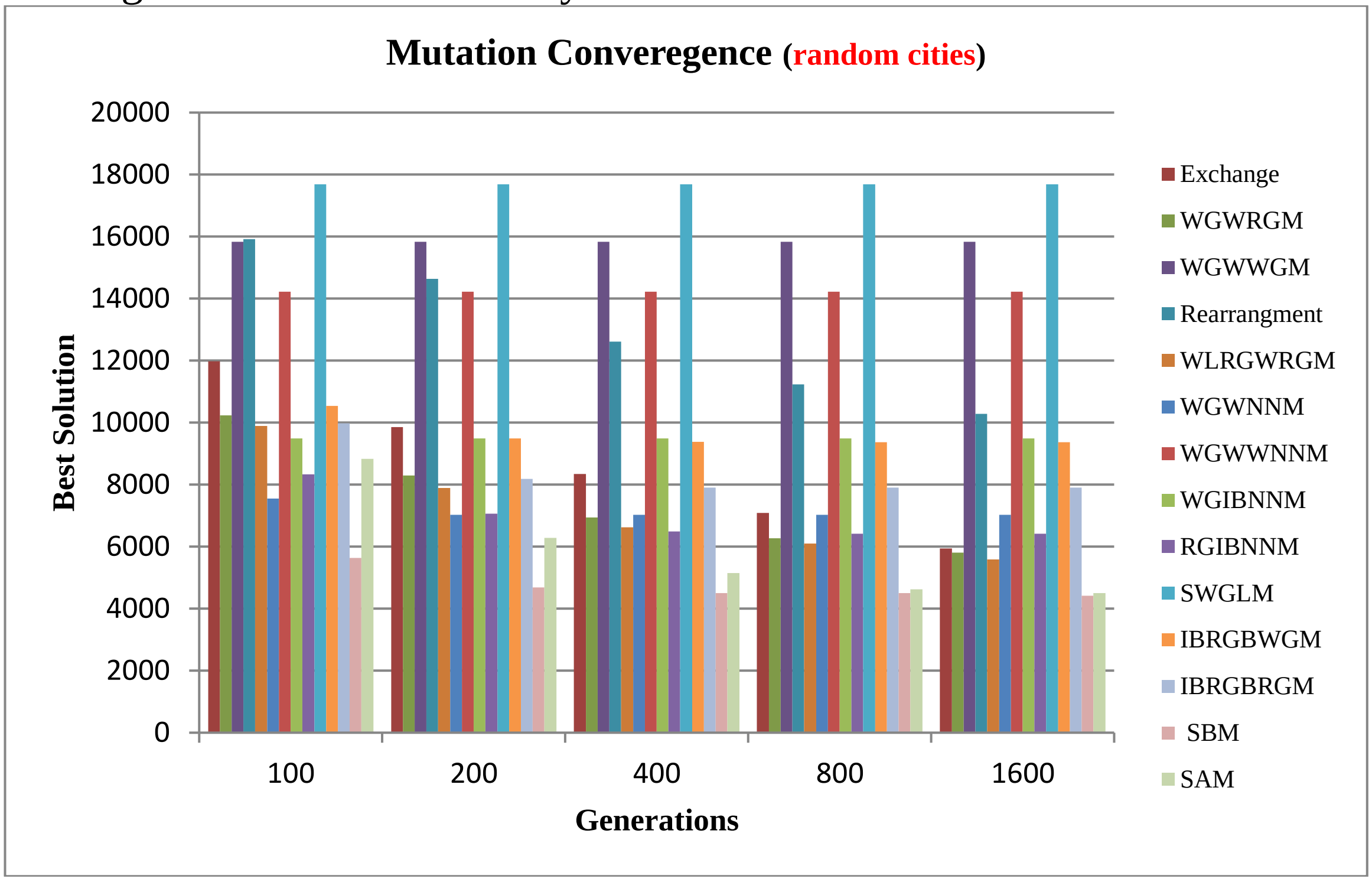

**Figure 4.9.** Mutation convergence to the minimum value (random cities)

As can be observed from Figure (4.9), the results indicate the superiority of the SBM and SAM algorithms, where the speed of convergence of the optimal solution with the progress of the generations is faster than the use of a certain type of mutation alone. The WLRGWRGM, followed by WGWRGM and exchange mutations, also showed the extent of their influence on the quality of the solution.

A result in Figure (4.10) indicates that the SBM algorithm showed faster convergence to the minimum value, followed by SAM, and these algorithms showed better performance than the remaining mutations. At the level of mutation alone, the exchange mutation, followed by RGIBNNM and IBRGBRGM, showed better performance than the other mutations.

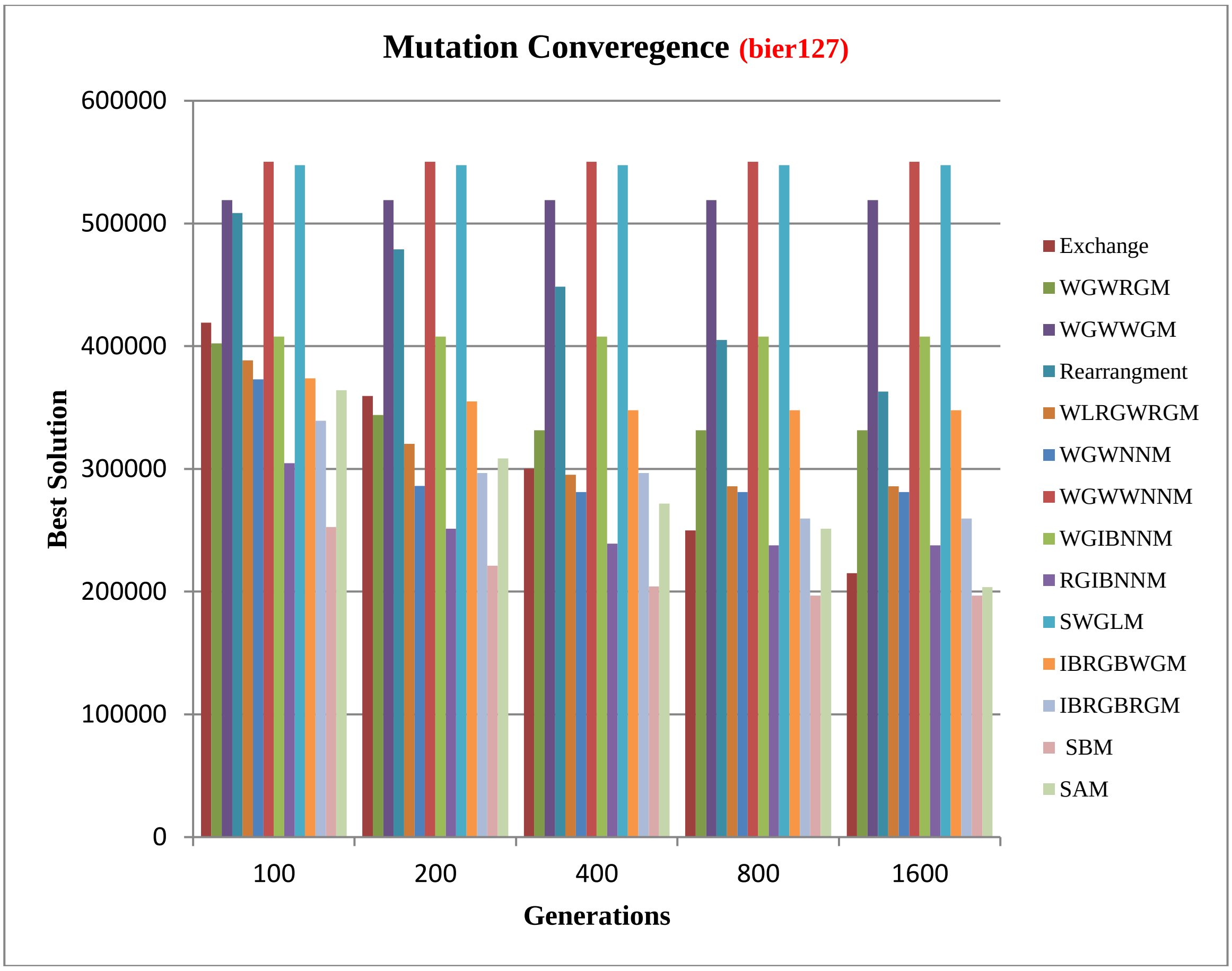

**Figure 4.10.** Mutation convergence to the minimum value (bier127)

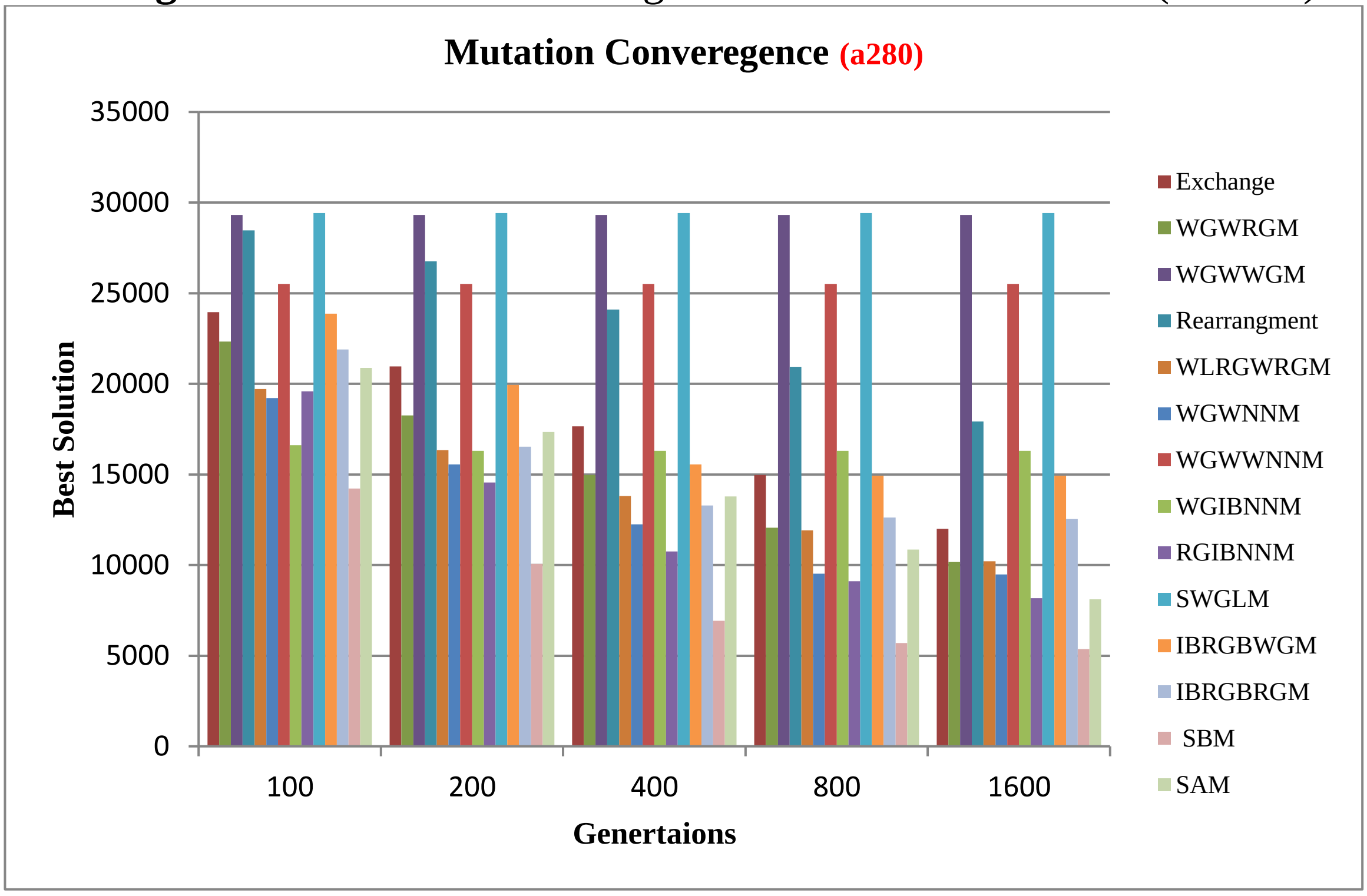

**Figure 4.11.** Mutation convergence to the minimum value (a280)

As can be seen from Figure (4.11), the best performance was recorded by the SBM algorithm, and especially after the 200 generation, which showed faster convergence to the minimum value than any other mutation, followed by the SAM algorithm. At the level of mutations alone, RGIBNNM, WGWNNM and WGWRGM mutations showed better performance than the rest of the mutations. Because of the slow convergence of the SWGLM and WGWWGM mutations, they achieved the worst result.

Although the SBM outperformed the SAM, SAM is still better than SBM in terms of time spent, because SBM tries all mutations available and chooses the best, while SAM selects any one randomly. Moreover, the difference between the two results is not great.

### 4.3.2 Experiments Set 2

In these experiments, we wanted to measure the effectiveness of the SBC and SAC algorithms in converging to an optimal solution. These algorithms, in addition to the Exchange mutation and rearrangement mutation, were tested using eleven real TSP problems taken from the TSPLIB, including: rat783, a280, u159, ch130 bier127, kroA100, pr76, berlin52, att48, eil51, pr144 (the numbers attached to the problem names represent the number of cities).

The GA parameters that were selected in the first test for all algorithms included the following: the crossover ratio: 0%, mutation ratio: 100%, the size of population was 200, and the maximum number of generations was 8000 (see Table (4.1)). In the second test the same parameters were used except for the size of population, which was reduced to 100 (see Table (4.2)).

**Table 4.1.** Result for eleven problems obtained for 4 mutation operators after 8000 generations (population size=200)

| Name | #City | SBM | SAM | Rearrangement | Exchange mutation | Optimal |
|---|---|---|---|---|---|---|
| **rat783** | 783 | **21056** | 29216 | 88898 | 52664 | 8806 |
| **a280** | 280 | **4563** | 4650 | 13946 | 8401 | 2579 |
| **u159** | 159 | 70524 | **63158** | 85693 | 86408 | 42080 |
| **ch130** | 130 | 8043 | **7976** | 18765 | 10234 | 6110 |
| **bier127** | 127 | 198160 | 175822 | 185636 | **165814** | 118282 |
| **kroA100** | 100 | 40401 | **29864** | 60430 | 32715 | 21282 |
| **pr76** | 76 | 136353 | 153034 | 149491 | **132752** | 108159 |
| **berlin52** | 52 | **8567** | 9388 | 9428 | 9240 | 7542 |
| **att48** | 48 | **36816** | 37158 | 66425 | 39310 | 10628 |
| **eil51** | 51 | **445** | 464 | 476 | 494 | 426 |
| **pr144** | 144 | 120686 | **91780** | 295822 | 183067 | 58537 |

From Table (4.1), results indicate the superiority of the SBM algorithm in most of the problems, such as: rat87, a280, berlin52, att48, eil51. It converges to the optimal faster than the exchange method, and most of the rest of the test data (cities) were outperformed by the SAM algorithm, such as: u159, ch130, kroA100, pr144.

**Table 4.2.** Result for eleven problems obtained for 4 mutation operators after 8000 generations (population size=100)

| Name | #City | SBM | SAM | Rearrangement | Exchange mutation | Optimal |
|---|---|---|---|---|---|---|
| **rat783** | 783 | **29814** | 32290 | 83441 | 52680 | 8806 |
| **a280** | 280 | 4864 | **4712** | 8502 | 8278 | 2579 |
| **u159** | 159 | **75797** | 81311 | 111792 | 91940 | 42080 |
| **ch130** | 130 | **8968** | 9244 | 11457 | 10883 | 6110 |
| **bier127** | 127 | 207139 | 179888 | 193343 | **178595** | 118282 |
| **kroA100** | 100 | 32693 | **26841** | 40591 | 39695 | 21282 |
| **pr76** | 76 | 146481 | **136987** | 166636 | 146275 | 108159 |
| **berlin52** | 52 | 9723 | **8375** | 9517 | 9316 | 7542 |
| **att48** | 48 | 38834 | **35054** | 37049 | 42188 | 10628 |
| **eil51** | 51 | 451 | **440** | 771 | 466 | 426 |
| **pr144** | 144 | 112839 | **85107** | 148957 | 154783 | 58537 |

As can be seen from Table (4.2), the best performance was recorded by the SAM algorithm, which outperformed the other methods in seven problems: a280, KroA100, pr76, berlin52, att48, eil51, pr144, followed by the SBM algorithm, which overcame most of the rest, such as: rat87, u159, ch130.

The solutions of these algorithms are close to near optimal solutions and none could achieve an optimal solution. This shows the importance of using appropriate parameters along with crossover (such as population size, crossover ratio), due to the effective impact of their convergence to the optimal solution.

## 4.4 Summary

In this chapter ten methods and two strategies for mutation operator have been proposed. For the proposed types of mutation, each of them provides an heuristic search process for mutation. The two strategies called SBM and SAM are trying to apply more than one mutation operator. While SBM applies all the specific types of mutation to the same parent and retains a good child, SAM applies a certain type of mutation randomly every time, in the hope that the algorithm will choose a different type every time. Both strategies aim to encourage diversity in the population, through different directions of the search space, by applying multi mutation.

The proposed method was tested using the well-known problem (TSP). Comparisons were also made between the proposed type and the well-known exchange mutation and rearrangement operator.

# Chapter 5
# Combining the Proposed Strategies

This chapter aims to evaluate the effect of combining the proposed methods on GA performance, attempting to find the best combination of the proposed operators using two sets of experiments.

## 5.1 Experiments Set1

The aim of these experiments was to measure the effectiveness of the combination of proposed crossover, mutation operators and strategies, and especially Modified crossover, Collision crossover, SBC, SAC, Exchange mutation, SBM and SAM.

These operators and strategies were tested using 100 random cities, where the coordinates of the cities were chosen randomly. The GA parameters that were selected included the following: population size: 100, the probability of crossover: (100%), mutation probability: (100%), and the maximum generation was 1600.

In this test, operators and strategies that were combined included:

1. Exchange mutation with Modified crossover
2. Exchange mutation with Collision crossover
3. Exchange mutation with SBC
4. Exchange mutation with SAC
5. SBM with Modified crossover
6. SBM with Collision crossover
7. SBM with SBC
8. SBM with SAC
9. SAM with Modified crossover
10. SAM with Collision crossover
11. SAM with SBC
12. SAM with SAC

Results from these experiments indicated that the best performance was recorded by the "SBM with Modified crossover", followed by the "SAM with SBC", which compared well with the rest of the combination operators, because it showed good convergence to a minimum value.

The efficiency of each of the twelve combined methods is shown in Figure (5.1), which reveals that the "Exchange with SBC" integrated operator outperformed all other methods that merged with the Exchange mutation in the speed of convergence, followed by "Exchange with Modified crossover".

For operators combined with SBM, the Modified crossover, followed by Collision crossover, showed rapid convergence to a near optimal solution as compared to other combined operators.

Finally, “SAM with SBC”, followed by “SAM with Collision crossover”, showed a better result when compared with the SAM combined operators, because it showed good convergence to a minimum value.

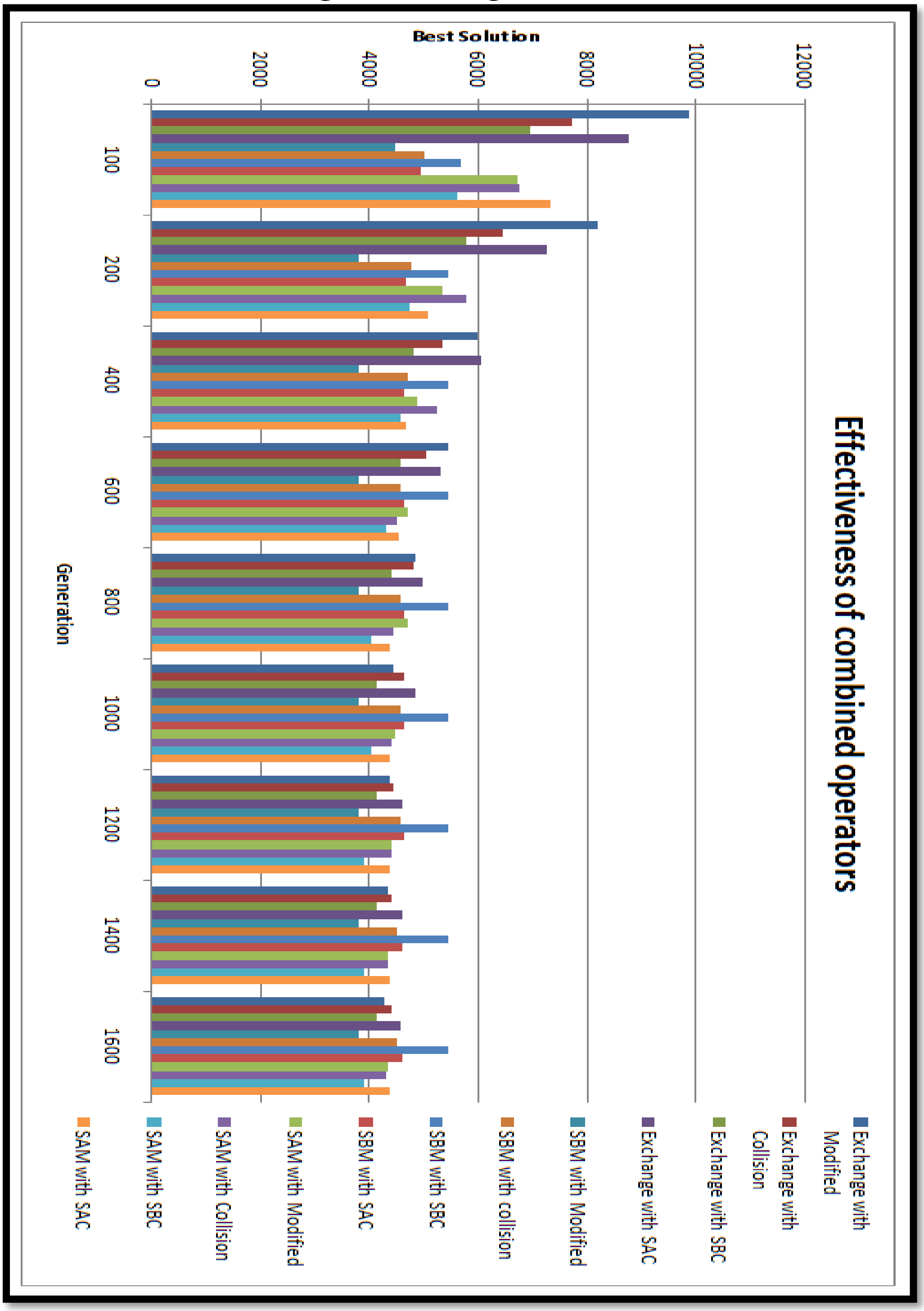

**Figure5.1.** Effectiveness of combined operators

## 5.2 Experiments Set 2

In this set of experiments, we wanted to compare the operators and strategies mentioned in the previous set of experiments.

The best combination of operator is the one that converges to an optimal solution or the best in-hand solution (near optimal). These operators and strategies were tested using eleven real TSP problems taken from the TSPLIB, including: rat783, a280, u159, ch130 bier127, kroA100, pr76, berlin52, att48, eil51and pr144.

The GA parameters that were selected in this test for all operators included the following: the crossover ratio: 100%, mutation ratio: 100%; the size of population was 100, and the maximum number of generations was 1600 (see Table (5.1)).

Each column in Table (5.1) represents the combination of operators, e.g. column1 represents “Exchange mutation with Modified crossover”, while column 2 represents “Exchange mutation with Collision crossover”, and so on.

As can be seen from the results in Table (5.1), the best performance was recorded by the “SAM with SBC”, which outperformed the other combined methods in four problems: a280, u159, bier127, berlin52, followed by “SBM with SAC” and “SAM with Modified crossover”. These combined methods converged to the near optimal solution of most of the rest of the test data over the other combined operators. This result confirms powerful use of more than one operator, and thus avoids falling into local minima.

The best combination operator with Exchange mutation is SBC, which converges to the optimal in most of the problems faster than the other methods with Exchange mutation, such as: rat87, a280, u159, ch130, bier127, pr144, followed by Exchange mutation with Modified crossover, which overcame most of the rest of the problems, such as:kroA100, pr76, att48.

“SBM combined with Collision crossover” recorded the best convergence to the near optimal solution, and showed faster convergence to the minimum value better than any mixed with SBM in most of the problems, such as: a280, u159, ch130, bier127,pr144,followed by “SBM with SAC”, where it overcame most of the rest of the problems, such as: rat783, kroA100, berlin52, att48, eil51.

Finally, “SAM combined with SBC” scored the best result compared to the rest of the methods together with SAM. The problems that outperformed them were: rat87, a280, u159, bier127, pr76, berlin52, followed by “SAM with Modified” and “SAM with SAC”.

It is worth mentioning that the solutions of these methods are close to near optimal solutions and none could achieve an optimal solution. This shows the importance of using appropriate parameters such as population size, crossover ratio, mutation ratio and maximum number of generations, because of the effective impact of their convergence to the optimal solution as described in the first chapter, knowing that GAs do not guarantee an optimal solution (Eiben, Michalewicz, Schoenauer, & Smith, 2007).

**Table 5.1.** Comparison of the twelve combined methods

| Name | 1 | 2 | 3 | 4 | 5 | 6 | 7 | 8 | 9 | 10 | 11 | 12 | Optimal |
|---|---|---|---|---|---|---|---|---|---|---|---|---|---|
| rat783 | 58219 | 61396 | 47203 | 61551 | 38823 | 39492 | **30010** | 30870 | 41280 | 46032 | 38183 | 40891 | **8806** |
| a280 | 7061 | 8366 | 6289 | 7666 | 5336 | 4989 | 5024 | 5079 | 5376 | 6300 | **4841** | 5737 | **2579** |
| u159 | 7139 | 93806 | 68017 | 72810 | 78249 | 71589 | 75962 | 80825 | 71933 | 85509 | **64268** | 80120 | **42080** |
| ch130 | 9221 | 12339 | 9039 | 10393 | 10815 | 9243 | 9559 | 9584 | **8850** | 10168 | 9535 | 9642 | **6110** |
| bier127 | 156334 | 189704 | 153901 | 162715 | 185555 | 181600 | 196323 | 187233 | 166540 | 184090 | **144722** | 161934 | **118282** |
| kroA100 | 30555 | 35574 | 34227 | 31855 | 31837 | 33931 | 30548 | 29264 | **26033** | 37706 | 32834 | 28417 | **21282** |
| pr76 | **124370** | 159666 | 144437 | 155023 | 125305 | 153020 | 155901 | 149463 | 139608 | 137146 | 129742 | 160265 | **108159** |
| berlin52 | 8905 | 8917 | 9437 | 8603 | 9090 | 8829 | 9018 | 8656 | 8749 | 8609 | **8631** | 9590 | **7542** |
| att48 | 36378 | 40260 | 39364 | 44315 | 35459 | 36950 | 35977 | **35039** | 38190 | 35273 | 36115 | 38508 | **10628** |
| eil51 | 492 | 448 | 520 | 466 | 454 | 465 | 449 | **444** | 511 | 461 | 464 | 460 | **426** |

## 5.3 Time complexity of the proposed work

The time complexity of the proposed work is the same as the traditional GA's time complexity, except for the SBC and SBM, where a number of crossovers and mutations were tested to find the best operation among a fixed number of operations; hence it is a constant number of operations. No significant time complexity will be added to the overall time complexity of traditional GAs, i.e. the order of the time complexity will be the same.

The time complexity of the GAs depends mainly on the representation of the chromosome (the size of the chromosome, which varies depending on the problem), the genetic parameters (number of generations, population size, mutation probability and crossover probability), and the selection time.

Consequently, the time complexity of the GA can be calculated as follows:

Suppose that the size of population is (P), the number of generations is (G), the probability of crossover and mutation are (PC and PM) respectively, and the chromosome size is (n), which is the variable number of cities, normally, $n >> $ a constant.

If we use the Quick Sorting algorithm to sort the population and take the best solutions, the number of operations needed for the selection process is:

$$(P \times \log(P))$$

The number of the mutated chromosomes is equal to ($PM \times P$), since we need to go along each chromosome; the number of operations needed for the mutation process is:

$$(PM \times P \times n)$$

The number of crossovers is equal to ($PC \times P$), since we need to go along each chromosome; the number of operations needed for the crossover process is:

$$(PC \times P \times n)$$

All the previous operations will be repeated (G) times. Therefore the time complexity of the overall GA (GATC) is:

$$GATC = G( (P \times \log(P)) + (PM \times P \times n) + (PC \times P \times n) ) \qquad (13)$$

Since, G and P are constants (ks) regardless of what the problem is, assuming that $k1=G$, $k2=P$, $k3= PM \times P$, $k4= PC \times P$ then theoretically :
$k=k1=k2=k3=k4$, and therefore GATC becomes:

$$GATC= k( (k \times \log(k)) + (k \times n) + (k \times n) ) \qquad (14)$$
$$= k^2 \log(k) + 2n\, k^2$$

Since k is a constant, then $k^2$ and $\log(k)$ are also constants, which makes equation (14) become:

$$GATC= k + (kn) \qquad (15)$$

and therefore the time complexity of the traditional GAs becomes linear (of order (*n*)), i.e. GATC depends mainly on the problem size (chromosome size) as it is the only variable in the process:

$$GATC= O(n) \qquad (16)$$

## 5.4 Summary

In this chapter we have presented two sets of experiments conducted on some of the proposed methods that provided the best results in crossover and mutation to find the best combination of three mutation operators Exchange mutation, SBM and SAM and four crossover operators Modified crossover, Collision crossover, SAC and SBC to create twelve combinations.

The combination operators were tested using two sets of test data. The first test used random cities and the second test used eleven real TSP problems taken from the TSPLIB. Comparison of these combination operators was also carried out.

At the end of the chapter, the time complexity of the proposed work was described and explained.

# Chapter 6
# Conclusions and Future Work

## 6.1 Conclusions

In this thesis new operators and strategies for GA are proposed. The new operators for both mutation and crossover trace particular guidance in every operator.

For the crossover operator we have proposed three crossover methods, namely: COWGC, COWLRGC and Collision Crossovers which are based on the physical rules of elastic collision, and they are considered one of the multi-point crossovers. Several experiments were conducted on several TSP problems to evaluate those methods which were compared to the two most well-known operators, namely: Modified crossover and PMX crossover.

For mutation operator we have proposed several mutation methods (WGWRGM, WGWWGM, WLRGWRGM, WGWNNM, WGWWNNM, WGIBNNM, RGIBNNM, SWGLM, IBRGBWGM and IBRGBRGM), and several experiments were conducted to evaluate those methods on several TSP problems. These were compared to the two most well-known operators, namely: exchange mutation and rearrangement mutation.

To overcome the problem of decision with regard to any crossover or mutation appropriate for use with any problem, four strategies are proposed: SAC, SBC, SBM, SAM for crossover and mutation respectively. These algorithms apply multiple crossover/mutation operators at the same time, and such strategies encourage diversity in the population, and thus avoid falling into local optima.

The experiments demonstrated the superiority of the four strategies (SAC, SBC, SBM, SAM) on the use of crossover/mutation alone.
The best combination method in this thesis is the SAM with SBC. These results confirm the findings of Contreras-Bolton and Parada where they stated:“The process of searching for good combinations was effective, yielding appropriate and synergic combinations of the crossover and mutation operators” (Contreras-Bolton & Parada, 2015). Another study by Spears, in which he stated: “But much of the performance stems from simply having two crossover operators at the GA’s disposal” (Spears, 1995).

At the level of crossover alone, the collision operator showed superiority over all crossover operators. The other proposed crossover methods are useful for SBC and SAC strategies because they encourage diversity through new patterns of individual. Regarding the use of each mutation alone, some mutations showed better performance than others, and this does not mean that the rest of the mutations have been proved to fail. They can be effective in dealing with other problems, because every problem has a search space different from the search space of other problems. In our work they can be effective in SBM and SAM, where they encourage diversity and hence increase the efficiency of both algorithms.

The experiments conducted in this thesis have also shown that the use of mutation alone or crossover alone is not enough to enhance the performance of the GA.

## 6.2 Future work

We are planning to develop some types of new crossovers and mutations and to apply the proposed methods to different problems, such as the Knapsack problem. In addition, we are planning to develop a new strategy for the initial population to make it more appropriate before starting the search.

The process of tuning parameter is exhausting and needs time, so we plan to use the self-adaptive parameter principle (Pellerin, Pigeon, & Delisle, 2004) in order to overcome parameter tuning such as mutation ratio and crossover ratio.

We will apply the proposed methods to the MPGA, and from this perspective we are planning to develop new types of migrations between the sub-population of MPGA using several basic trends, such as using a different topology, and depending on the negative Assortative mating concept, we are also thinking of developing migration depending on the amount of diversity of each sub-population. Thus, we are seeking to improve on the results of a GA to reach the optimal solution or near to optimal solution.

In addition, we are thinking of a way to encourage diversity in the MPGA, through the use of the concept of multi-operators. We consider that with the use of multi-migration strategies and with or without a dynamic migration rate and migration interval, we hope that in each period of different generations, there will be a certain type of migration, thus promoting the concept of out-breeding.

## المعلومات الشخصية

**الاسم:** إسراء عمر الكفاوين

**التخصص:** ماجستير علم الحاسوب

**الكلية:** العلوم

**السنة:** 2015